\documentclass[10pt,twocolumn,letterpaper]{article}

\usepackage[pdftex]{graphicx}
\usepackage{float}
\usepackage[justification=raggedright]{caption}	
\usepackage{lscape}                                         
\usepackage{wrapfig}

\usepackage[lined,ruled,linesnumbered]{algorithm2e}
\usepackage{animate} 

\usepackage{booktabs}                   
\usepackage{multirow}

\usepackage{makecell}

\usepackage{paralist}
\usepackage{enumitem}
\setlist[enumerate]{itemsep=0mm}

\usepackage{bm}                          
\usepackage{epsfig}                       
\usepackage{times}
\usepackage{mathptmx}
\usepackage{mathtools}
\usepackage{amssymb,amsmath}   

\usepackage{units}
\usepackage{color}

\usepackage{comment}

\usepackage{url}  
\usepackage{hyperref}
\hypersetup{pagebackref,breaklinks,colorlinks,linkcolor=blue,citecolor=blue,urlcolor=blue,bookmarks=false}
\usepackage{xspace}
\usepackage[table]{xcolor}
\usepackage{setspace}
\usepackage{grfext}
\PrependGraphicsExtensions*{.jpg,.png,.PNG}

\usepackage{gensymb}

\usepackage{soul}
\usepackage{svg}

\usepackage[accsupp]{axessibility} 

\usepackage[final]{changes}

\def\eg{e.g.,~}               
\def\ie{i.e.,~}               

\DeclareMathAlphabet{\altmathcal}{OMS}{cmsy}{m}{n}
\DeclareMathAlphabet{\mathbfit}{OT1}{ptm}{bx}{it}

\newlength\paramargin
\newlength\figmargin

\newlength\secmargin
\newlength\figcapmargin
\newlength\tabcapmargin

\newcommand{\mpage}[2]
{
\begin{minipage}{#1\linewidth}\centering
#2
\end{minipage}
}

\newcommand{\topic}[1]
{
\vspace{1mm}\noindent\textbf{#1}
}

\long\def\ignorethis#1{}

\newbox\jsavebox%

\makeatletter
\newcommand{\providelength}[1]{%
  \@ifundefined{\expandafter\@gobble\string#1}
   {
    \typeout{\string\providelength: making new length \string#1}%
    \newlength{#1}%
   }
   {
    \sdaau@checkforlength{#1}%
   }%
}

\newcommand{\sdaau@checkforlength}[1]{%
  \edef\sdaau@temp{\expandafter\sdaau@getfive\meaning#1TTTTT$}%
  \ifx\sdaau@temp\sdaau@skipstring
    \typeout{\string\providelength: \string#1 already a length}%
  \else
    \@latex@error
      {\string#1 illegal in \string\providelength}
      {\string#1 is defined, but not with \string\newlength}%
  \fi
}
\def\sdaau@getfive#1#2#3#4#5#6${#1#2#3#4#5}
\edef\sdaau@skipstring{\string\skip}
\makeatother

\usepackage[capitalize]{cleveref}
\crefname{section}{Sec.}{Secs.}
\Crefname{section}{Section}{Sections}
\Crefname{table}{Table}{Tables}
\crefname{table}{Tab.}{Tabs.}

\def\xi{\mathbf{x}_i}

\graphicspath{{figures}, {examples}}

\usepackage{textcomp}  
\usepackage{scalerel}  
\def\emojidog{\scalerel*{\includegraphics{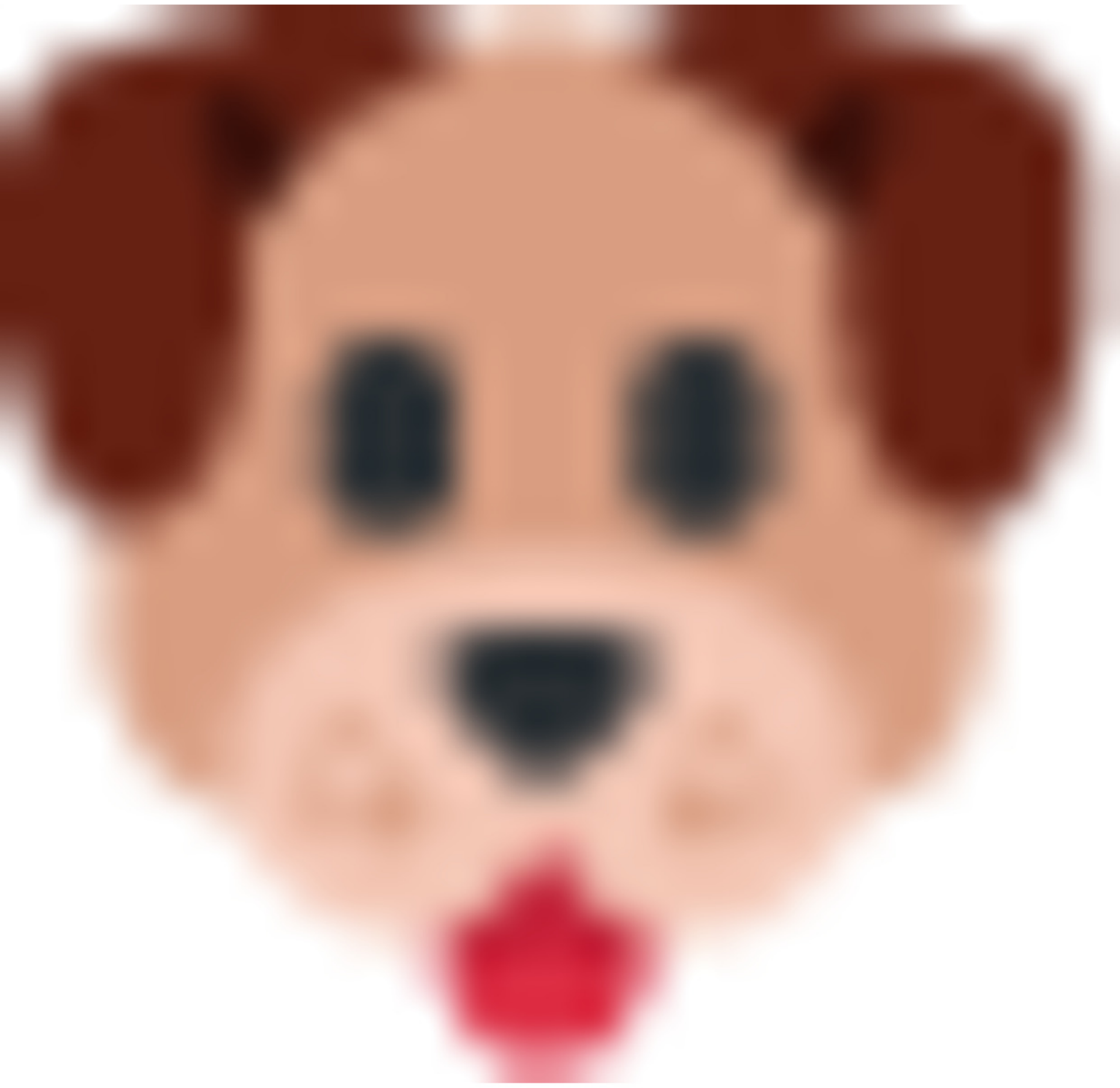}}{\textrm{\textbigcircle}}}
\def\emojifox{\scalerel*{\includegraphics{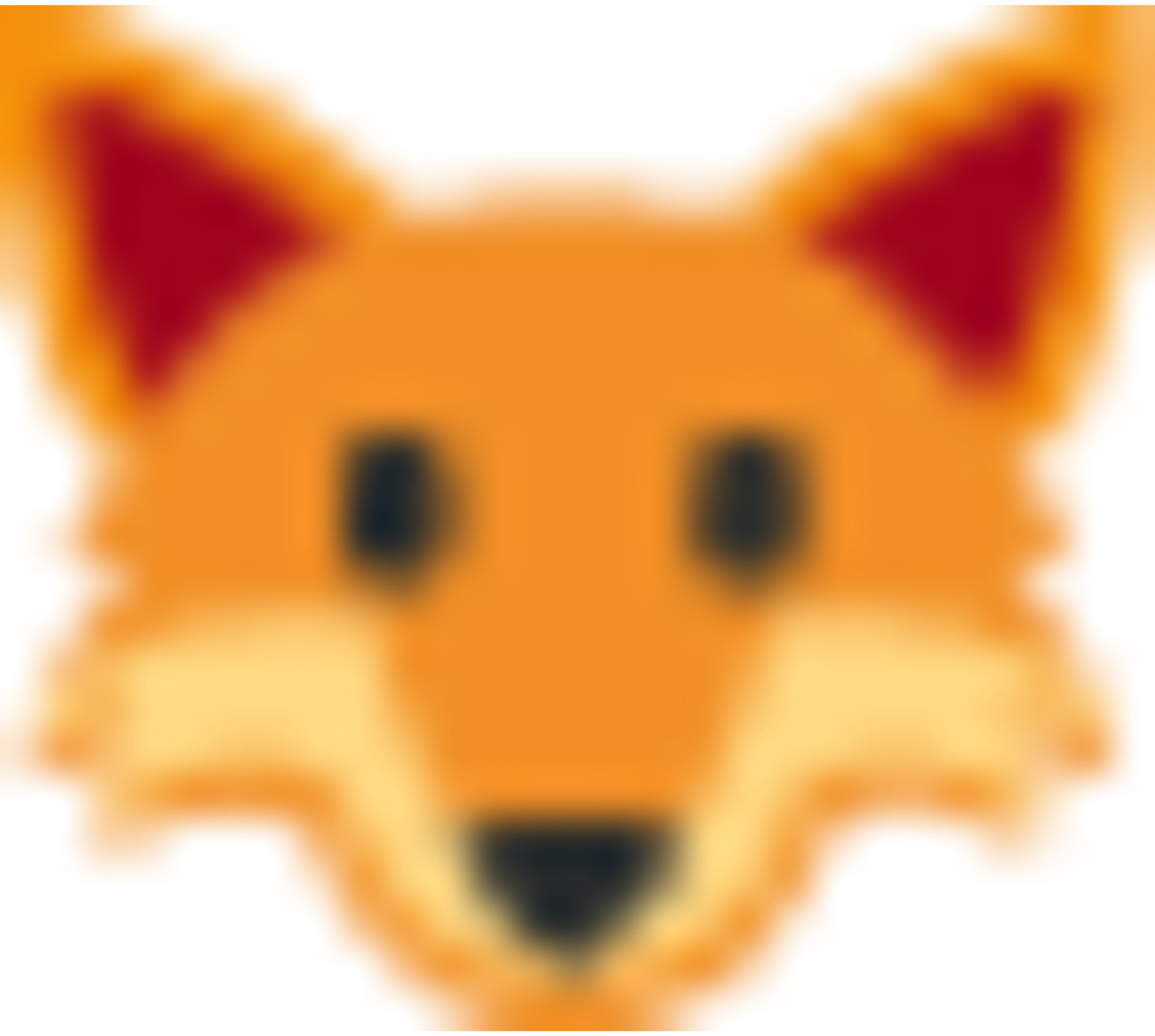}}{\textrm{\textbigcircle}}}
\def\emojilion{\scalerel*{\includegraphics{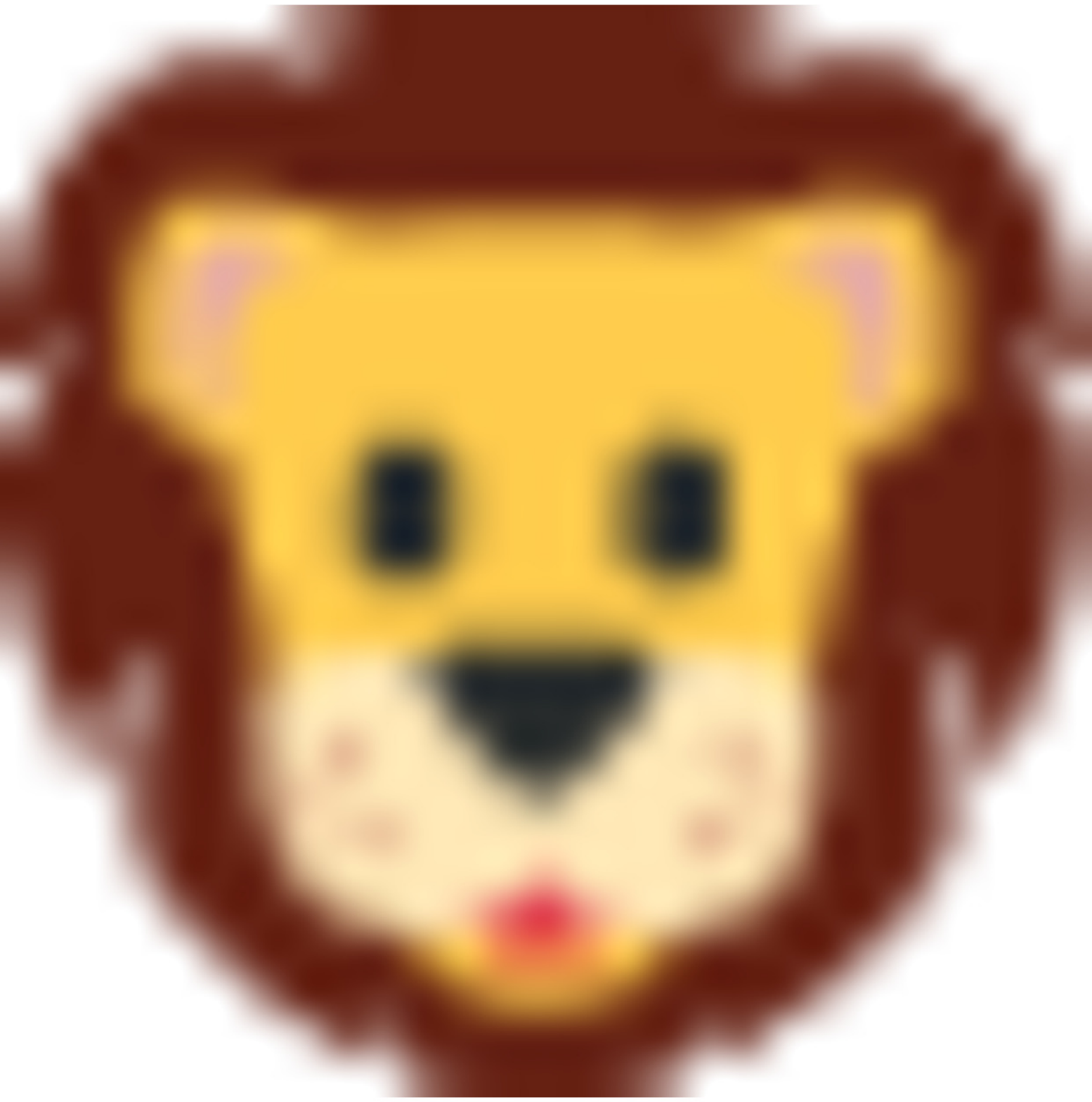}}{\textrm{\textbigcircle}}}
\def\emojitiger{\scalerel*{\includegraphics{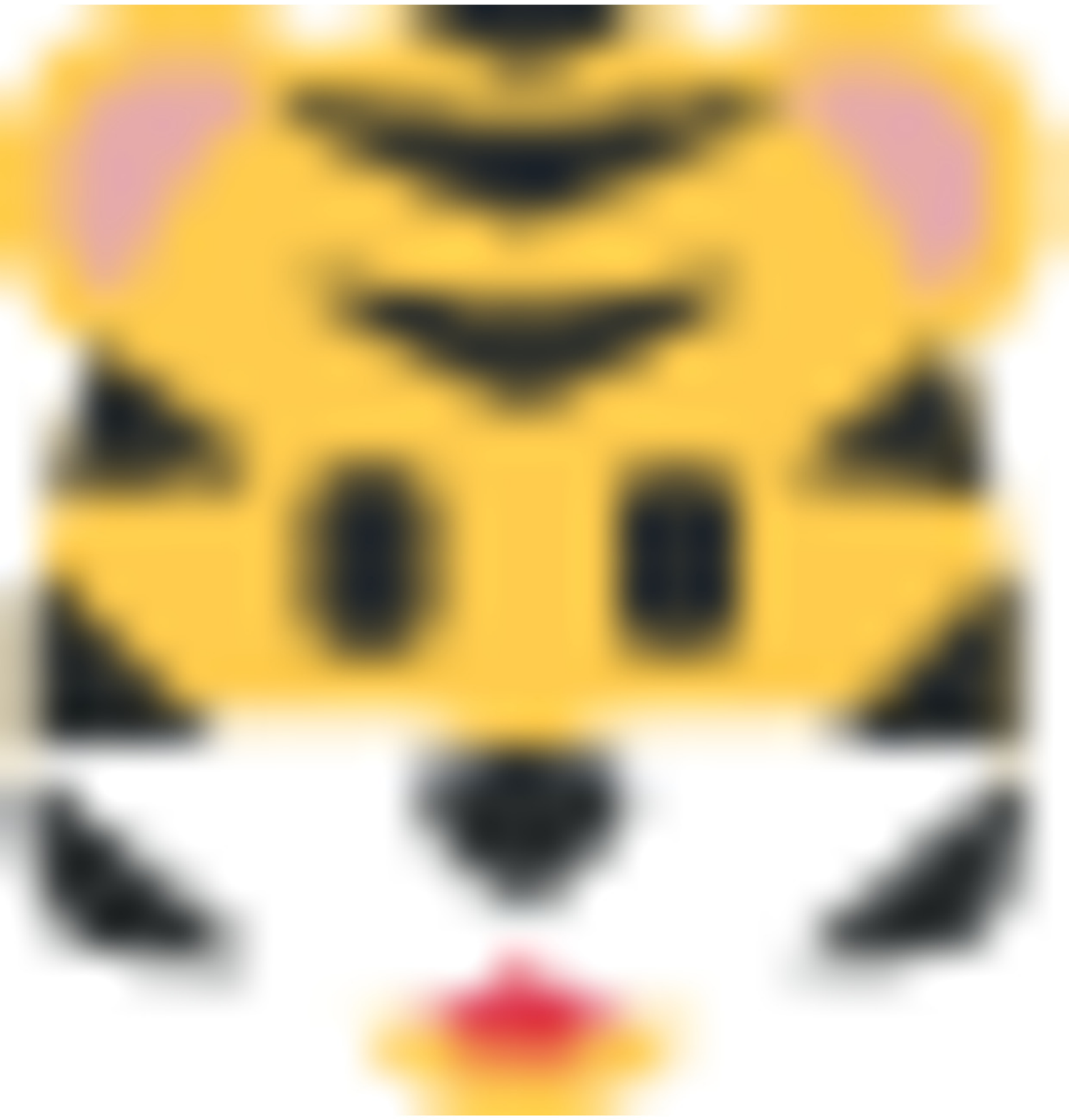}}{\textrm{\textbigcircle}}}
\def\emojiwolf{\scalerel*{\includegraphics{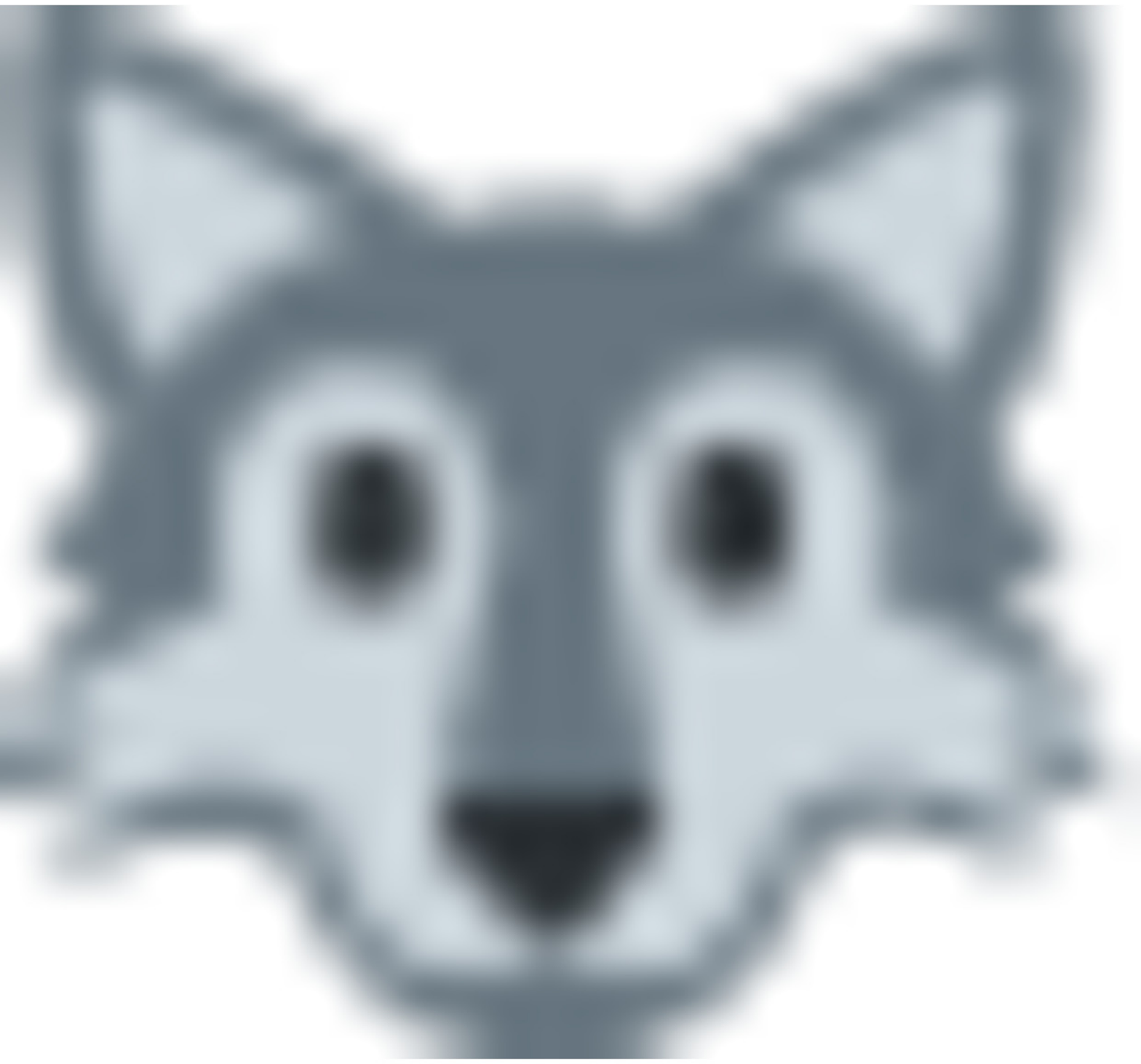}}{\textrm{\textbigcircle}}}
\def\emojihamster{\scalerel*{\includegraphics{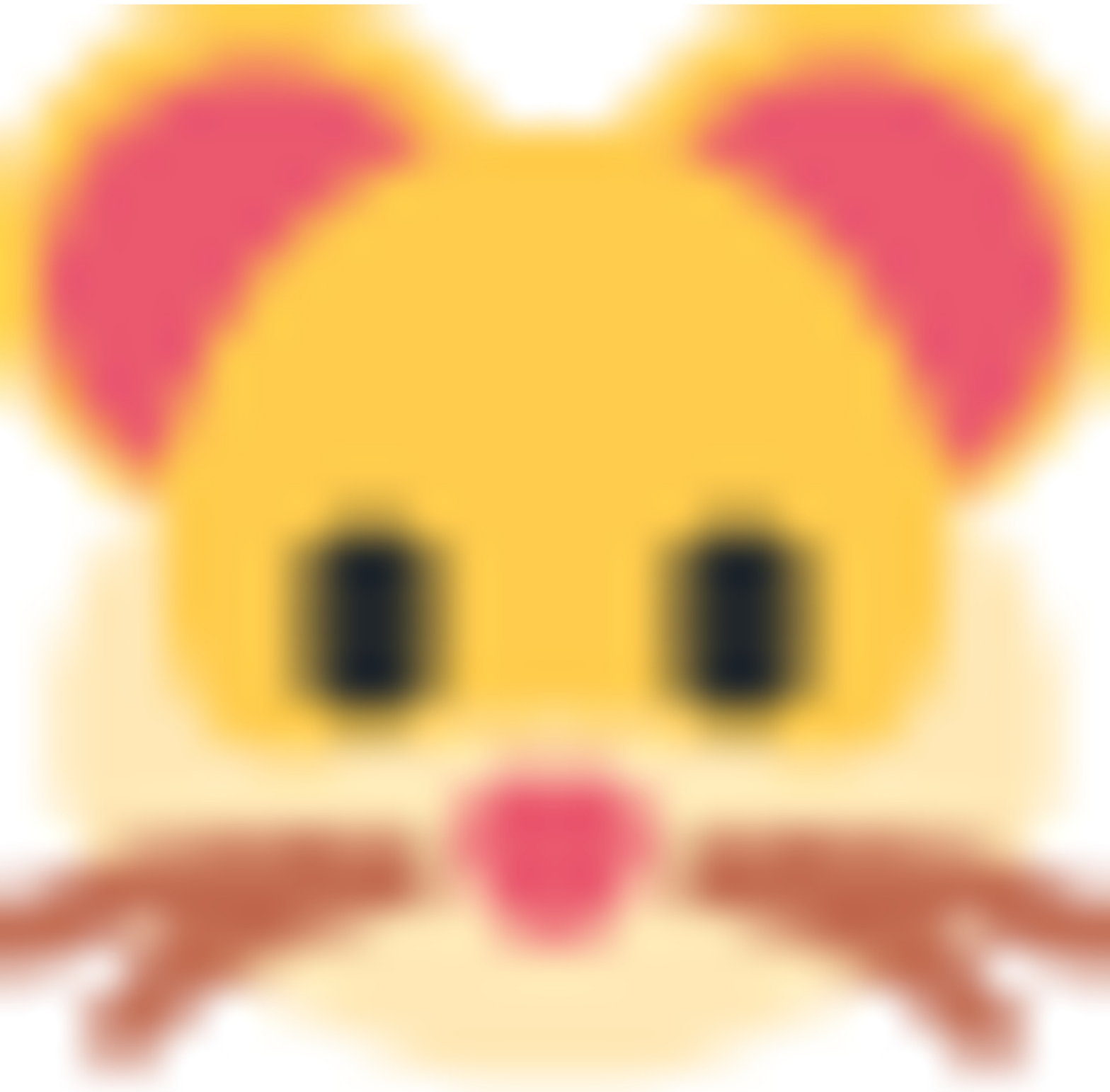}}{\textrm{\textbigcircle}}}
\def\emojibadger{\scalerel*{\includegraphics{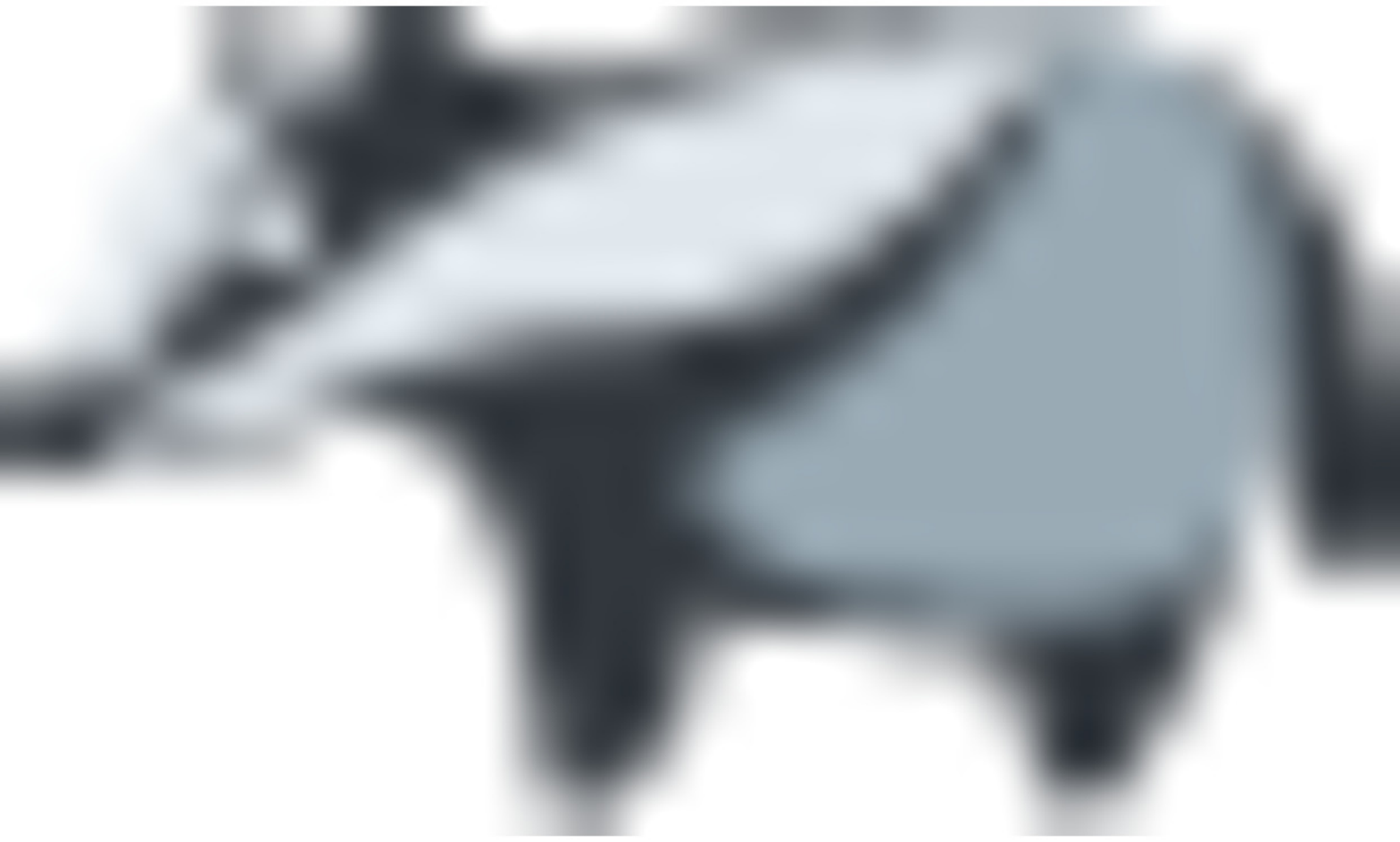}}{\textrm{\textbigcircle}}}
\def\emojiotter{\scalerel*{\includegraphics{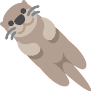}}{\textrm{\textbigcircle}}}
\def\emojibear{\scalerel*{\includegraphics{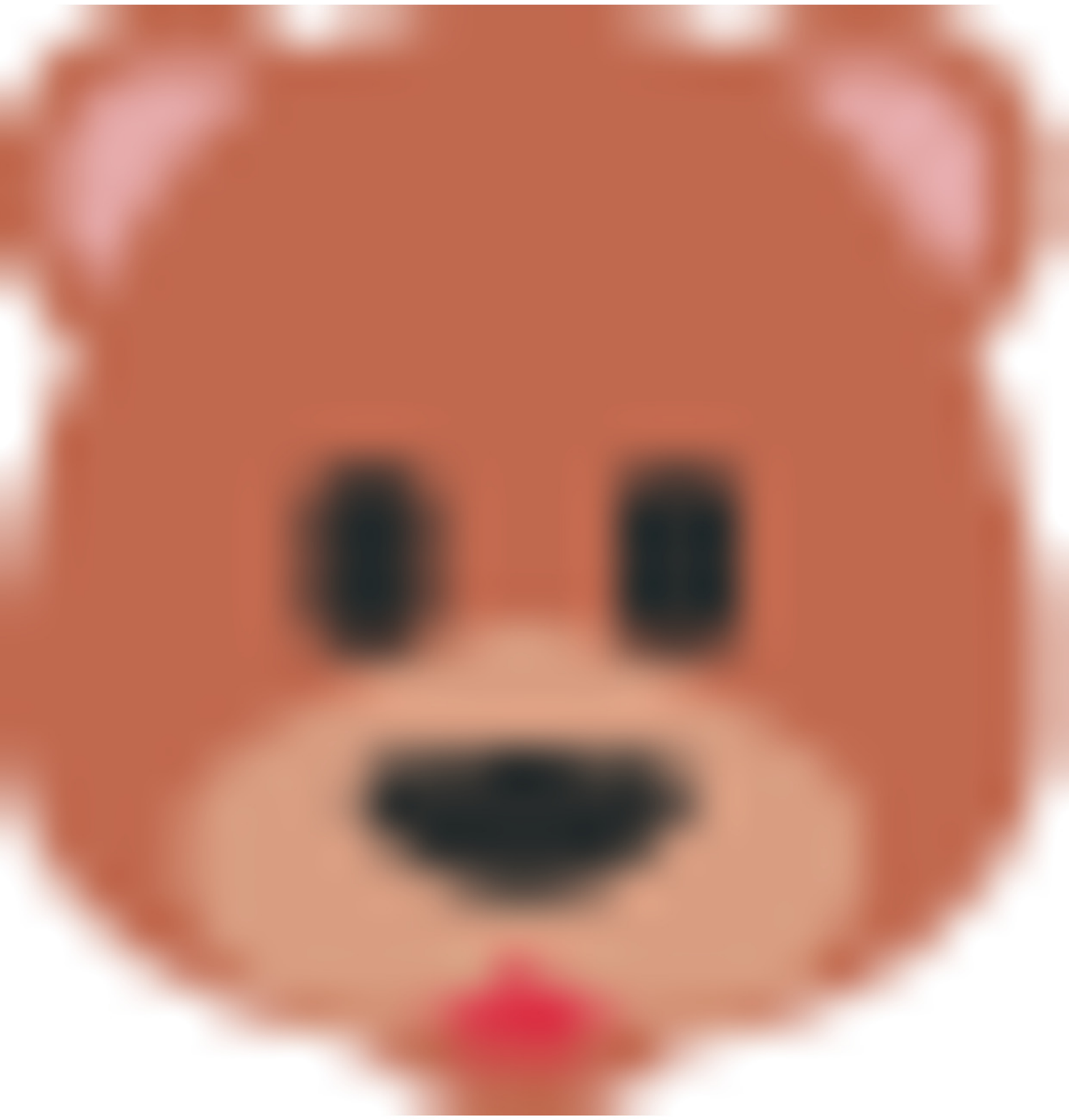}}{\textrm{\textbigcircle}}}
\def\emojipig{\scalerel*{\includegraphics{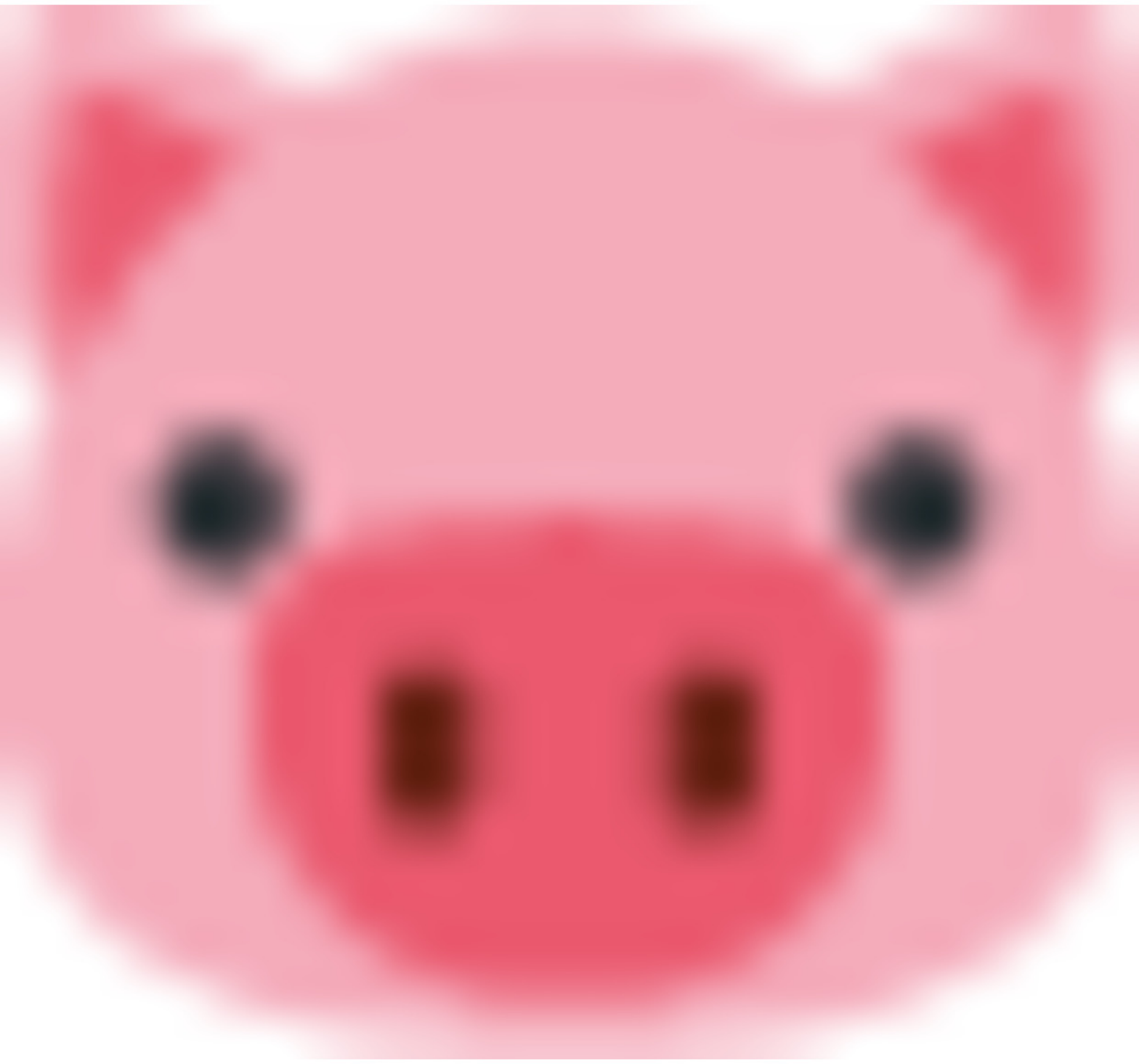}}{\textrm{\textbigcircle}}}
 
\makeatletter

\def\@fnsymbol#1{\ensuremath{\ifcase#1\or \dagger\or \ddagger\or
\mathsection\or \mathparagraph\or \|\or **\or \dagger\dagger
\or \ddagger\ddagger \else\@ctrerr\fi}}
\makeatother

\graphicspath{{figures/}}

\usepackage{iccv}

\iccvfinalcopy 

\def\iccvPaperID{7123} 

\ificcvfinal\fi

\begin{document}

\title{
Text-driven Visual Synthesis with Latent Diffusion Prior
}

\author{\hspace{-0.4cm}Ting-Hsuan Liao
\hspace{0.65cm}
Songwei Ge
\hspace{0.65cm}
Yiran Xu
\hspace{0.65cm}
Yao-Chih Lee
\hspace{0.65cm}
Badour AlBahar
\hspace{0.65cm}
Jia-Bin Huang\\
\vspace{0.2cm}
University of Maryland, College Park\\
\url{http://latent-diffusion-prior.github.io} \\
}

\twocolumn[{
\renewcommand\twocolumn[1][]{#1}
\maketitle
\begin{center}
\centering

\mpage{0.02}{\raisebox{-10pt}{\rotatebox{90}{Text-to-3D}}}
\mpage{0.23}{\frame{\includegraphics[width=\linewidth, trim=0 0 0 1cm, clip]{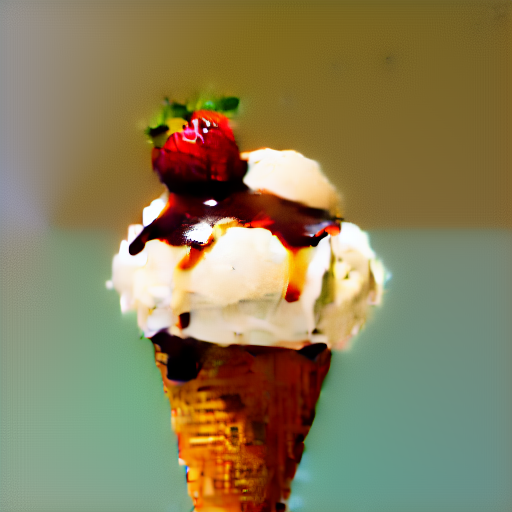}}}\hfill
\mpage{0.23}{\frame{\includegraphics[width=\linewidth, trim=0 1cm 0 0, clip]{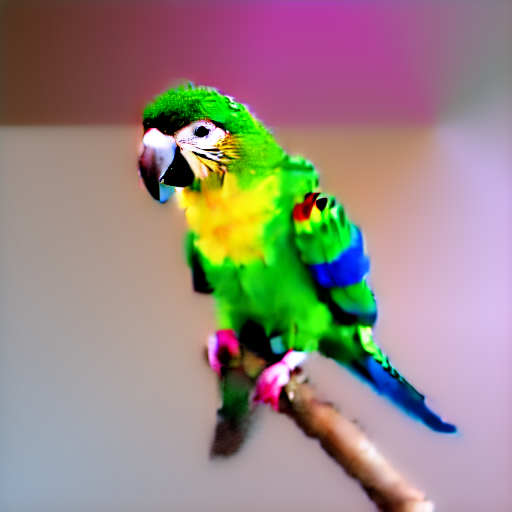}}}\hfill
\mpage{0.23}{\frame{\includegraphics[width=\linewidth, trim=0 0 0 1cm, clip]{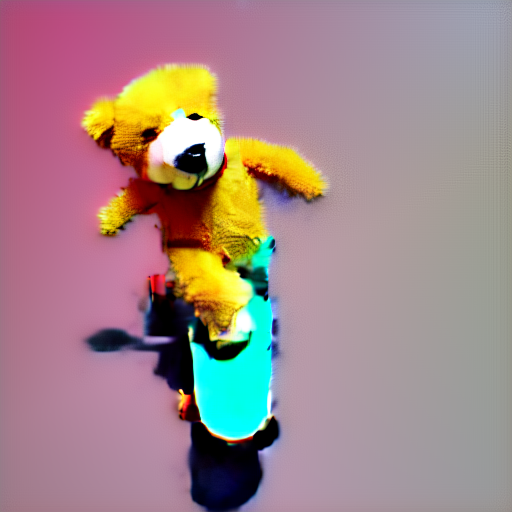}}}\hfill
\mpage{0.23}{\frame{\includegraphics[width=\linewidth, trim=0 0 0 1.9cm, clip]{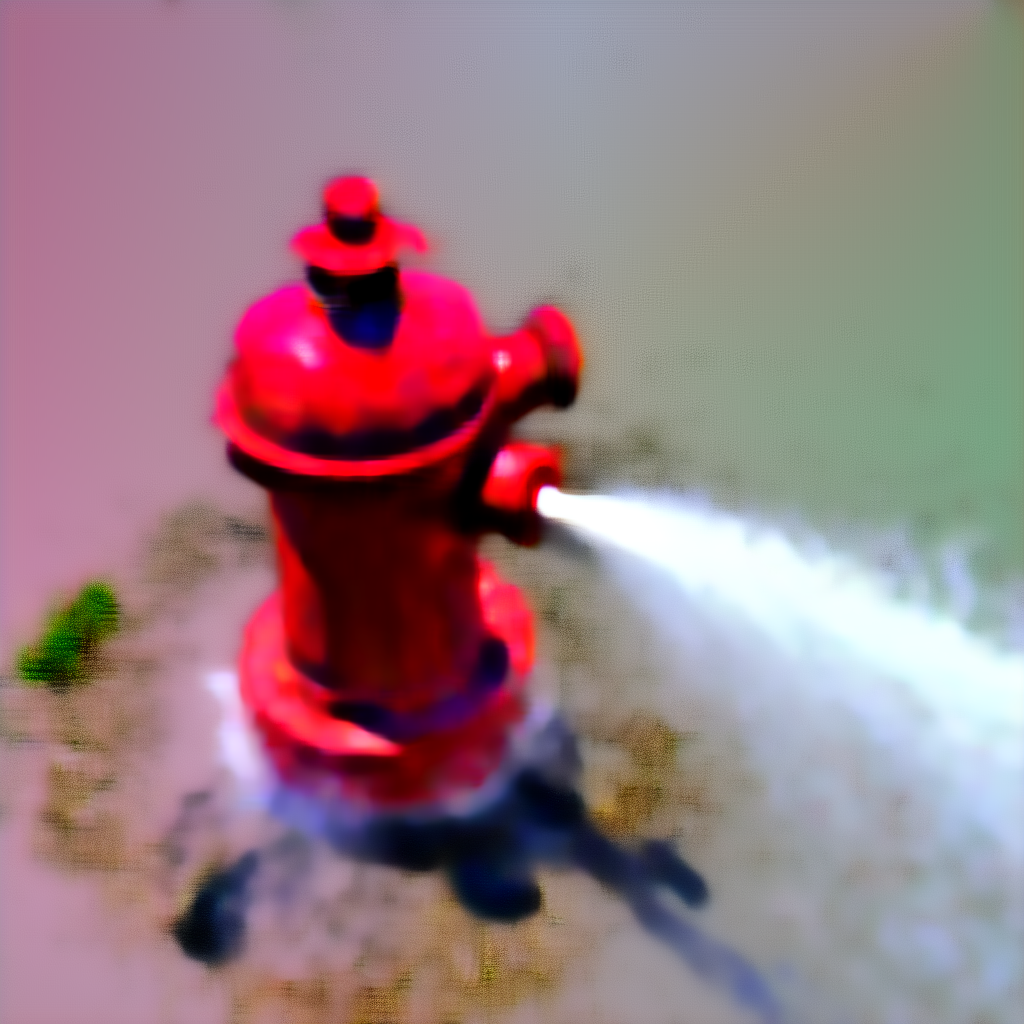}}}\\

\vspace{0.3mm}
\mpage{0.02}{\raisebox{-10pt}{\rotatebox{90}{}}}
\mpage{0.23}{\small{``A high quality photo of an ice cream sundae.''}}\hfill
\mpage{0.23}{\footnotesize{``A high-quality photo of a colorful parrot, highly detailed.''}}\hfill
\mpage{0.23}{\small{``A stuffed animal that is frowning is on a skateboard.''}}\hfill
\mpage{0.23}{\small{``A red fire hydrant spraying water.''}}\\

\vspace{0.5mm}

\mpage{0.02}{\raisebox{-10pt}{\rotatebox{90}{Image editing}}}
\mpage{0.23}{\frame{\includegraphics[width=\linewidth, trim=0 0 0 0, clip]{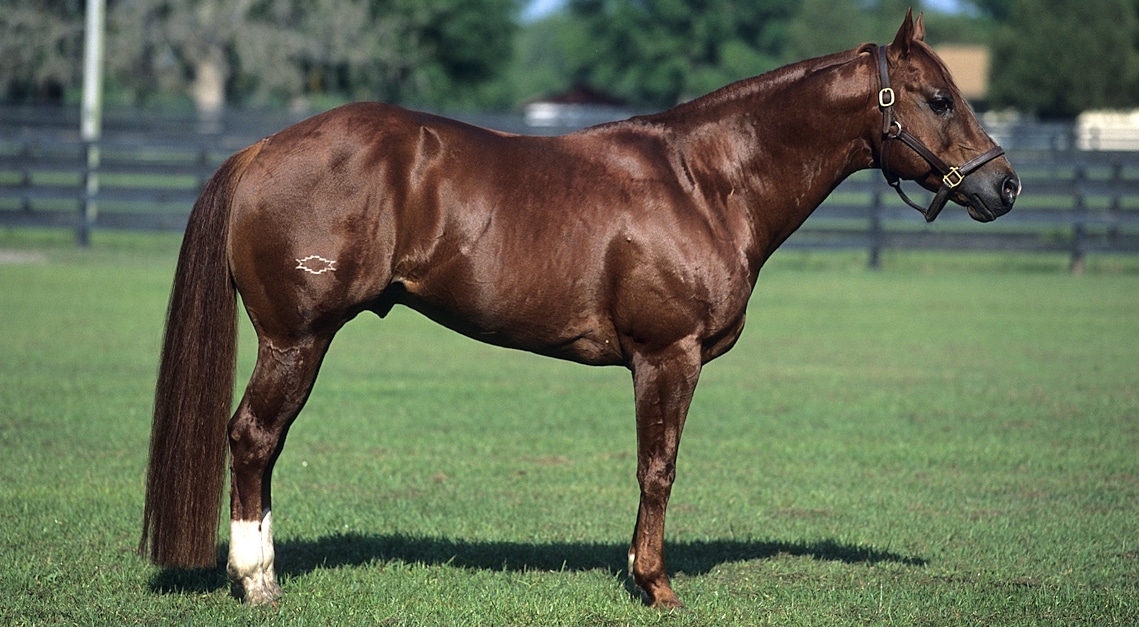}}}\hfill
\mpage{0.23}{\frame{\includegraphics[width=\linewidth, trim=0 0 0 0, clip]{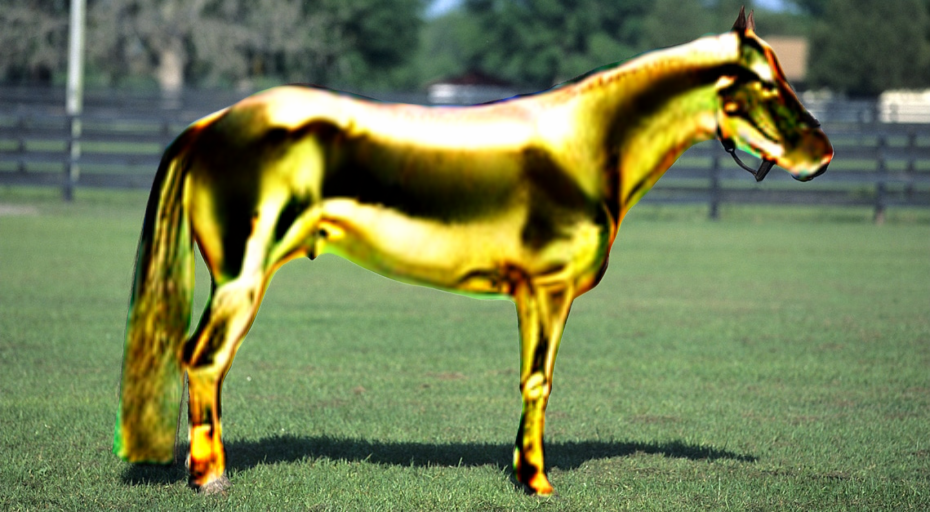}}}\hfill
\mpage{0.23}{\frame{\includegraphics[width=\linewidth, trim=0 0 0 0, clip]{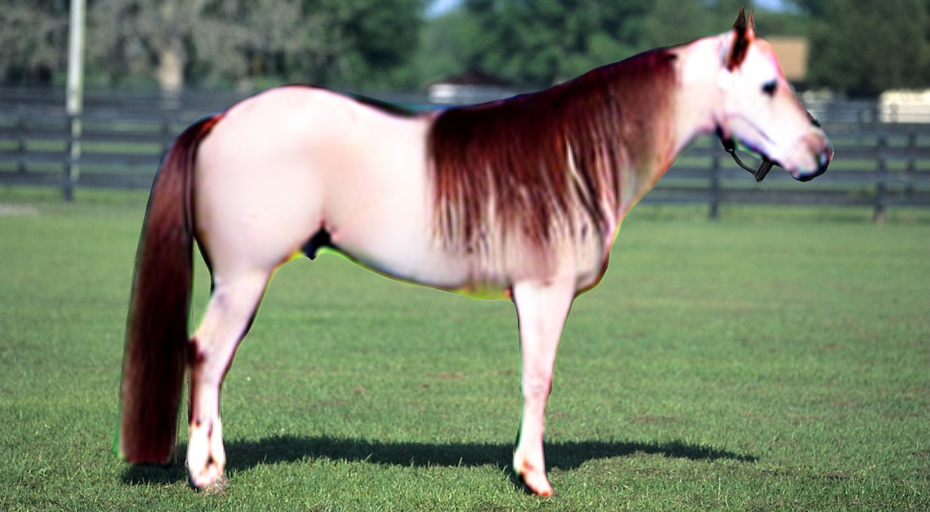}}}\hfill
\mpage{0.23}{\frame{\includegraphics[width=\linewidth, trim=0 0 0 0, clip]{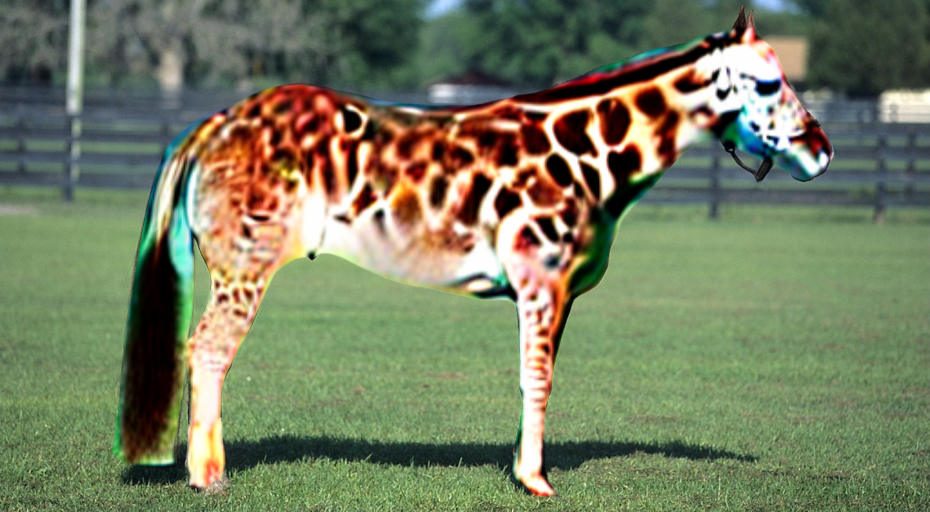}}}\\
\vspace{0.3mm}
\mpage{0.02}{\raisebox{-10pt}{\rotatebox{90}{}}}
\mpage{0.23}{Input}\hfill
\mpage{0.23}{\small{``golden horse''}}\hfill
\mpage{0.23}{\small{``long-mane horse''}}\hfill
\mpage{0.23}{\small{``horse with giraffe skin''}} \\

\vspace{0.5mm}

\mpage{0.02}{\raisebox{-10pt}{\rotatebox{90}{StyleGAN adaptation}}}
\mpage{0.23}{\frame{\includegraphics[width=\linewidth, trim=0 0 0 0, clip]{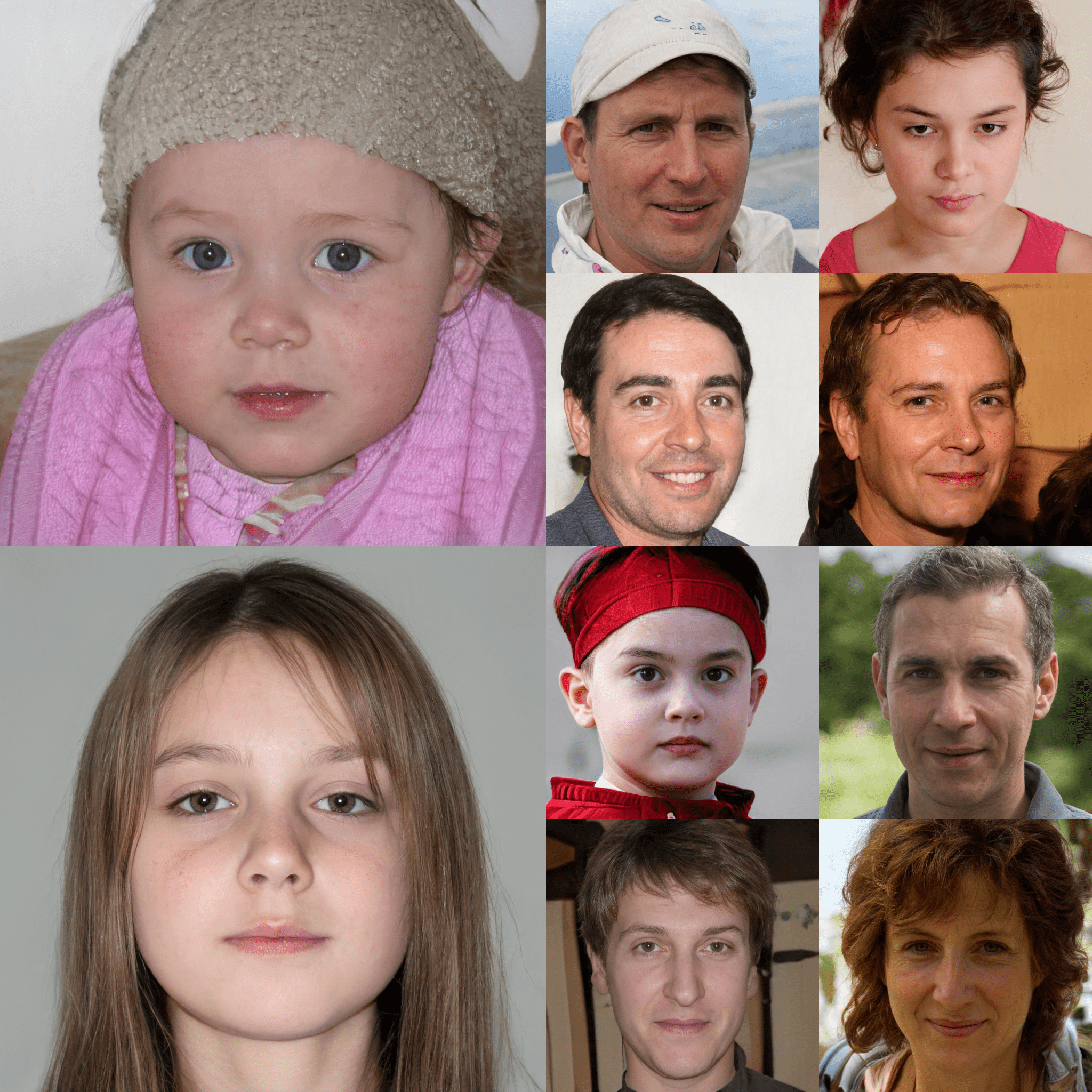}}}\hfill
\mpage{0.23}{\frame{\includegraphics[width=\linewidth, trim=0 0 0 0, clip]{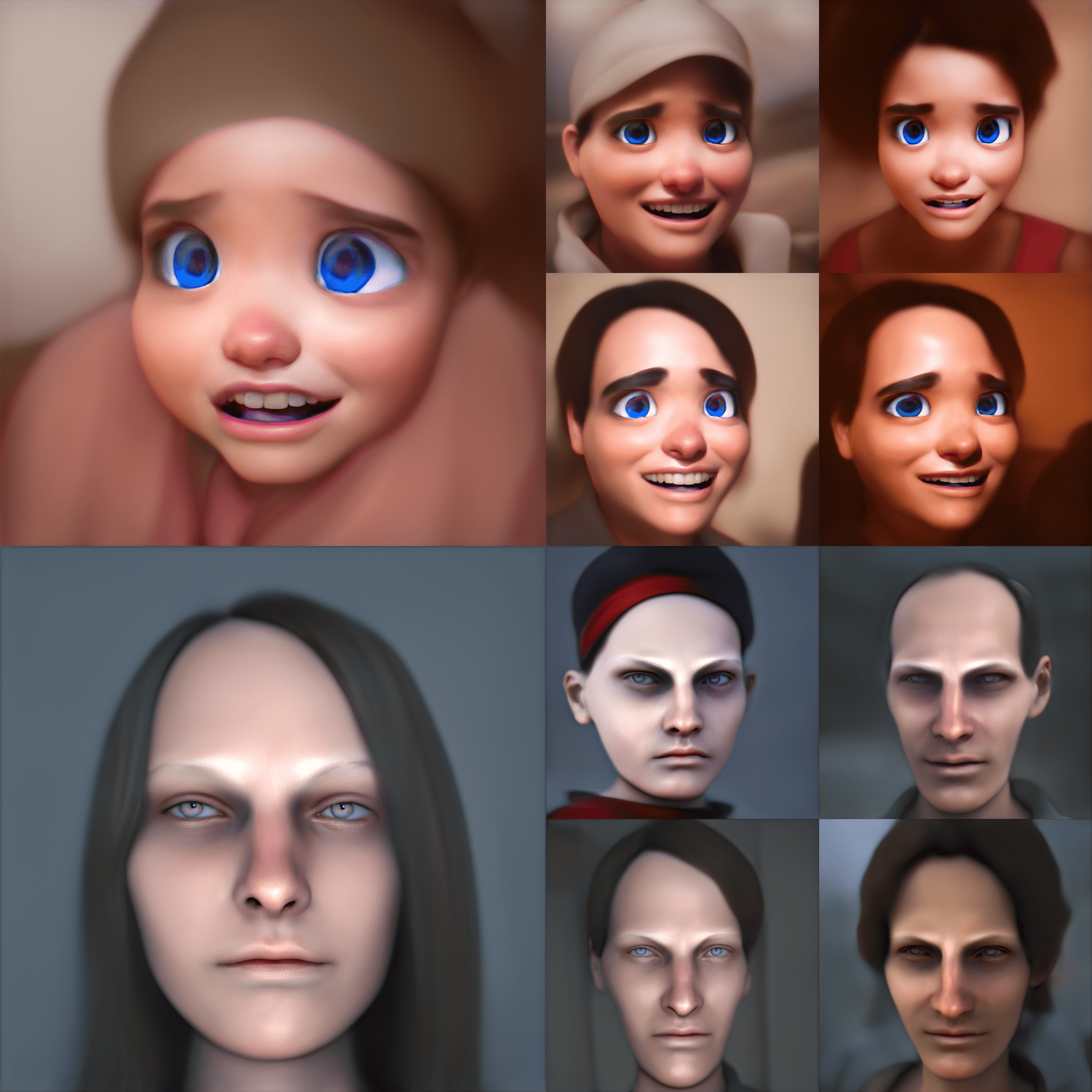}}}\hfill
\mpage{0.23}{\frame{\includegraphics[width=\linewidth, trim=0 0 0 0, clip]{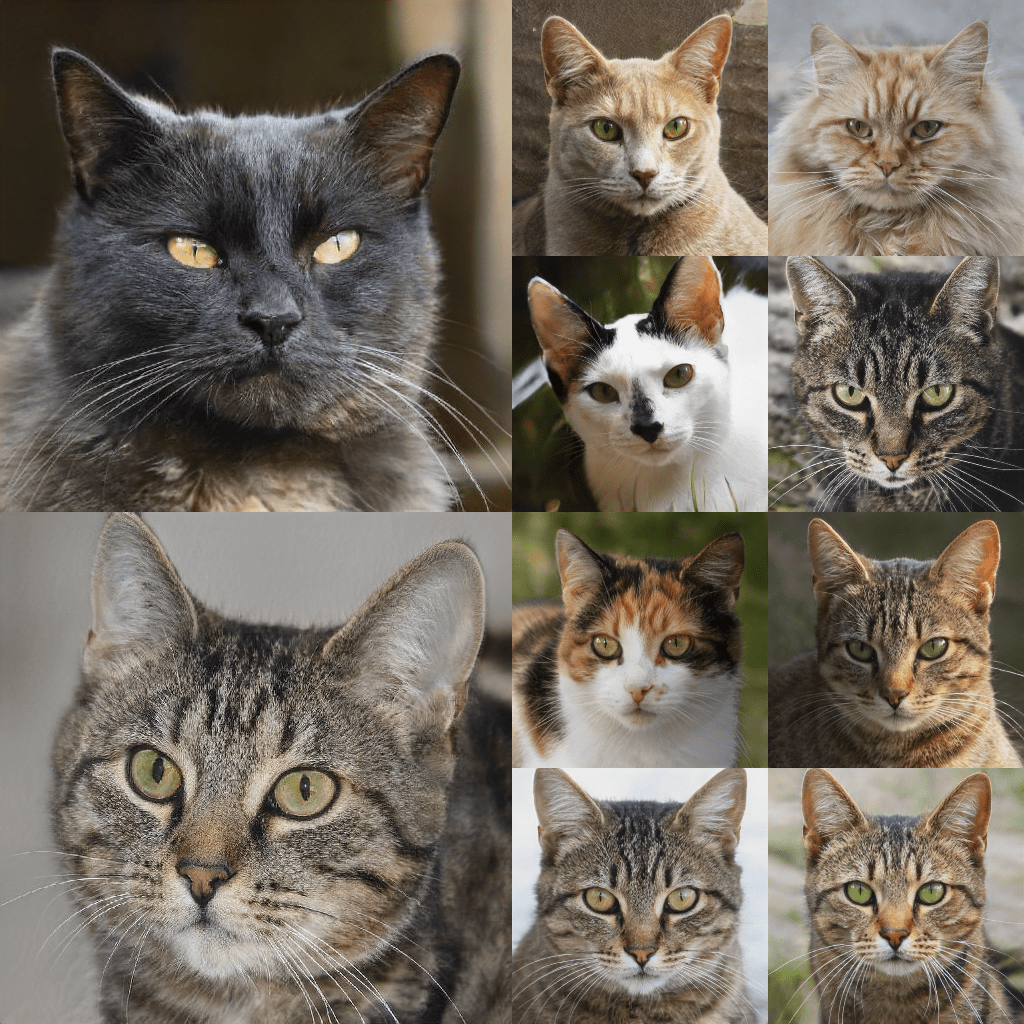}}}\hfill
\mpage{0.23}{\frame{\includegraphics[width=\linewidth, trim=0 0 0 0, clip]{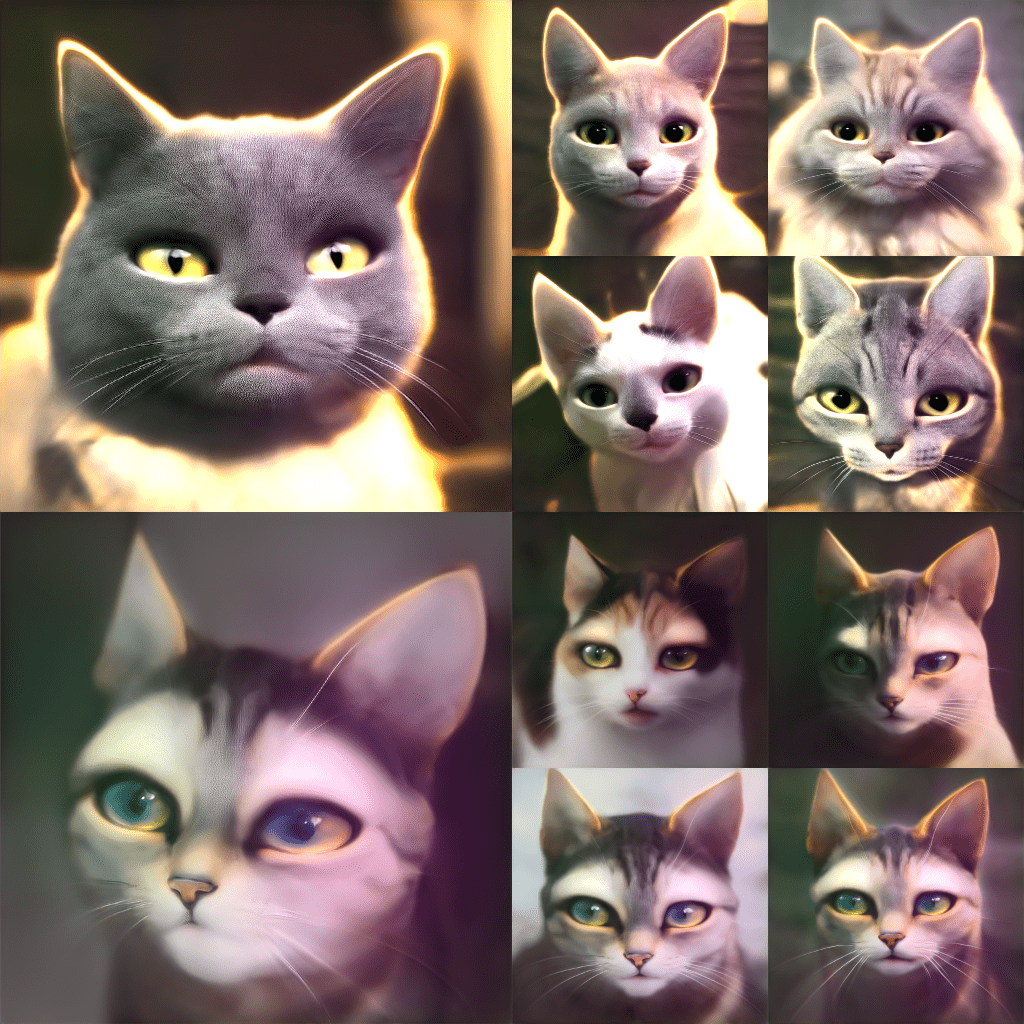}}}\\

\vspace{0.3mm}
\mpage{0.02}{\raisebox{-10pt}{\rotatebox{90}{}}}
\mpage{0.23}{\small{Source domain (FFHQ)}}\hfill
\mpage{0.23}{\small{Target domains }}\hfill
\mpage{0.23}{\small{Source domain (AFHQ)}}\hfill
\mpage{0.23}{\small{Target domains }}\\
\vspace{0.3mm}

\captionof{figure}{
\textbf{Applications of the proposed latent diffusion prior.} 
We showcase various visual synthesis applications from the proposed latent diffusion prior, including Text-to-3D, StyleGAN adaptation, and layered image editing.
}

\label{fig:intro}
\end{center}

}]

\maketitle

\thispagestyle{empty}
\begin{abstract}
There has been significant progress in using diffusion models for large-scale text-to-image synthesis. 
This has led to versatile downstream applications such as 3D object synthesis, image editing, and customized image generation.
In this paper, we present a generic approach that uses latent diffusion models as powerful image priors for various visual synthesis tasks.
Existing methods that use these priors do not fully exploit the models' capabilities.
To address this issue, we propose 
1) a feature matching loss that provides detailed guidance by comparing features from different decoder layers and 
2) a KL divergence loss that regularizes predicted latent features to stabilize the training process.
We apply our method to three applications: text-to-3D, StyleGAN adaptation, and layered image editing, and demonstrate its efficacy through extensive experimentation. 
Our results show that our method performs favorably compared to baseline methods.
\end{abstract}

\section{Introduction}
\label{sec:introduction}

Diffusion models have shown impressive image generation capabilities in terms of photorealism and compositionality~\cite{
balaji2022eDiff-I,GLIDE,rombach2022high,saharia2022photorealistic}. 
Through guidance control and embedding techniques, text-to-image diffusion models have been applied to various visual editing and processing tasks, 
\eg customized image generation~\cite{ruiz2022dreambooth}, video generation~\cite{wu2022tune}, image and video editing~\cite{kawar2022imagic,kim2022diffusion}, text-to-3D generation~\cite{lin2022magic3d, poole2022dreamfusion}, and image generator adaptation~\cite{song2022diffusion}. 
With a free-form text prompt, these methods utilize a pre-trained text-to-image diffusion model to guide the synthesis tasks. 
However, specialized methods are required for each task to produce good results.
While these methods may share similar objectives, a unified approach is underexplored.

In this paper, we leverage a pretrained diffusion model as a generic image prior for various visual synthesis applications. 
This is similar to the recent line of work that uses a CLIP model~\cite{clip}. 
CLIP models are trained to encode a paired image and its caption and maximize their cosine similarity. 
Several works have used a pretrained CLIP model as a prior to facilitate different text-guided synthesis tasks, \eg, image and video editing~\cite{bar2022text2live, patashnik2021styleclip, tzaban2022stitch,xu2022temporally}, generator adaptation~\cite{gal2022stylegannada}, text-to-3D synthesis~\cite{jain2021dreamfields, mohammad2022clip}.
However, as CLIP models are not generative models, their contrastive objective may not preserve the visual information useful for synthesis-oriented tasks.

Diffusion models have recently shown great potential in these synthesis tasks~\cite{kawar2022imagic,kim2022diffusion,lin2022magic3d,poole2022dreamfusion, song2022diffusion}.
They can achieve competitive or even better performance compared to CLIP-based approaches.
However, unlike CLIP models where the original CLIP objective was uniformly adopted, these works have applied diffusion models differently under different tasks. 
In this work, we present a framework to utilize diffusion models for various tasks.

We build our framework on Score Distillation Sampling~\cite{poole2022dreamfusion}, which uses an image diffusion model~\cite{saharia2022photorealistic} as prior to train a NeRF model without costly backpropagating through the diffusion model itself. 
Other works~\cite{lin2022magic3d,metzer2022latent,song2022diffusion,sjc} follow a similar approach but are based on latent diffusion models~\cite{rombach2022high}. 
The score distillation loss is computed in the latent space, which we term ``latent score distillation (LSD)''. 
Although the methods using score distillation with trained diffusion models have shown promising results, the loss is computed in a limited spatial resolution (\eg, $64\times64$) and thus cannot provide sufficient detailed guidance.

To achieve detailed guidance, we propose a new \emph{Feature Matching Loss} (FM) loss that  
uses features in multiple decoder layers of the latent diffusion model to guide the optimization process. 

Another limitation of recent works utilizing LSD is the lack of regularization over the optimized latent code. 
While minimizing the optimization loss in an unconstraint manner, these methods are likely to produce out-of-distribution latent code that is not seen by the decoder during the training, resulting in a lower-quality output. 
To mitigate this issue, we propose to use a KL divergence loss to regularize the optimized latent so that they stay close to the prior distribution during the training.

We evaluate our method on three text-driven synthesis applications,
text-to-3D generation, layered image editing, and StyleGAN adaptation (as shown in Fig.~\ref{fig:intro}).
Our approach achieves competitive results over baselines using the CLIP model as the image prior and diffusion models with latent score distillation.
 
We make the following contributions:
\begin{itemize}

\item
We propose a feature matching loss to extract detailed information from the decoder to guide text-based visual synthesis tasks.
\item
We propose a KL divergence loss to regularize the optimized latent, stabilizing the optimization process.
\item 
We extensively evaluate our method on three downstream tasks and show competitive results with strong baselines. 
\end{itemize}

\begin{figure*}[t]
\begin{center}
\centering
\includegraphics[trim=0 0 0 0, clip,width=.9\textwidth]{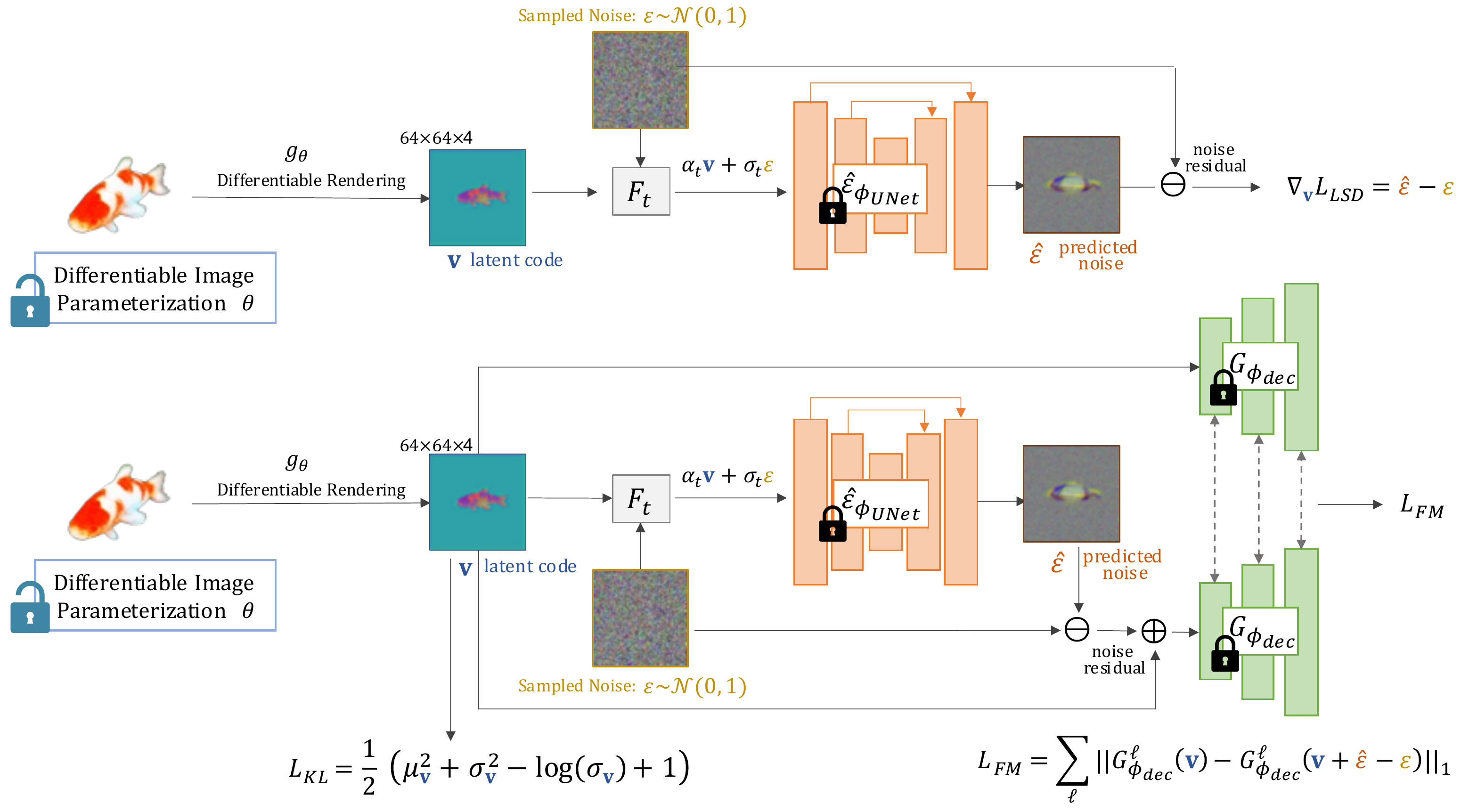}




\vspace{-1mm}

\caption{
\textbf{Method overview.} 
Our method aims to guide the generation and editing process based on a text prompt. 
We obtain the latent code from a differentiable renderer under different applications, including the generator from Text2LIVE~\cite{bar2022text2live}, StyleGAN-based generator, or a NeRF model.
This latent code $\mathbf{v}$ is perturbed following the latent diffusion model's scheduler at a random time step $t$, such that $F_t: \mathbf{z}_t = \alpha_t \mathbf{v}+ \sigma_t \mathbf{\epsilon}$.
This perturbed latent code $\mathbf{z}_t$ is then passed to the UNet to generate the predicted noise $\hat \epsilon$.
We then use the predicted noise $\hat \epsilon$ to derive the \emph{latent score distillation} gradient. 
To derive the \emph{feature matching} gradient, we input the latent code $v$ and noised latent code $\mathbf{v}+(\hat{\mathbf{\epsilon}} - \mathbf{\epsilon})$ into the decoder $G_{\phi_{dec}}(\cdot)$.
We compute the difference between the multi-level features from three different layers of the decoder to compute the feature matching loss.
Finally, both the \emph{latent score distillation} and \emph{feature matching} gradients are backpropagated to the differentiable renderer. 
}
\vspace{-5mm}
\label{fig:method_overview}
\end{center}
\end{figure*}

\section{Related Work}
\label{sec:related}

\topic{Text-to-image diffusion models.}
Diffusion models~\cite{ho2020denoising,kawar2022denoising,song2020denoising} synthesize images by denoising independent noises drawn from the standard Gaussian diffusion. 
Impressive progress has been made in photorealistic and zero-shot text-to-image generation~\cite{balaji2022eDiff-I,Make-A-Scene,GLIDE,DALLE, rombach2022high, saharia2022photorealistic, Parti}. 
When combined with tricks such as classifier-free guidance~\cite{Classifier} and gradient guidance~\cite{ma2020structure}, the flexibility of the diffusion model makes it readily adapt to different conditional generation tasks that are not part of its training objectives, such as image editing~\cite{kawar2022imagic,valevski2022unitune}, 
and personalized image generation~\cite{ruiz2022dreambooth}. 
Our work capitalizes on the recent progress text-to-image diffusion model and uses it as image priors for various visual synthesis tasks.

\topic{Text-driven 3D generative models.}
With the guidance from a Contrastive Language-Image Pre-training (CLIP) model~\cite{clip}, text-guided 3D generation becomes possible together with a differentiable renderer~\cite{ jain2021dreamfields, Jetchev2021ClipMatrixTC, mohammad2022clip, sanghi2021clip}. 
By optimizing the 3D representation through the CLIP objective computed on the rendered image, they can explicitly control the pose of generated 3D shapes, and generate creative appearances and shapes from the text freely. 
More recently, DreamFusion~\cite{poole2022dreamfusion} has brought diffusion models to this task and shown impressive results. 
In contrast to~\cite{graikos2022diffusion} backpropagating through the pre-trained diffusion model itself, DreamFusion leverages noise residual predicted by the pre-trained diffusion model as gradients for efficient backpropagation. 
Follow-up works~\cite{metzer2022latent,sjc} adapt the method to latent diffusion models~\cite{rombach2022high}. 
However, the 3D models resulting from these methods still lack details since the latent score distillation is computed in the latent space with a limited spatial resolution (\eg, $64\times64$ latent space for a $512\times 512$ image). 
In this work, we leverage the knowledge embedded in the \emph{latent diffusion decoder} to provide more detailed guidance. 
We show better details can be achieved in the synthesized 3D models.

\topic{Image generator domain adaptation.}
Some works aim to fine-tune a pre-trained generator with few-shot or text-guided zero-shot domain adaptation to reduce the cost of training an image generator.
Few-shot adaption aims to use fewer data (typically hundreds or fewer) to train an image generator.
Previous works control the learnable parameters~\cite{mo2020freeze,noguchi2019image,robb2020few,wang2018transferring},
design regularizers~\cite{li2020few,pinkney2020resolution,tseng2021regularizing}, 
or use auxiliary tasks~\cite{liu2020towards,yang2021data} to improve the quality of the generator.  
With the advancement of large fundamental models, some works use them as guidance to achieve zero-shot adaptation. 
StyleGAN-NADA~\cite{gal2022stylegannada} uses pre-trained CLIP model~\cite{clip} as guidance and shows diverse domain adaption results on StyleGAN~\cite{stylegan1}.
StyleGANFusion~\cite{song2022diffusion} leverages pre-trained StableDiffusion~\cite{rombach2022high} and follow the score distillation approach proposed by DreamFusion~\cite{poole2022dreamfusion} to achieve even better results.
Building upon StyleGANFusion~\cite{song2022diffusion}, we show that using our proposed method produces a significantly better FID score and a competitive CLIP score.

\topic{Text-driven image editing.}
Pioneering image manipulation methods~\cite{li2020manigan,nam2018text} utilize GANs to achieve editing of appearances while preserving the shape. 
However, the training image domains and text expression constraints often restrict the GAN-based approaches. 
The advanced methods have leveraged embeddings from a pretrained CLIP~\cite{clip}. 
These embeddings can be applied to update the generative models with test-time optimization~\cite{bar2022text2live,gal2022stylegannada,kwon2022clipstyler,patashnik2021styleclip,xu2022predict}. 
Recently, text-to-image diffusion models have shown exceptional success in manipulation tasks~\cite{balaji2022eDiff-I,brooks2022instructpix2pix,kawar2022imagic,mokady2022null,GLIDE,ruiz2022dreambooth}. 
We demonstrate the application of our diffusion prior to the image editing task. 
Unlike existing diffusion-based editors, our method manipulates images using test-time optimization with the proposed diffusion guidance. 
Our method produces more detailed results than the latent diffusion-guided baselines and the CLIP-guided method, Text2LIVE~\cite{bar2022text2live}.

\def\D{\altmathcal{D}}
\def\I{\altmathcal{I}}
\def\O{\altmathcal{O}}
\def\res{\altmathcal{R}}

\def\b{\mathbfit{b}}
\def\c{\mathbfit{c}}
\def\d{\mathbfit{d}}
\def\o{\mathbfit{o}}
\def\p{\mathbfit{p}}
\def\t{\mathbfit{t}}
\def\x{\mathbfit{x}}
\def\z{\mathbfit{z}}

\def\K{\mathbfit{K}}
\def\R{\mathbfit{R}}

\def\ang{\phi}
\def\dehom{\mu}
\def\proj{\pi}
\def\sigmoid{S}
\def\vis{\nu}
\def\r{\mathbfit{r}}

\def\bp{(\p\!)} 
\def\bt{(t\!)} 
\def\bx{(\x\neg)} 

\def\ok{\o_{\neg k}}
\def\tk{\t_{\neg k}}
\def\wk{w_{\neg k}}
\def\xi{\x_{\neg i}}
\def\zk{\z_{\neg k}}
\def\Kk{\K_{\neg k}}
\def\Rk{\R_{\neg k}}

\def\ng{\hspace{-0.1mm}}
\def\neg{\hspace{-0.2mm}}
\def\pos{\hspace{0.2mm}}

\makeatletter
\newcommand*\MY@rightharpoonupfill@{%
    \arrowfill@\relbar\relbar\rightharpoonup
}
\newcommand*\overrightharpoon{%
    \mathpalette{\overarrow@\MY@rightharpoonupfill@}%
}
\makeatother

\newlength{\depthofsumsign}
\setlength{\depthofsumsign}{\depthof{$\sum$}}
\newcommand{\nsum}[1][1.4]{
    \mathop{%
        \raisebox
            {-#1\depthofsumsign+1\depthofsumsign}
            {\scalebox
                {#1}
                {$\displaystyle\sum$}%
            }
    }
}
\section{Proposed Method}
\label{sec:method}

\subsection{Background}
\label{sec:background}

\topic{Score distillation.}
DreamFusion~\cite{poole2022dreamfusion} first proposes using the Imagen~\cite{saharia2022photorealistic} diffusion model for a text-driven 3D generation. 
Imagen can generate \emph{high-resolution} images by cascade diffusion.
However, the base model only operates on a \emph{low-resolution} $64 \times 64$ image space.
They build upon the formulation introduced by~\cite{graikos2022diffusion} and propose a more stable training process.
Their approach optimizes the underlying 3D representation model by using the training objective of the diffusion model.
The resulting approach, score distillation sampling, involves perturbing the rendered image with a random noise $\mathbf{\epsilon}$, and then using the pretrained diffusion model to predict the noise $\hat{\mathbf{\epsilon}}$. 
The training is based on the gradient computed from the \emph{noise residual} between the added random noise and the predicted one: $\hat{\mathbf{\epsilon}} - \mathbf{\epsilon}$. 
Note that the Imagen model~\cite{saharia2022photorealistic} only operates on low-resolution image space.

\topic{Latent score distillation.}
Jacobian NeRF~\cite{sjc}, Latent NeRF~\cite{metzer2022latent} and StyleGANFusion~\cite{song2022diffusion} have recently incorporated score distillation into the latent diffusion models~
\cite{rombach2022high}. A latent diffusion model generally contains two components, an autoencoder with encoder $E_{\phi_{enc}}$ and decoder $G_{\phi_{dec}}$, and a diffusion model $\hat{\epsilon}_{\phi_{UNet}}$ operating in the latent space. 
However, the decoder is not utilized in the latent score distillation of~\cite{metzer2022latent,song2022diffusion,sjc}, which can lead to inferior results since the knowledge of converting low-resolution latent to high-resolution RGB images is not used.
To address this, we propose using feature matching and KL losses to reintroduce the decoder into the optimization procedure.

\subsection{Feature matching Loss}
\label{sec:feature_matching}

We propose a \emph{feature matching} loss to leverage the generative capacity of the decoder $G_{\phi_{dec}}$ and provide finer-grained guidance to the differentiable renderer. 
We use the decoder of the stable diffusion autoencoder as shown in Figure~\ref{fig:method_overview}. 
We feed the original latent code $\mathbf{v}$ and the updated latent code $\mathbf{v}' = \mathbf{v}+(\hat{\mathbf{\epsilon}} - \mathbf{\epsilon})$ to the decoder $G_{\phi_{dec}}(\cdot)$, and then compute the \emph{feature matching} loss: 
$$L_{FM} = \sum_\ell \|G^{\ell}_{\phi_{dec}}\left(\mathbf{v}\right) - G^{\ell}_{\phi_{dec}}\left(\mathbf{v}+\hat{\mathbf{\epsilon}} - \mathbf{\epsilon}\right)\|_1,$$ 
where $\ell$ is the specific level in the decoder $G_{\phi_{dec}}$.

Our proposed feature matching loss, denoted as $L_{FM}$, is inspired by the GAN discriminator feature matching loss proposed in pix2pixHD~\cite{wang2018pix2pixHD}, which aims to align the features of multiple discriminator layers of real and synthetic images.
Our feature matching loss $L_{FM}$ operates similarly to the latent score distillation loss $L_{LSD}$, but its signal is \emph{amplified} through the use of the decoder $G_{\phi_{dec}}$.
Our approach differs from the feature matching loss in pix2pixHD in two major aspects.
First, we measure the similarity of the extracted features from the pretrained (and fixed) \emph{decoder}, not from an additional trainable discriminator.
Second, we use latent code \emph{with added noise residual} and the clean latent as the decoder input, as opposed to real/fake images.

We denote $\mathbf{\theta}$ as the target parameters in the renderer $g_{\mathbf{\theta}}(\cdot)$ to optimize, \eg NeRF model for text-to-3D generation and the StyleGAN model for generator adaptation. 
The direct gradient computation involves calculating the UNet Jacobian, which is computationally expensive~\cite{poole2022dreamfusion}. 
We therefore consider the gradient backpropagation to the generator $\frac{d L_{FM}}{d \mathbf{\theta}}$ using the chain rule: 
\begin{equation}
\begin{aligned}
    \frac{d L_{FM}}{d \mathbf{\theta}}
&=
\frac{\partial \mathbf{v}}{\partial \mathbf{\theta}}
\frac{\partial L_{FM}}{\partial \mathbf{v}} \\
&=
\frac{\partial \mathbf{v}}{\partial \mathbf{\theta}}
\left(
\alpha_t  \frac{\partial \hat{\mathbf{\epsilon}}}{\partial \mathbf{z}_t}\frac{\partial \mathbf{x}'}{\partial \mathbf{v}'}\frac{\partial L_{FM}}{\partial \mathbf{x}'} +
\frac{\partial \mathbf{x}'}{\partial \mathbf{v}'}\frac{\partial L_{FM}}{\partial \mathbf{x}'}+
\frac{\partial \mathbf{x}}{\partial \mathbf{v}}\frac{\partial L_{FM}}{\partial \mathbf{x}}
 \right), \\
\end{aligned}
\end{equation}
where 
$\frac{\partial \hat{\mathbf{\epsilon}}}{\partial \mathbf{z}_t}$ 
refers to the UNet Jacobian term, which we ignore due to the high optimization cost. 
$\mathbf{x}$ and $\mathbf{x}'$ denote the latent feature within the decoder obtained by the original $\mathbf{v}$ and updated latent code $\mathbf{v}'$. $\mathbf{z}_t$ refers to the perturbed latent code defined as $\mathbf{z}_t = \alpha_t \mathbf{v}+ \sigma_t$, where $\alpha_t$ and $\sigma_t$ are time-dependent constant defined in DDPM~\cite{ho2020denoising}.

The final gradient can be derived as follows:
\begin{equation}
\begin{aligned}
\frac{d L_{FM}}{d \mathbf{\theta}}
&= 
\frac{\partial \mathbf{v}}{\partial \mathbf{\theta}}
\left(
\left(1+ \alpha_t\right) \frac{\partial \mathbf{x}'}{\partial \mathbf{v}'}
\frac{\partial L_{FM}}{\partial \mathbf{x}'} + \frac{\partial \mathbf{x}}{\partial \mathbf{v}}\frac{\partial L_{FM}}{\partial \mathbf{x}} \right).\\
\end{aligned}
\end{equation}

\begin{figure*}[t]
\begin{center}
\centering
\vspace{1mm}

\mpage{0.22}{{\includegraphics[width=\linewidth, trim=0 0 0 0, clip]{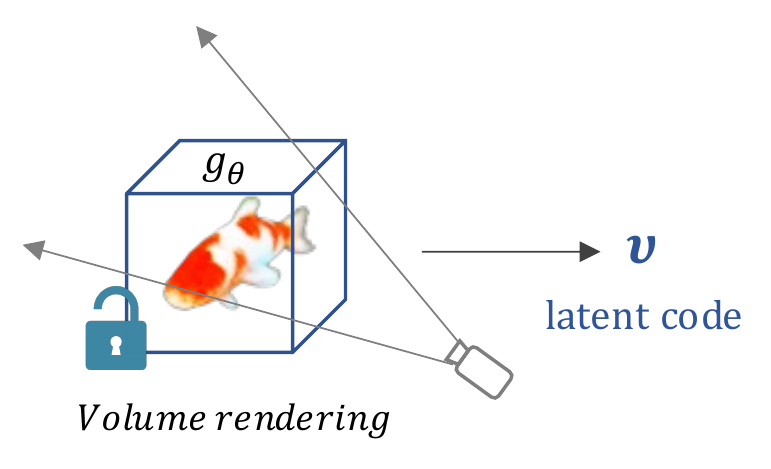}}}\hfill
\mpage{0.30}{{\includegraphics[width=\linewidth, trim=0 0 0 0, clip]{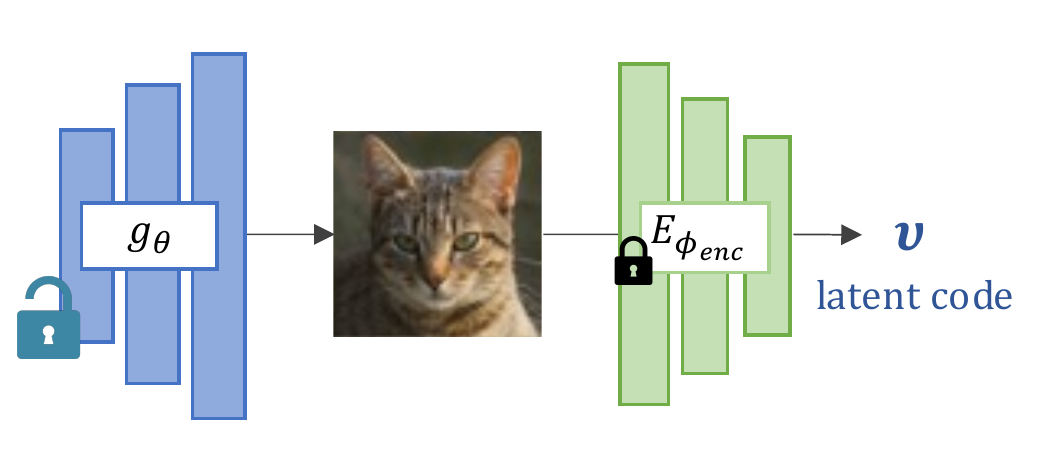}}}\hfill
\mpage{0.45}{{\includegraphics[width=\linewidth, trim=0 0 0 0, clip]{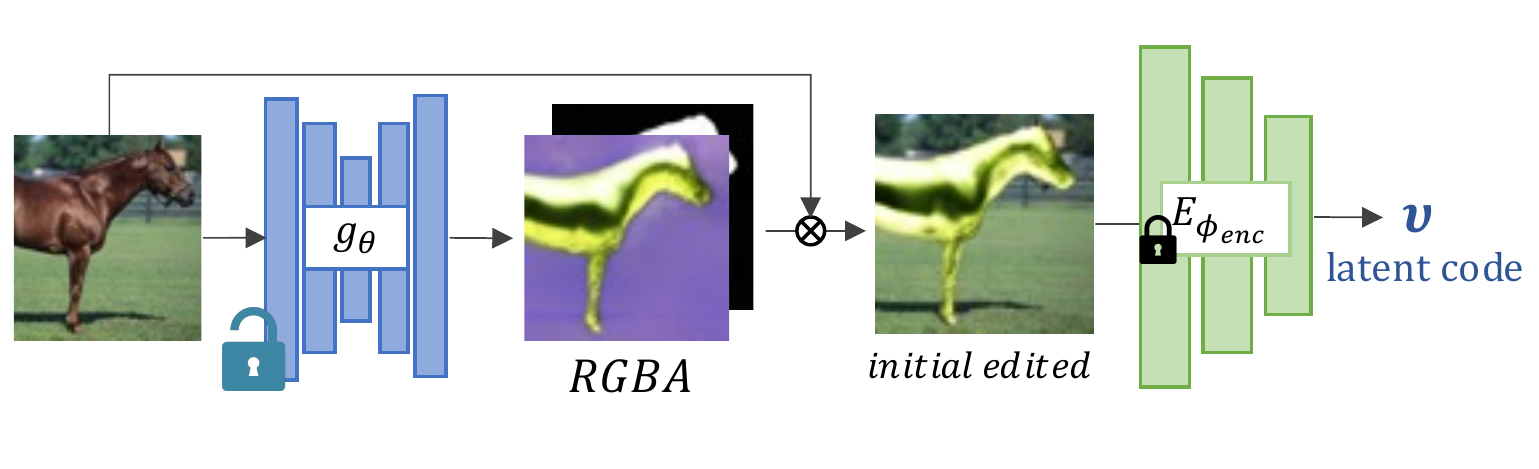}}}\hfill

\mpage{0.22}{{{\small{(a) Text-to-3D}}}}\hfill
\mpage{0.30}{{{\small{(b) StyleGAN adaptation}}}}\hfill
\mpage{0.45}{{{\small{(c) Layered image editing}}}}\\

\caption{
\textbf{Applications pipeline.} 
To apply our proposed feature matching loss $L_{FM}$,  KL regularizer loss $L_{KL}$ and score distillation loss $L_{LSD}$, we first obtain the latent code $\mathbf{v}$ using the differentiable renderer in each application. As illustrated in the figure, to obtain the image that produces $\mathbf{v}$ with StableDiffusion encoder $E_{\phi_{enc}}$, in (a) Text-to-3D, we render from a NeRF model with a random camera viewpoint; (b) StyleGAN adaptation, we generate the image with a pretrained StyleGAN model; (c) Layered image editing application, we use the generator of Text2LIVE to synthesize the edited image, alpha map, and the alpha blending of the initial and edited images.
}
\label{fig:app_overview}
\end{center}
\end{figure*}

\subsection{Kullback-Leibler divergence regularizer.}
\label{sec:KL_loss}
When conducting optimization through the latent code in an unconstrained way~\cite{metzer2022latent,poole2022dreamfusion,sjc}, the resulting latent code likely drifts away from the original distribution. 
Consequently, the decoder must handle an \emph{unseen input} during the training and produces poor quality. 
Our feature matching loss mitigates this issue by incorporating the gradient from the decoder. 
However, it is insufficient in practice as we still observe artifacts or unrealistic styles in the decoded image. 
Therefore, we propose further regularizing the latent with a KL divergence loss.

Stable Diffusion and VAE models both use a KL penalty to regularize the variance of the latent space. However, in the text-to-3D task, a latent radiance field is built to directly predict the latent code without an encoder $E_{\phi_{enc}}$. Thus, different from the training process, we do not compute the KL penalty on the mean and variance output by the encoder but directly on the latent sample $v$ as below:

\begin{equation}
L_{KL} = \frac{1}{2} \left(
\mu_{\mathbf{v}}^2 + \sigma^2_{\mathbf{v}} - log\left(\sigma^2_{\mathbf{v}}\right) + 1
\right),
\end{equation}
where $\mu_{\mathbf{v}} = \frac{1}{N}\sum_i \mathbf{v}_i $ and $\sigma^2_{\mathbf{v}} = \frac{1}{N}\sum_i(\mathbf{v}_i-\mu_{\mathbf{v}})^{2} $ represent the mean and variance of the latent code $\mathbf{v}$, respectively.
And $N$ denotes the number of elements of the latent code $\mathbf{v}$.
We find this to be effective in improving the stability of the optimization process. 

\subsection{Training procedure}
\label{sec:training_procedure}
The diffusion prior includes three parts, \emph{latent score distillation}, \emph{feature matching loss}, and \emph{KL regularizer}.

During the optimization, the latent code $\mathbf{v}$ is first perturbed following the DDPM~\cite{ho2020denoising} scheduler at a random time step $t$, such that perturbed latent code $\mathbf{z}_t = \alpha_t \mathbf{v}+ \sigma_t \mathbf{\epsilon}$. 
This perturbed latent code $\mathbf{z}_t$ is then passed to the UNet to generate the predicted noise $\hat{\mathbf{\epsilon}}$.
We then use the predicted noise $\hat{\mathbf{\epsilon}}$ to derive the \emph{latent score distillation} gradient. We define the latent score distillation loss as $L_{LSD}$ in this paper.
To compute the feature matching loss, we input the latent code $\mathbf{v}$ and updated latent code $\mathbf{v}+(\hat{\mathbf{\epsilon}} - \mathbf{\epsilon})$ into the decoder $G_{\phi_{dec}}(\cdot)$.
We use the decoded features at three different layers
from the decoder, to compute the feature matching loss.

We compute the gradient partial to the latent code $\mathbf{v}$, and propagate back to optimize $g_\mathbf{\theta}$ using the final loss $L_{final}$:
\begin{equation}
L_{final} = \lambda_{1} L_{FM} + \lambda_{2} L_{KL} + \lambda_{3} L_{LSD},
\end{equation}
where $\lambda_{1}, \lambda_{2}, \lambda_{3}$ are the balancing factor for each loss.
\section{Experiments}
\label{sec:exp}

\label{sec:setup}
We evaluated three applications using Stable Diffusion~\cite{rombach2022high} v1.4 for Text-to-3D (Latent NeRF) and StyleGAN adaptation, and Stable Diffusion v1.5 for Text-to-3D (Jacobian NeRF) and layered image editing as our pretrained diffusion model.
Fig.~\ref{fig:app_overview} depicts the overall pipeline for each application and demonstrates how we acquire the latent code $\mathbf{v}$ from each differentiable generator used in the three applications. We include the implementation details and more results in the supplementary material and we will make the source code publicly available.

\subsection{Applications}
\begin{figure*}[t]
\centering

\mpage{0.01}{\raisebox{0pt}{\rotatebox{90}{\small{burger\footnotemark}}}}  \hfill
\mpage{0.155}{\frame{\includegraphics[width=\linewidth, trim=3cm 3cm 3cm 3cm, clip]{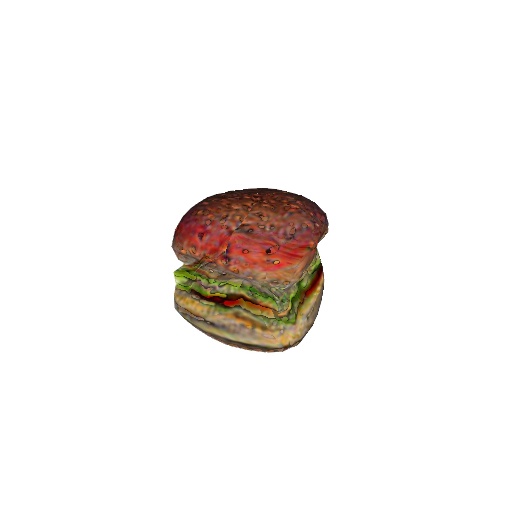}}}\hfill
\mpage{0.155}{\frame{\includegraphics[width=\linewidth, trim=2.5cm 2.5cm 2.5cm 2.5cm, clip]{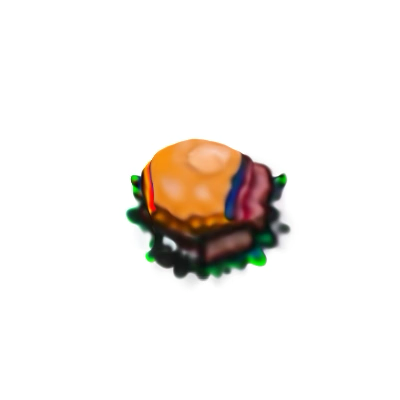}}}\hfill
\mpage{0.155}{\frame{\includegraphics[width=\linewidth, trim=0 0 0 0, clip]{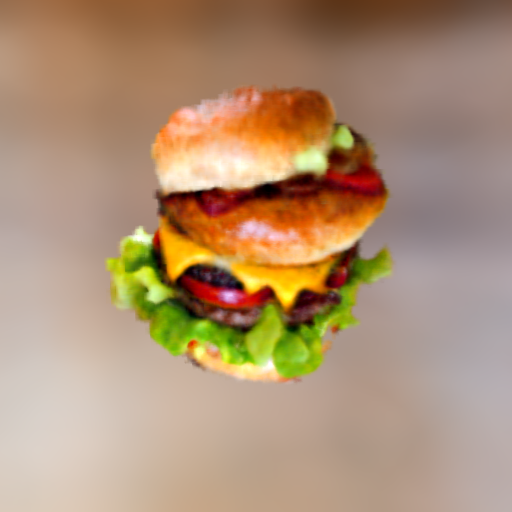}}}\hfill
\mpage{0.155}{\frame{\includegraphics[width=\linewidth, trim=0 0 0 0, clip]{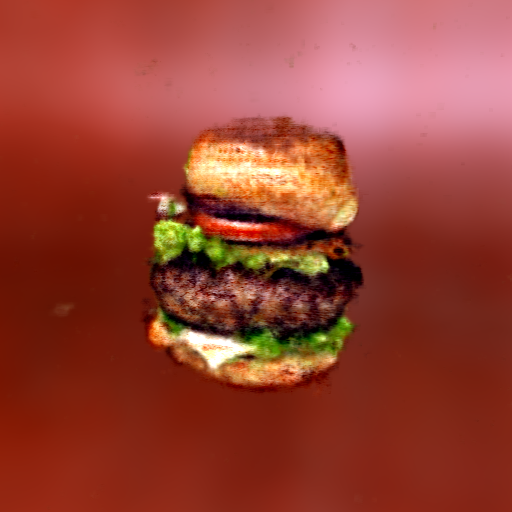}}}\hfill
\mpage{0.155}{\frame{\includegraphics[width=\linewidth, trim=0 0 0 0, clip]{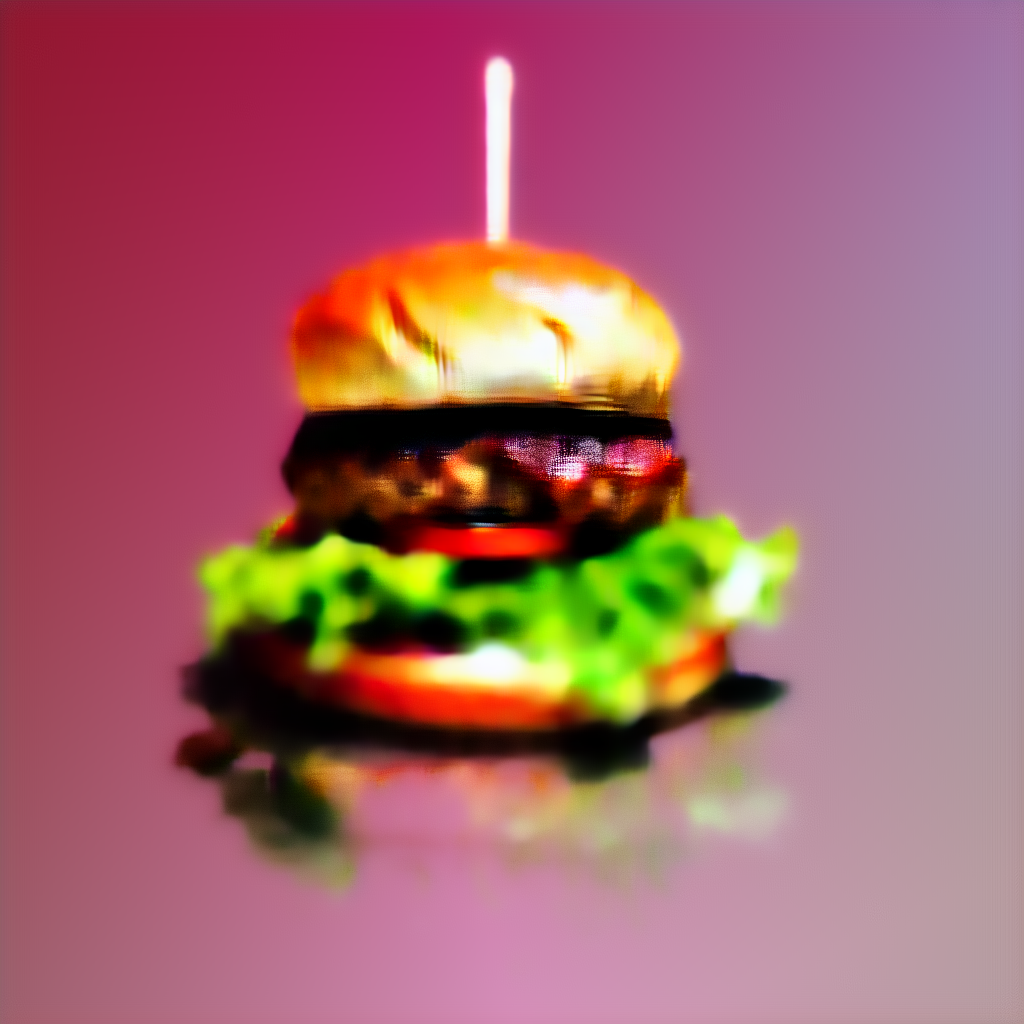}}}\hfill
\mpage{0.155}{\frame{\includegraphics[width=\linewidth, trim=0 0 0 0, clip]{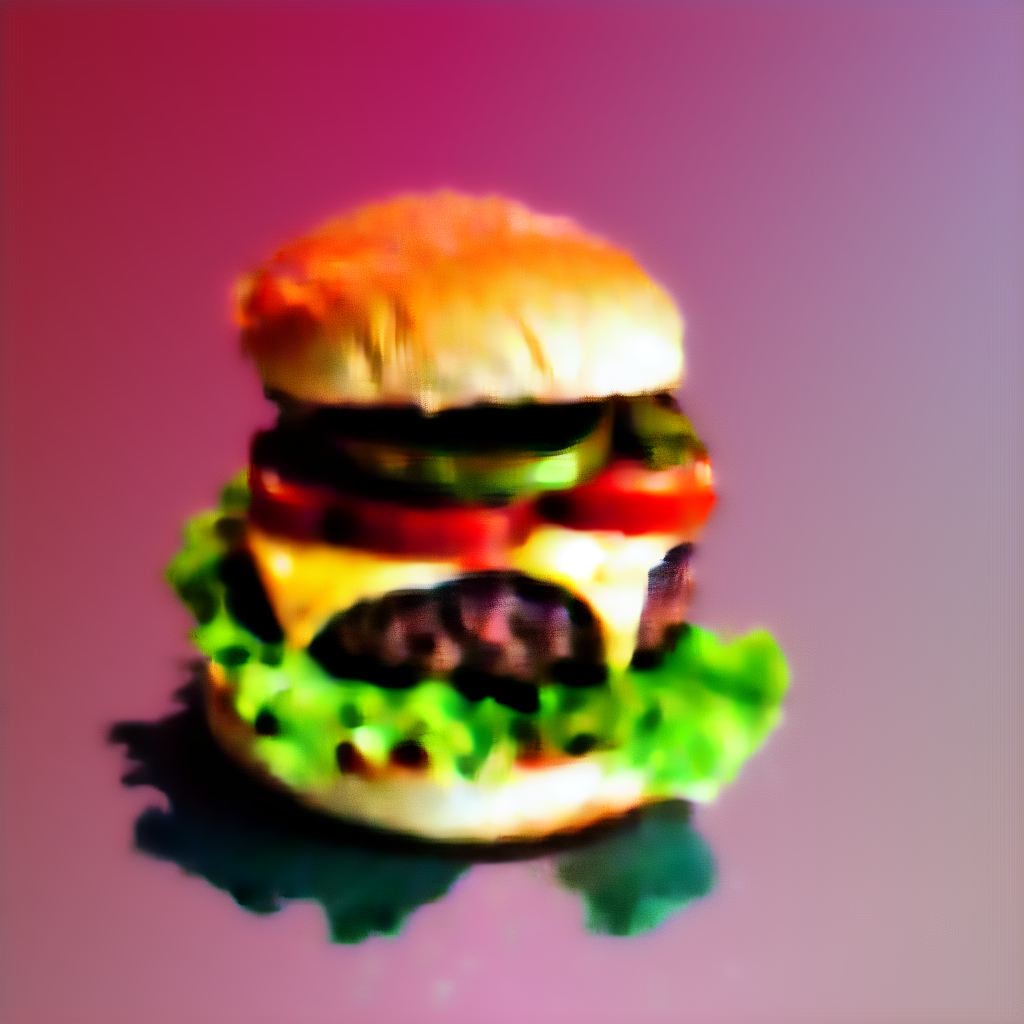}}}\\

\vspace{1mm}

\mpage{0.01}{\raisebox{0pt}{\rotatebox{90}{\small{koi fish}}}}  \hfill
\mpage{0.155}{\frame{\includegraphics[width=\linewidth, trim=2cm 2cm 2cm 2cm, clip]{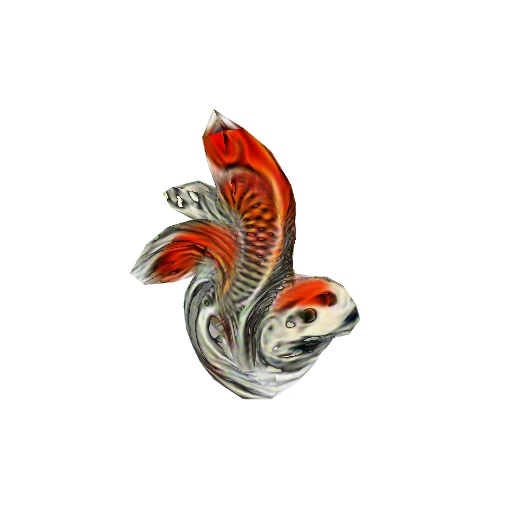}}}\hfill
\mpage{0.155}{\frame{\includegraphics[width=\linewidth, trim=0 0 0 0, clip]{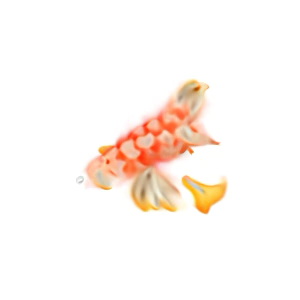}}}\hfill
\mpage{0.155}{\frame{\includegraphics[width=\linewidth, trim=0 0 0 0, clip]{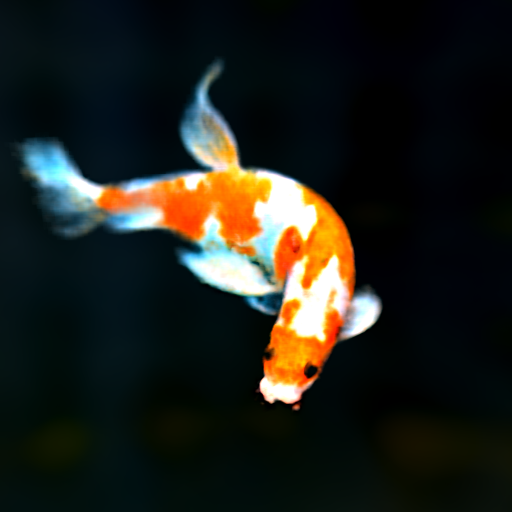}}}\hfill
\mpage{0.155}{\frame{\includegraphics[width=\linewidth, trim=0 0 0 0, clip]{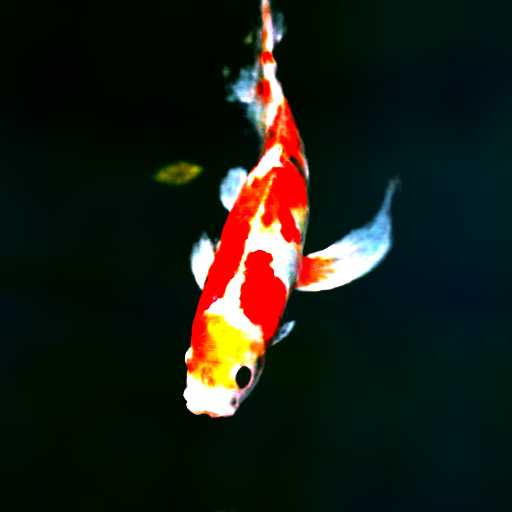}}}\hfill
\mpage{0.155}{\frame{\includegraphics[width=\linewidth, trim=0 0 0 0, clip]{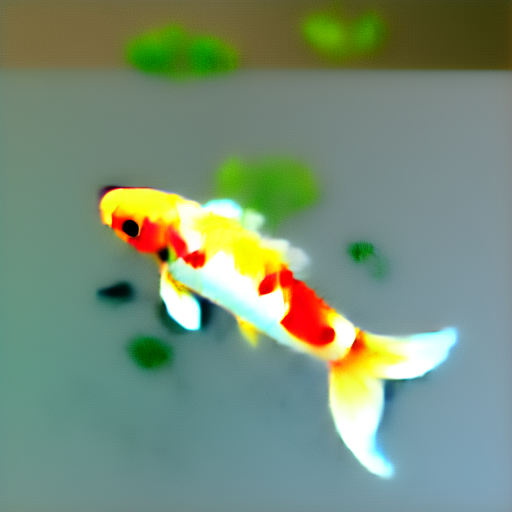}}}\hfill
\mpage{0.155}{\frame{\includegraphics[width=\linewidth, trim=0 0 0 0, clip]{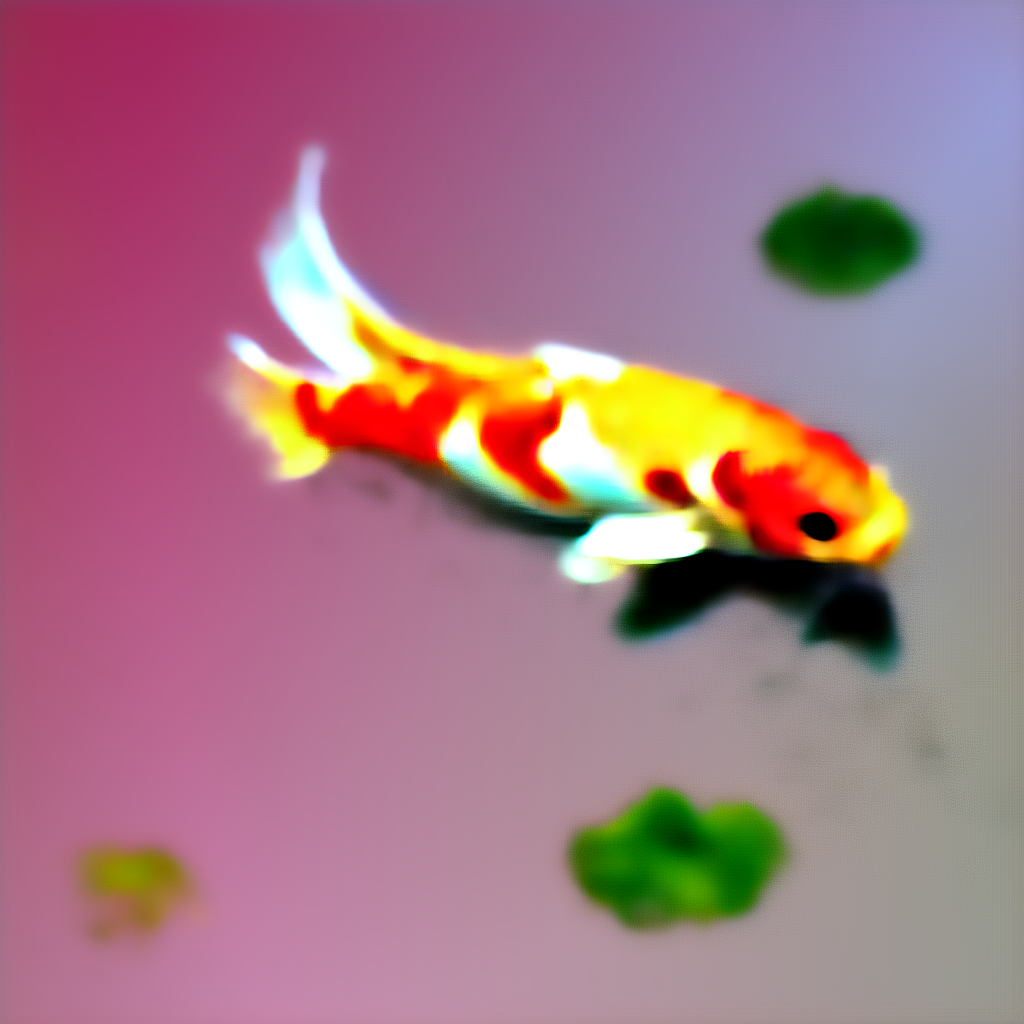}}}\\

\vspace{1mm}

\mpage{0.01}{\raisebox{0pt}{\rotatebox{90}{\small{blue jug\footnotemark}}}}  \hfill
\mpage{0.155}{\frame{\includegraphics[width=\linewidth, trim=1.5cm 1cm 1.5cm 2.0cm, clip]{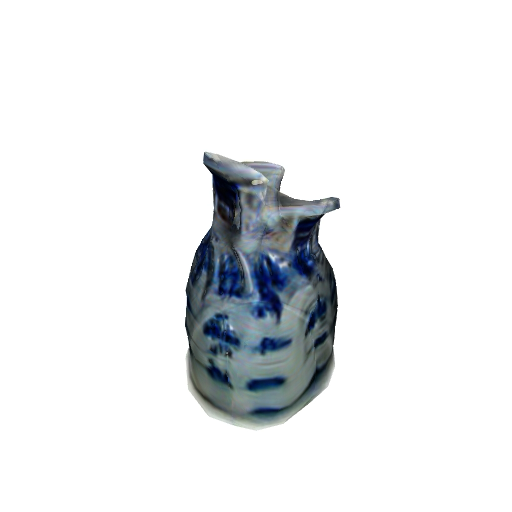}}}\hfill
\mpage{0.155}{\frame{\includegraphics[width=\linewidth, trim=1.5cm 1.5cm 1.5cm 1.5cm, clip]{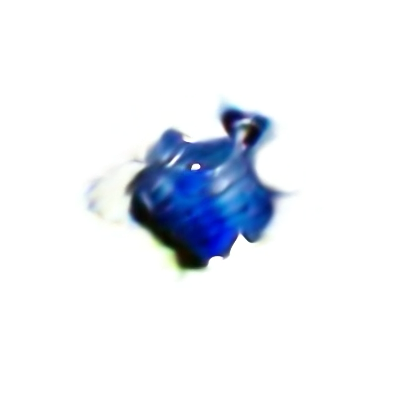}}}\hfill
\mpage{0.155}{\frame{\includegraphics[width=\linewidth, trim=0 0 0 0, clip]{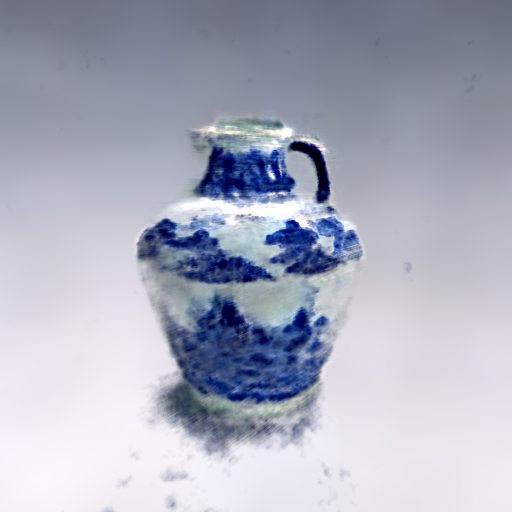}}}\hfill
\mpage{0.155}{\frame{\includegraphics[width=\linewidth, trim=1.5cm 2.0cm 1.5cm 1.0cm, clip]{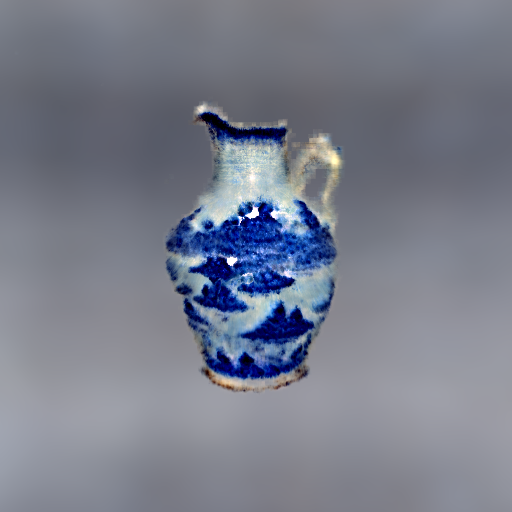}}}\hfill
\mpage{0.155}{\frame{\includegraphics[width=\linewidth, trim=0 0 0 0, clip]{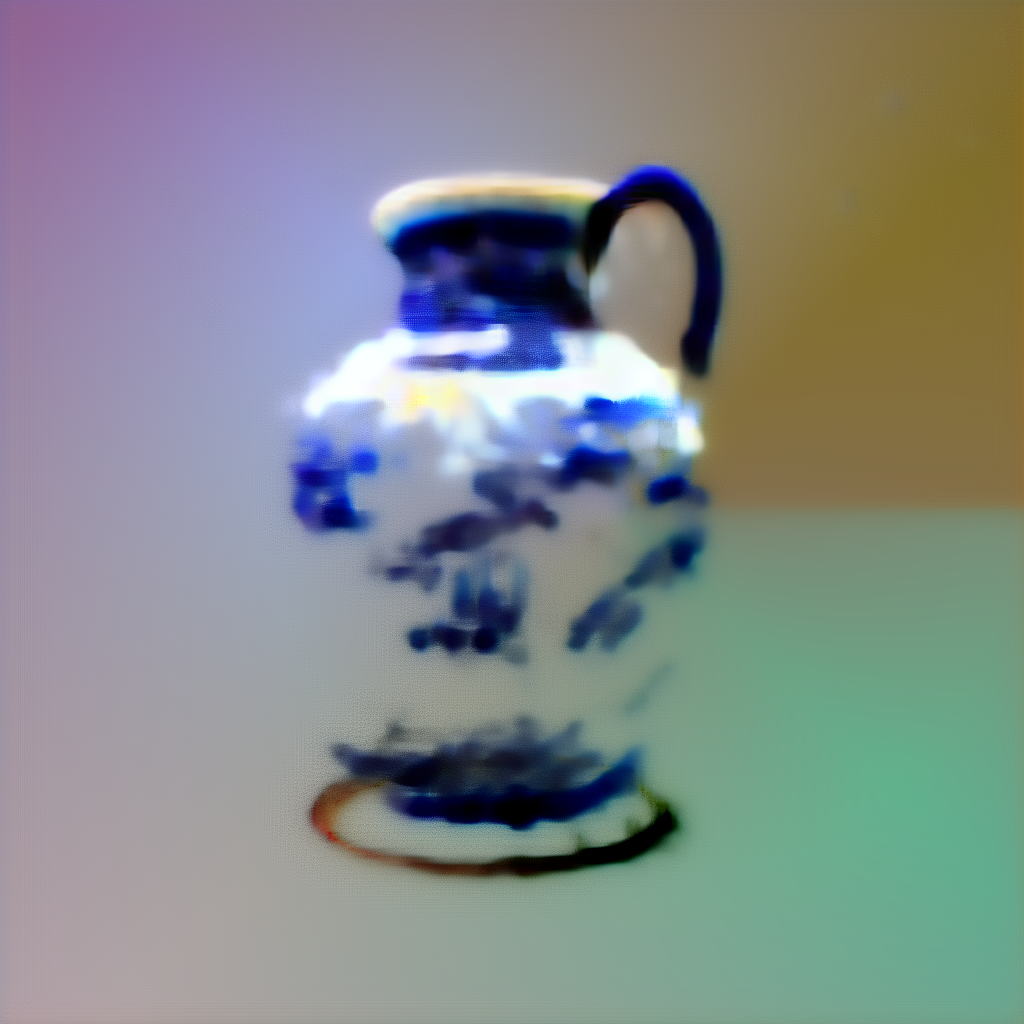}}}\hfill
\mpage{0.155}{\frame{\includegraphics[width=\linewidth, trim=0.5cm 0 0.5cm 1cm, clip]{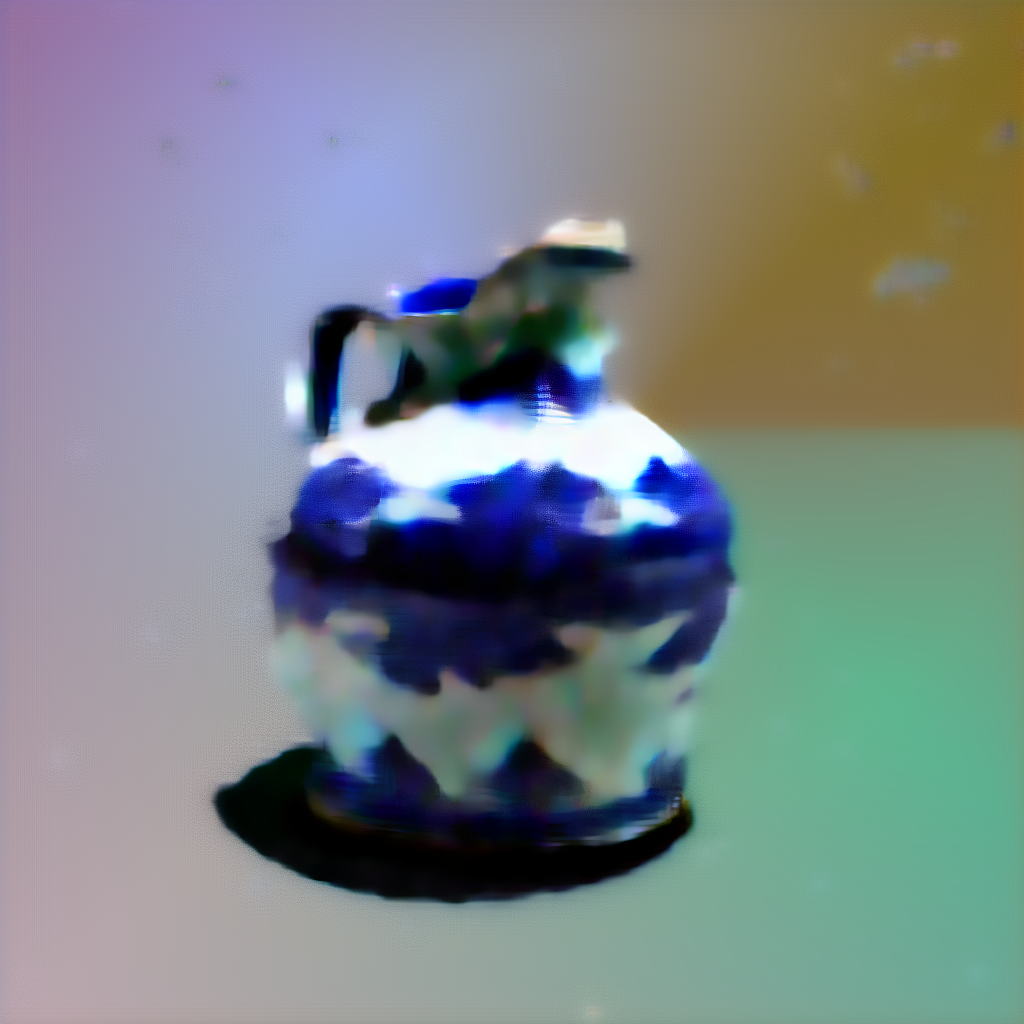}}}\\

\vspace{1mm}

\mpage{0.01}{{{\small{\!}}}}\hfill
\mpage{0.155}{{{\small{CLIP-Mesh~\cite{mohammad2022clip}}}}}\hfill
\mpage{0.155}{{{\small{DreamField~\cite{jain2021dreamfields}}}}}\hfill
\mpage{0.155}{{{\small{Latent NeRF~\cite{metzer2022latent}}}}}\hfill
\mpage{0.155}{{{\small{{Ours (Latent NeRF)}}}}}\hfill
\mpage{0.155}{{{\small{Jacobian NeRF~\cite{sjc}}}}}\hfill
\mpage{0.155}{{{\small{{Ours (Jacobian NeRF)}}}}}\\

\caption{\textbf{Comparisons on Text-to-3D.} We compare our method with CLIP-Mesh~\cite{mohammad2022clip}, DreamField~\cite{jain2021dreamfields}, Latent NeRF~\cite{metzer2022latent} and Jacobian NeRF~\cite{sjc}. Our approach can provide more details and a better appearance. For instance, the hamburger appears with clearer layers, while the fish and vase possess a more solid structure. We show the complete text prompts in the footnote.
}
\vspace{-3mm}
\label{fig:comparison_3D}
\end{figure*}

\topic{Text-to-3D.}
The text-to-3D task aims to generate a 3D model from a free-form text description. 
We evaluate our method on two text-guided 3D generative models, Jacobian NeRF ~\cite{sjc}, and Latent NeRF ~\cite{metzer2022latent} using the Stable Diffusion as guidance.
Both methods learn and predict the latent code within each viewing point and optimize with the latent score distillation.
We applied the proposed feature matching loss $L_{FM}$ and KL regularizer $L_{KL}$ to both methods. 
We also compare our results with two other baselines, CLIP-Mesh~\cite{mohammad2022clip} and DreamFields~\cite{jain2021dreamfields}, which leverage CLIP~\cite{clip} as guidance instead. 
As shown in Fig.~\ref{fig:comparison_3D}, 
Ours (Latent NeRF) and Ours (Jacobian NeRF) using our proposed losses upon Latent NeRF and Jacobian NeRF offer better quality with more details. 

We further evaluate the generated 3D model with the CLIP-R precision score~\cite{cliprprecision}, comparing with DreamFields~\cite{jain2021dreamfields}, CLIP-Mesh~\cite{mohammad2022clip}, Latent NeRF~\cite{metzer2022latent} and Jacodian NeRF~\cite{sjc}. 
We follow the same experiment setup outline in CLIP-Mesh~\cite{mohammad2022clip}, which involved generating one 3D model per prompt in 153 text prompts. 
We take a rendered image from a random camera pose during the evaluation to compute the CLIP score with 99 random prompts and the generated text prompt. 
We test two different-sized CLIP models for computing the precision. 
As shown in Table~\ref{tab:quan_RPrecision}, 
our method consistently improves over both baseline methods~\cite{metzer2022latent, sjc}.
Note that both DreamFields and CLIP-Mesh~\cite{jain2021dreamfields, mohammad2022clip} \emph{optimize directly with CLIP loss}, which can lead to potential overfitting issues as discussed in ~\cite{jain2021dreamfields}.
Our focus is \emph{not} on achieving state-of-the-art text-to-3D results but showing how our proposed losses complement the commonly used latent score distillation. 
The SOTA methods \cite{lin2022magic3d,poole2022dreamfusion} rely on proprietary \emph{image-based} diffusion models (Imagen~\cite{saharia2022photorealistic} and e-Diff~\cite{balaji2022eDiff-I}) that are not publicly available.
Magic3D~\cite{lin2022magic3d}, however, does use a latent diffusion model in their second stage of training. 
We believe that our proposed method can also be applied to improve the results.


\begin{table}[htbp]
    \centering
    \footnotesize
    \caption{R-Precision: the coherence of our model generations with their caption using different CLIP retrieval models.}
    \vspace{1mm}
        \begin{tabular}{l|c|c}
        \toprule
            Method & CLIPB/32 & CLIPB/16 \\
        \midrule
            GT Images & 77.1 & 79.1 \\
        \midrule
            DreamFields~\cite{jain2021dreamfields} & 68.3 & 74.2  \\
            CLIP-Mesh~\cite{mohammad2022clip} & 67.8 & 75.8  \\
        \midrule
            Latent NeRF~\cite{metzer2022latent} & 29.8 $\pm$ 1.54 & 37.7 $\pm$ 2.74 \\
            Ours (Latent NeRF)& 35.1 $\pm$ 1.63 & 39.4 $\pm$ 3.03 \\
            Jacobian NeRF~\cite{sjc} & 31.2 $\pm$ 0.82  & 44.0 $\pm$ 2.41 \\
            Ours (Jacobian NeRF)  & 33.3 $\pm$ 0.53 & 46.0 $\pm$ 1.34 \\
        \toprule
        \end{tabular}
    \label{tab:quan_RPrecision}
\end{table}


\begin{table}[htbp]
  \centering
  \small
  \caption{We compare our method with two baselines, StyleGANFusion~\cite{song2022diffusion} and StyleGAN-NADA~\cite{gal2022stylegannada}, in Cat-to-Animals and report the average FID scores (the lower the better).
  Our method outperforms other baselines in most of the experiments.}
  \resizebox{\columnwidth}{!}{%
    \begin{tabular}{lrrr}
    \toprule
          & \multicolumn{1}{l}{StyleGANFusion} & \multicolumn{1}{l}{StyleGAN-NADA} & \multicolumn{1}{l}{Ours} \\
    \midrule
    \emojidog~Dog  &  133.68   &     228.59   &       \textbf{132.35}   \\
    \emojifox~Fox   &      \textbf{47.05}  &   144.89   &     {52.73}    \\
    \emojilion~Lion  &     28.70   &   170.42    &      \textbf{25.69}    \\
    \emojitiger~Tiger  &    45.67     &   173.47      &     \textbf{39.54}  \\
    \emojiwolf~Wolf  &      59.43 &  161.47     &      \textbf{53.82}    \\
    \bottomrule
    \end{tabular}%
    }
  \label{tab:quan_FID}%
\end{table}%
\topic{StyleGAN adaptation.}
We also apply our proposed method to image generator adaptation. 
We conduct experiments on StyleGAN2~\cite{stylegan2} with our feature matching loss $L_{FM}$ and  KL loss $L_{KL}$. 
To demonstrate the effectiveness of our method, we compare our approach StyleGANFusion~\cite{song2022diffusion} and StyleGAN-NADA~\cite{gal2022stylegannada}.
We follow the three metrics used in ~\cite{song2022diffusion} to evaluate different methods: 
FID score~\cite{heusel2017gans}, CLIP score~\cite{radford2021learning}, and LPIPS score~\cite{zhang2018perceptual}.
For our method, we apply the feature matching loss $L_{FM}$ to StyleGANFusion as additional guidance with KL loss $L_{KL}$.

\begin{figure*}
\centering

\mpage{0.01}{\raisebox{-10pt}{\rotatebox{90}{\small{bear \textrightarrow polar bear}}}}
\mpage{0.235}{\frame{\includegraphics[width=\linewidth, trim=0 0 0 0, clip]{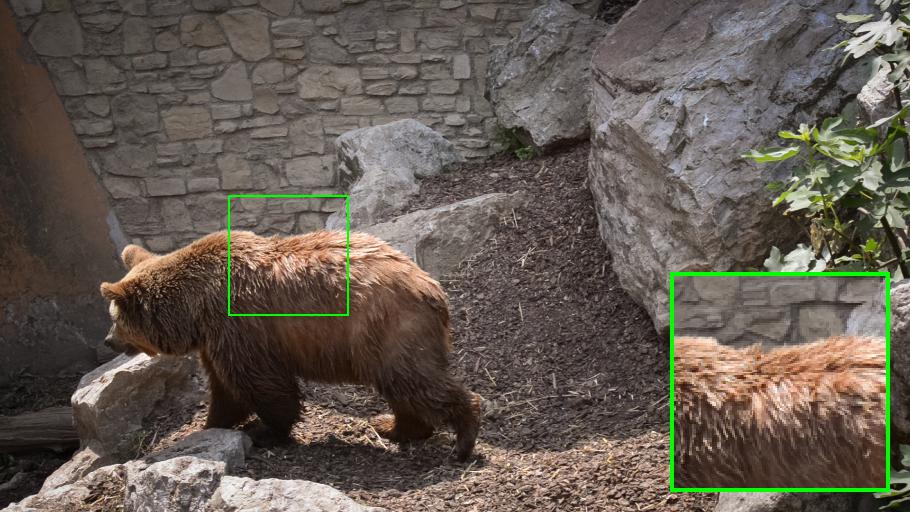}}}\hfill
\mpage{0.235}{\frame{\includegraphics[width=\linewidth, trim=0 0 0 0, clip]{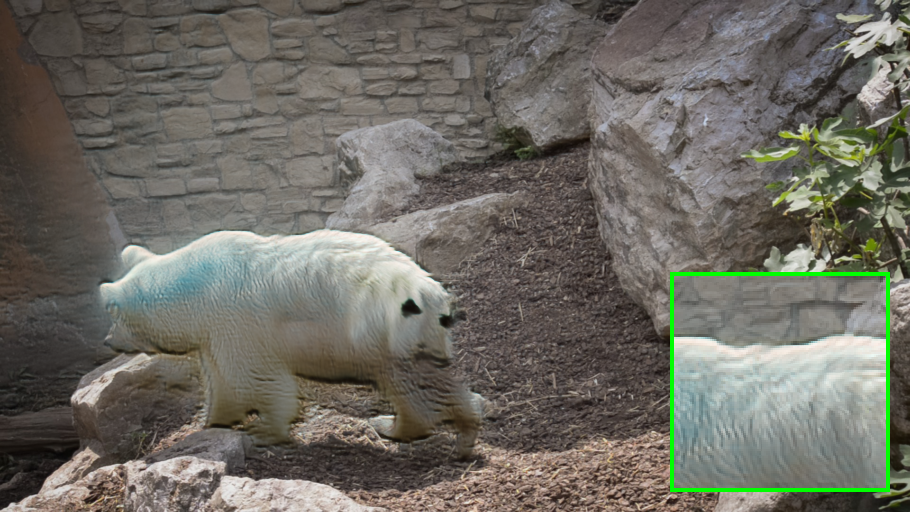}}}\hfill
\mpage{0.235}{\frame{\includegraphics[width=\linewidth, trim=0 0 0 0, clip]{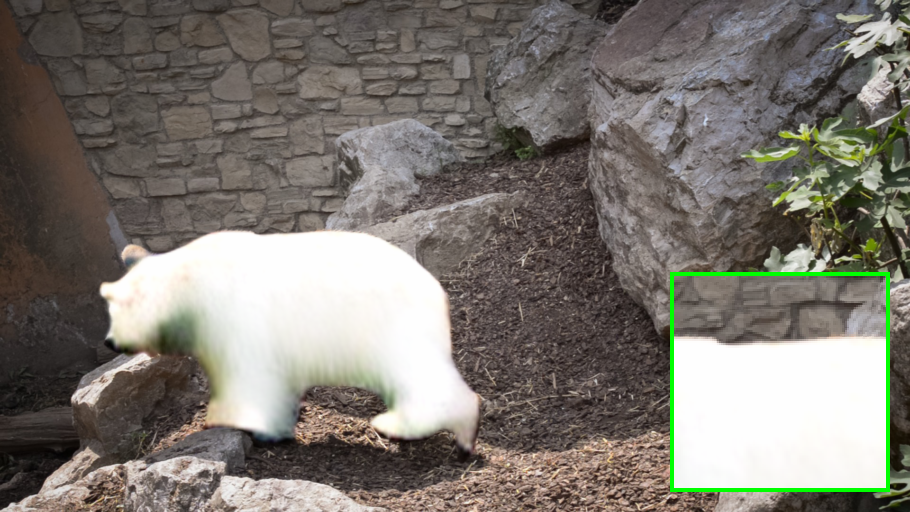}}}\hfill
\mpage{0.235}{\frame{\includegraphics[width=\linewidth, trim=0 0 0 0, clip]{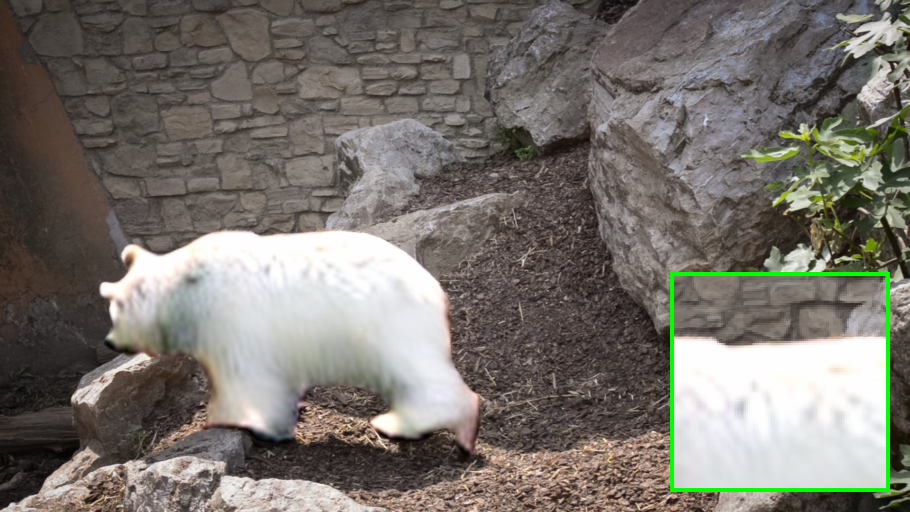}}}\\

\vspace{1mm}

\mpage{0.01}{\raisebox{-10pt}{\rotatebox{90}{\scriptsize{blackswan\textrightarrow whiteswan}}}}
\mpage{0.235}{\frame{\includegraphics[width=\linewidth, trim=0 0 0 0, clip]{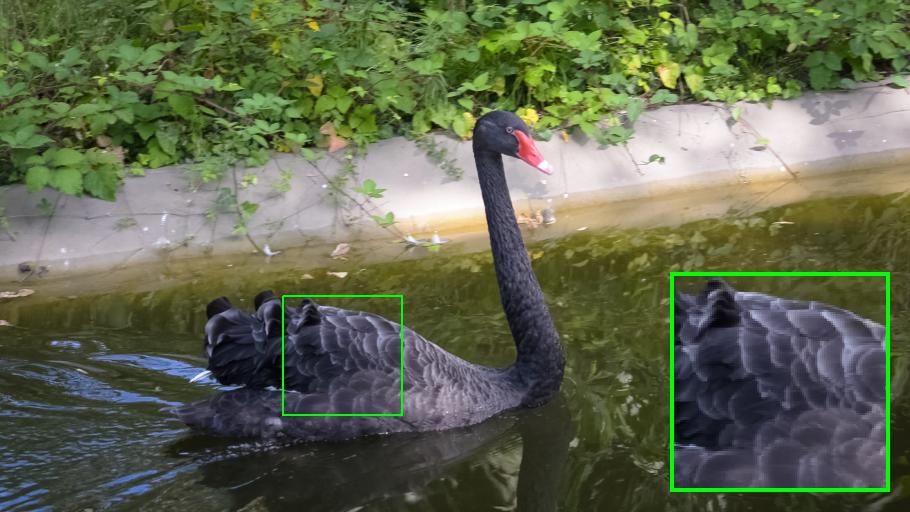}}}\hfill
\mpage{0.235}{\frame{\includegraphics[width=\linewidth, trim=0 0 0 0, clip]{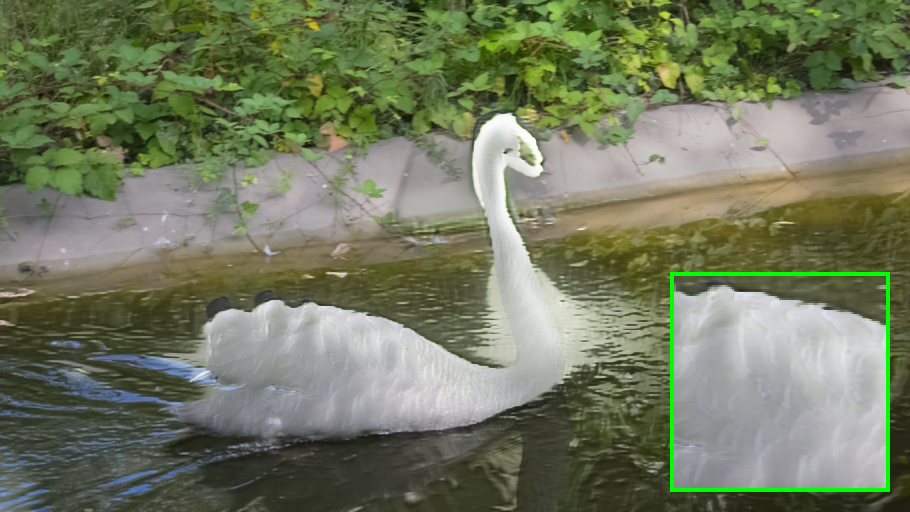}}}\hfill
\mpage{0.235}{\frame{\includegraphics[width=\linewidth, trim=0 0 0 0, clip]{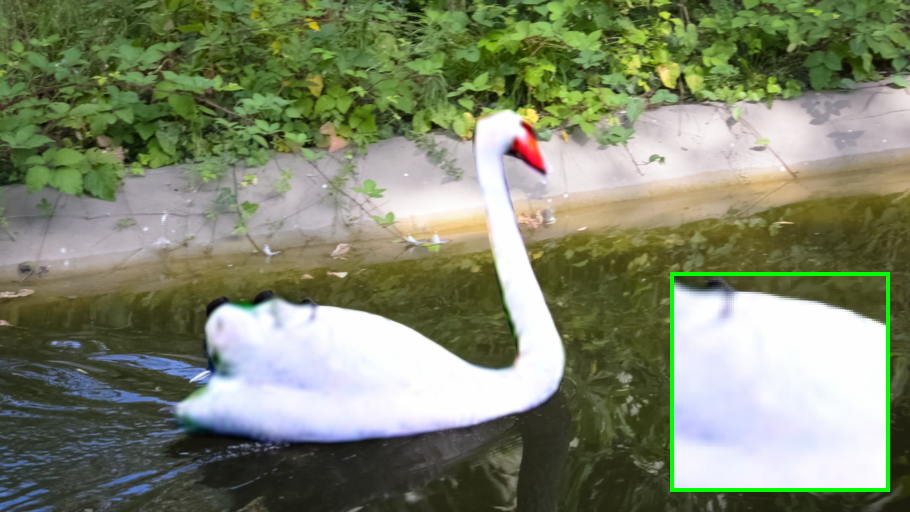}}}\hfill
\mpage{0.235}{\frame{\includegraphics[width=\linewidth, trim=0 0 0 0, clip]{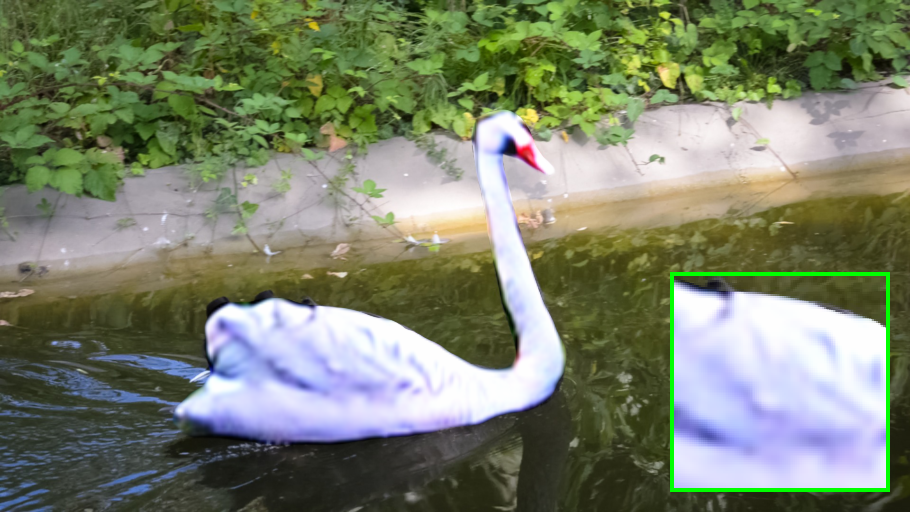}}}\\

\vspace{1mm}

\mpage{0.01}{\raisebox{-10pt}{\rotatebox{90}{\small{horse \textrightarrow zebra}}}}
\mpage{0.235}{\frame{\includegraphics[width=\linewidth, trim=0 0 0 0, clip]{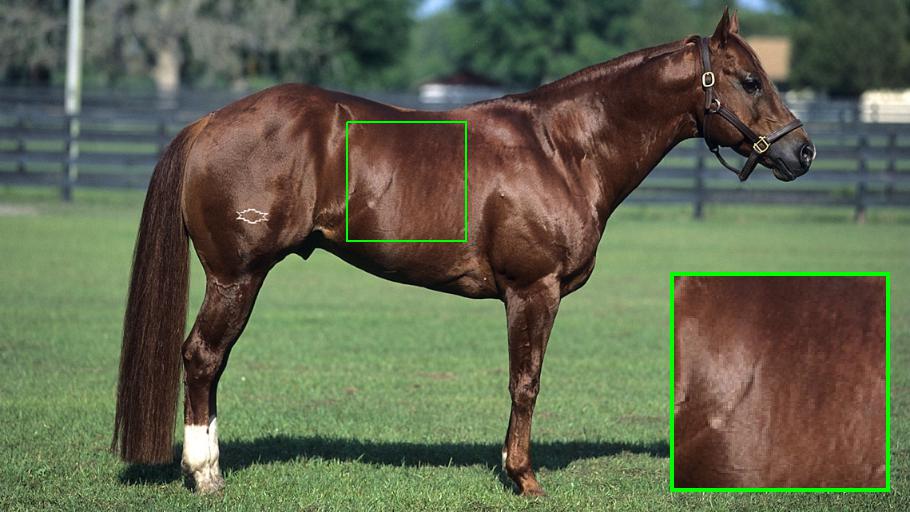}}}\hfill
\mpage{0.235}{\frame{\includegraphics[width=\linewidth, trim=0 0 0 0, clip]{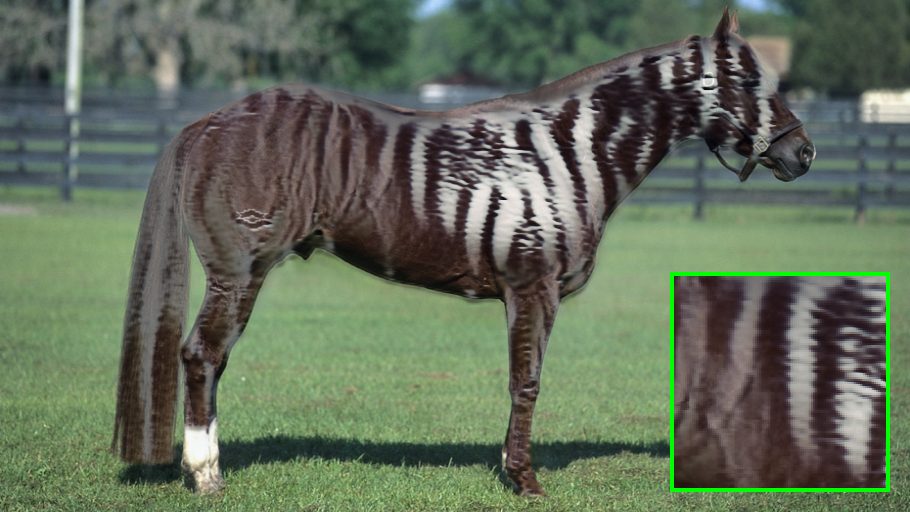}}}\hfill
\mpage{0.235}{\frame{\includegraphics[width=\linewidth, trim=0 0 0 0, clip]{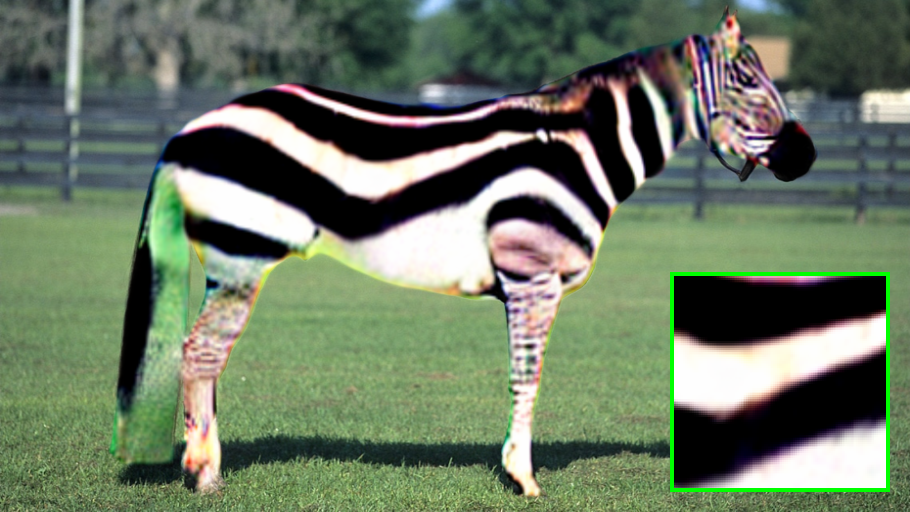}}}\hfill
\mpage{0.235}{\frame{\includegraphics[width=\linewidth, trim=0 0 0 0, clip]{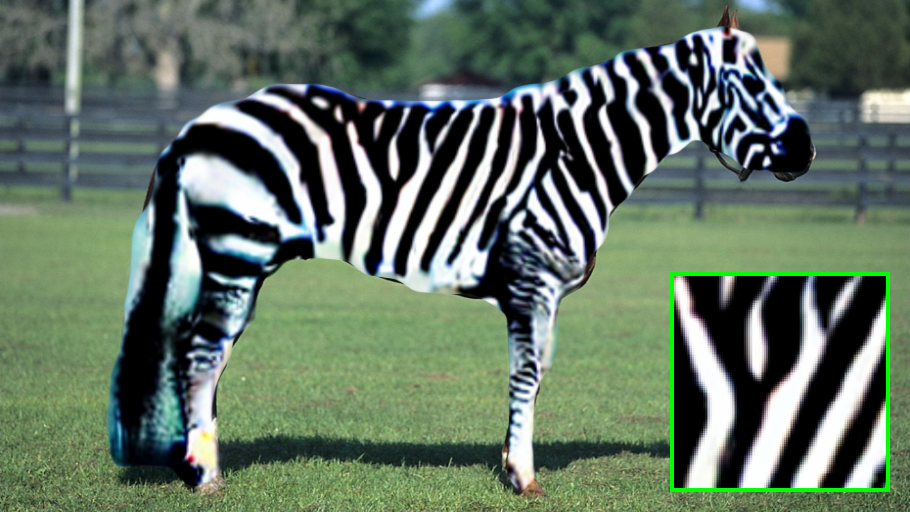}}}\\

\vspace{1mm}

\mpage{0.01}{\raisebox{-10pt}{\rotatebox{90}{\small{cat\textrightarrow tiger}}}}
\mpage{0.235}{\frame{\includegraphics[width=\linewidth, trim=0 0 0 0, clip]{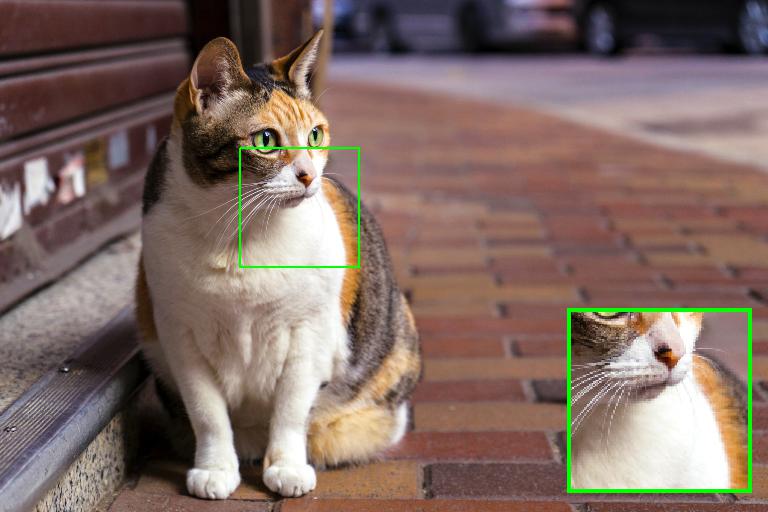}}}\hfill
\mpage{0.235}{\frame{\includegraphics[width=\linewidth, trim=0 0 0 0, clip]{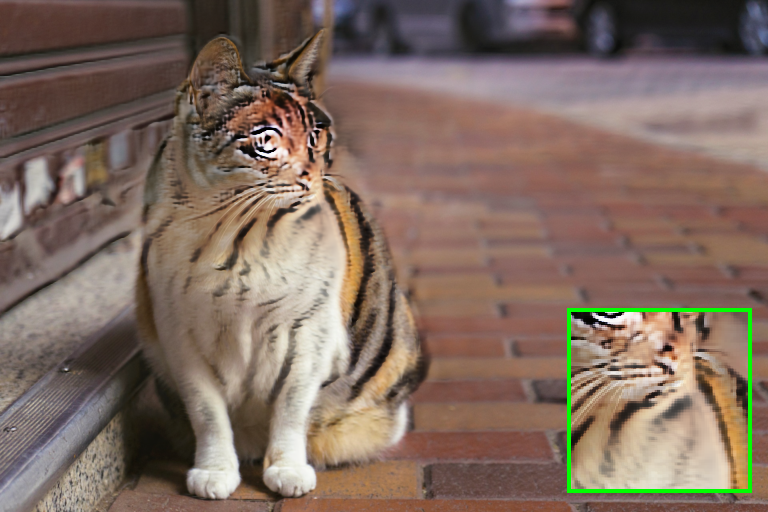}}}\hfill
\mpage{0.235}{\frame{\includegraphics[width=\linewidth, trim=0 0 0 0, clip]{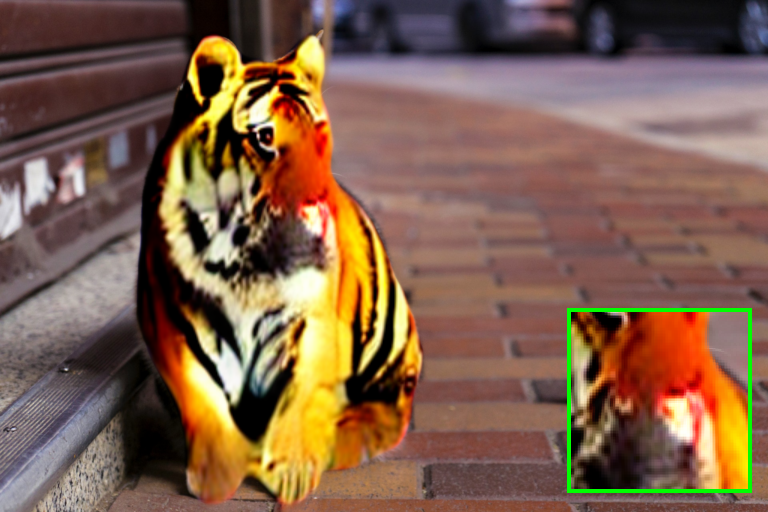}}}\hfill
\mpage{0.235}{\frame{\includegraphics[width=\linewidth, trim=0 0 0 0, clip]{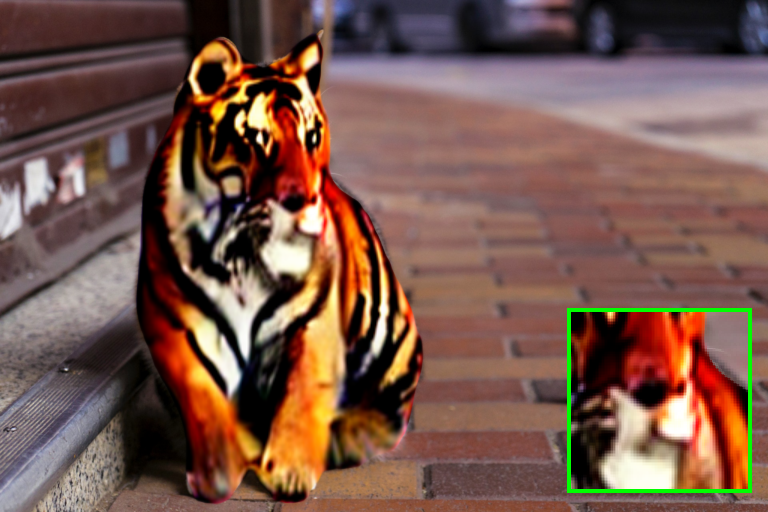}}}\\

\vspace{1mm}

\mpage{0.01}{\raisebox{0pt}{\rotatebox{90}{}}}
\mpage{0.235}{{{\small{Input}}}}\hfill
\mpage{0.235}{{{\small{Text2LIVE~\cite{bar2022text2live}}}}}\hfill
\mpage{0.235}{{{\small{$L_{LSD}$ baseline}}}}\hfill
\mpage{0.235}{{{\small{Ours ($L_{FM}+L_{LSD}+L_{KL}$)}}}}\\

\caption{\textbf{Comparisons on layered image editing.} We compare our method with the SOTA Text2LIVE~\cite{bar2022text2live} and the latent diffusion prior $L_{LSD}$ baseline. The leftmost column shows the editing text prompts. The zoom-in windows (highlighted in green) are shown at the right-bottom of the images to allow comparisons in detail areas. Our method can produce reasonable editing and more details than the baseline.}
\label{fig:comparison_layered_editing}
\end{figure*}

\begin{table}[htbp]
  \centering
  \caption{We compare our method with StyleGAN-NADA~\cite{gal2022stylegannada} and StyleGANFusion~\cite{song2022diffusion} on animal experiments. CLIP score is computed on generated samples and input text prompts to examine the text-image similarity. LPIPS is computed between all possible pairs in the generated samples to evaluate the diversity of images.
  }
  \resizebox{\columnwidth}{!}{%
    \begin{tabular}{lrrrrrr}
    \toprule
          & \multicolumn{2}{c}{StyleGANFusion} & \multicolumn{2}{c}{StyleGAN-NADA} & \multicolumn{2}{c}{Ours} \\
    \cmidrule(lr){2-3} \cmidrule(lr){4-5} \cmidrule(lr){6-7}
          Cat $\rightarrow$ & \multicolumn{1}{l}{CLIP$\uparrow$} & \multicolumn{1}{l}{LPIPS$\uparrow$} & \multicolumn{1}{l}{CLIP$\uparrow$} & \multicolumn{1}{l}{LPIPS$\uparrow$} & \multicolumn{1}{l}{CLIP$\uparrow$} & \multicolumn{1}{l}{LPIPS$\uparrow$} \\
    \midrule
    \emojidog~Dog    &   0.285    &  0.584    &   0.288    &  0.457      &   \textbf{0.289}    & \textbf{ 0.594}\\
    \emojihamster~Hamster   &   0.333    &  0.544   &   \textbf{0.352}    &  0.460      &   0.329    &   \textbf{0.569} \\
    \emojibadger~Badger  &   0.311    &  0.554    &   \textbf{0.368}    &  0.435       &   0.297    &   \textbf{0.555}\\
    \emojifox~Fox   &   0.321    &  \textbf{0.560}    & \textbf{ 0.343}     &   0.554      &   0.318    &   0.536 \\
    \emojiotter~Otter   &   0.323    &  0.523   &   \textbf{0.368}    &   0.389      &  0.317     &   \textbf{0.542}\\
    \emojilion~Lion    &  0.299     &   \textbf{0.577}   &  0.316     &   0.478    &    \textbf{0.302}   &   0.555\\
    \emojibear~Bear  &   0.294    &  \textbf{0.567}     &   \textbf{0.332}    &    0.399    &   0.305    &   0.555\\
    \emojipig~Pig   &  0.318     &  0.586   &   \textbf{0.322}    &  0.549      &   0.306    &  \textbf{0.596} \\
    \bottomrule
    \end{tabular}%
    }
 \vspace{-1mm}
  \label{tab:quan_CLIP}%
\end{table}%
\begin{figure*}
\centering

\mpage{0.01}{\raisebox{0pt}{\rotatebox{90}{\small{Cat $\rightarrow$} fox}}}  \hfill
\mpage{0.11}{\frame{\includegraphics[width=\linewidth, trim=0 0 0 0, clip]{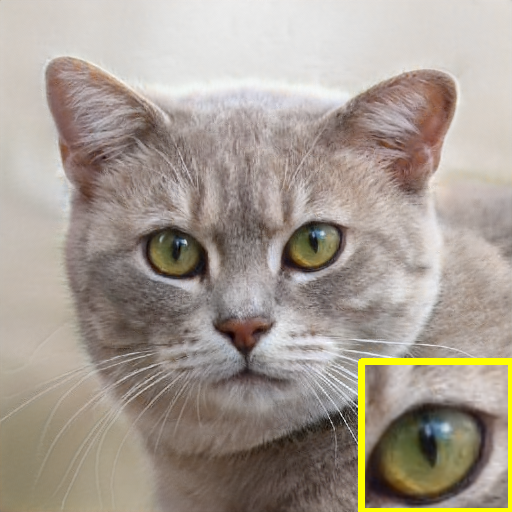}}}\hfill
\mpage{0.11}{\frame{\includegraphics[width=\linewidth, trim=0 0 0 0, clip]{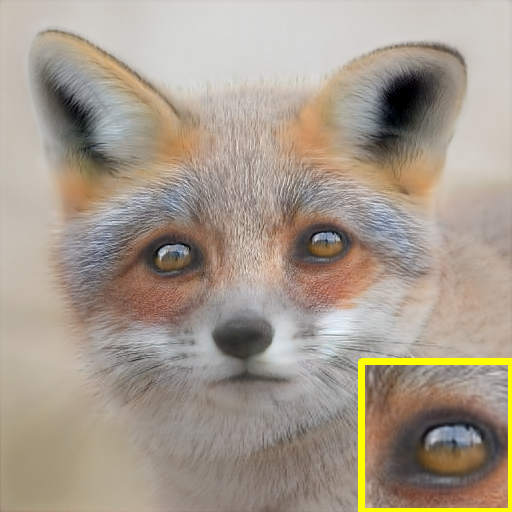}}}\hfill
\mpage{0.11}{\frame{\includegraphics[width=\linewidth, trim=0 0 0 0, clip]{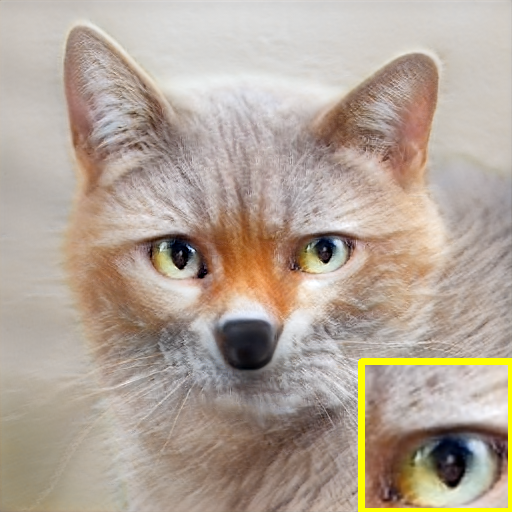}}}\hfill
\mpage{0.11}{\frame{\includegraphics[width=\linewidth, trim=0 0 0 0, clip]{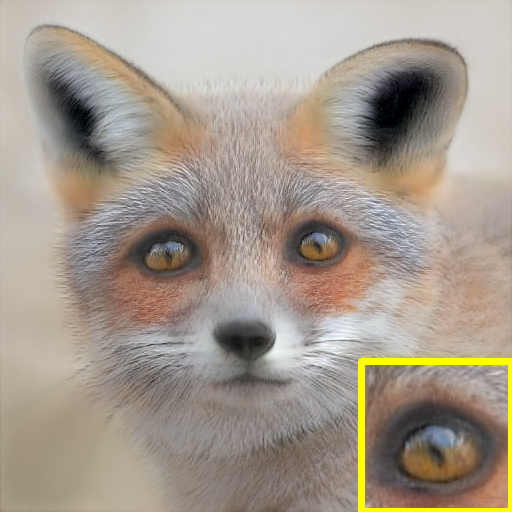}}}\hfill
\mpage{0.01}{\raisebox{0pt}{\rotatebox{90}{\small{prompt 1\footnotemark}}}}  \hfill
\mpage{0.11}{\frame{\includegraphics[width=\linewidth, trim=0 0 0 0, clip]{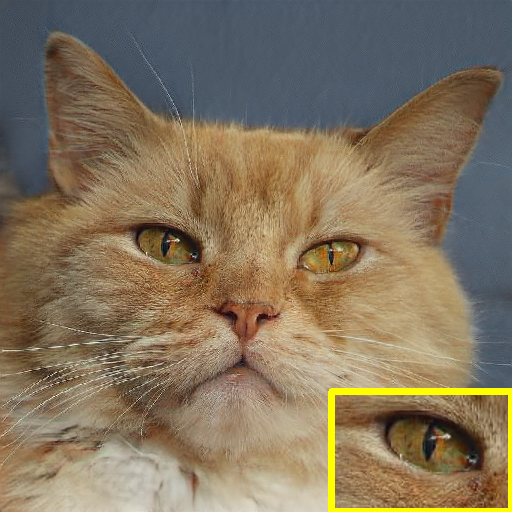}}}\hfill
\mpage{0.11}{\frame{\includegraphics[width=\linewidth, trim=0 0 0 0, clip]{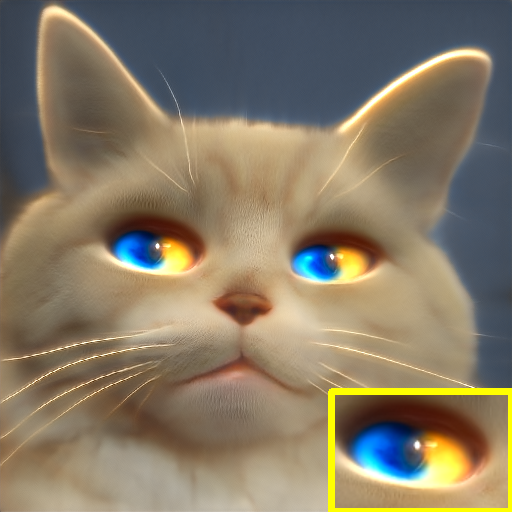}}}\hfill
\mpage{0.11}{\frame{\includegraphics[width=\linewidth, trim=0 0 0 0, clip]{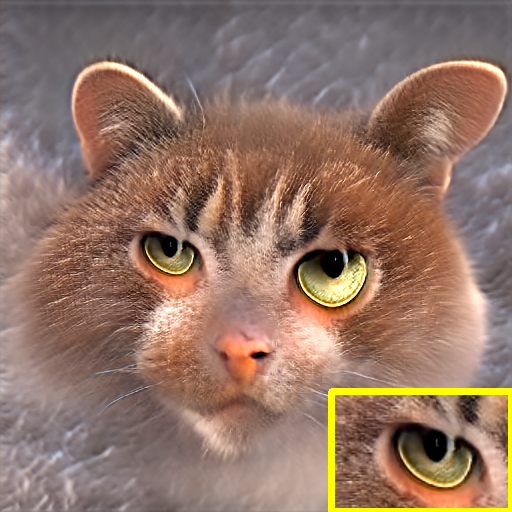}}}\hfill
\mpage{0.11}{\frame{\includegraphics[width=\linewidth, trim=0 0 0 0, clip]{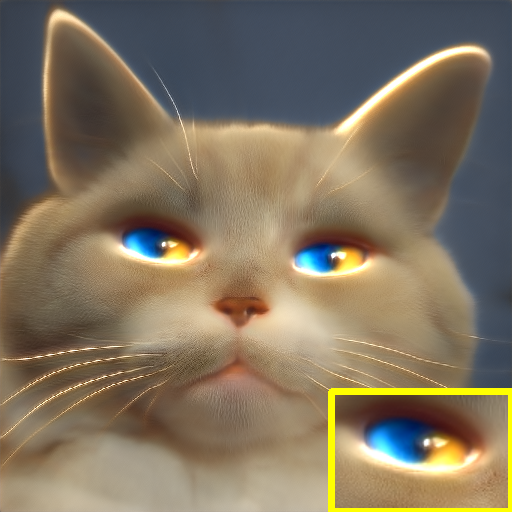}}}\\

\vspace{1mm}

\mpage{0.01}{\raisebox{0pt}{\rotatebox{90}{\small{Cat $\rightarrow$} hamster}}}  \hfill
\mpage{0.11}{\frame{\includegraphics[width=\linewidth, trim=0 0 0 0, clip]{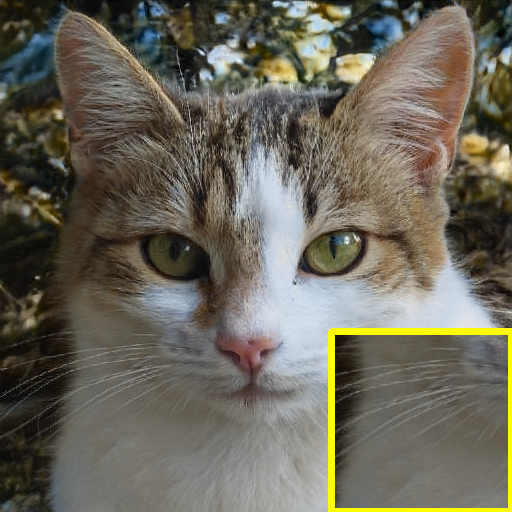}}}\hfill
\mpage{0.11}{\frame{\includegraphics[width=\linewidth, trim=0 0 0 0, clip]{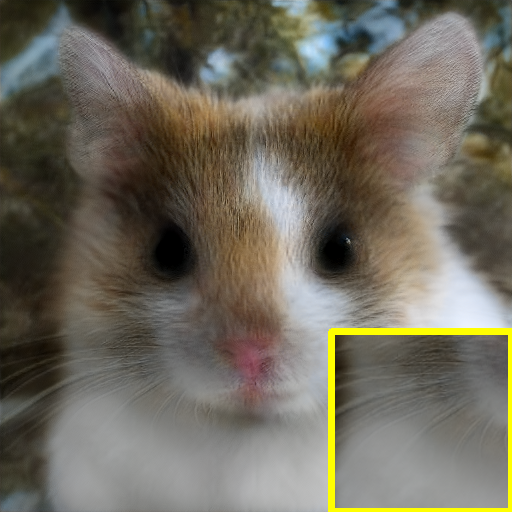}}}\hfill
\mpage{0.11}{\frame{\includegraphics[width=\linewidth, trim=0 0 0 0, clip]{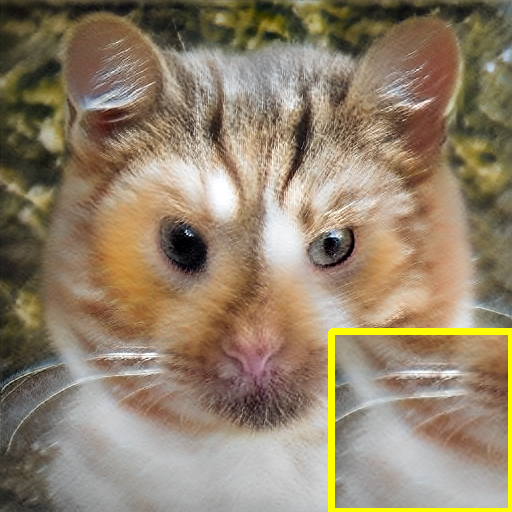}}}\hfill
\mpage{0.11}{\frame{\includegraphics[width=\linewidth, trim=0 0 0 0, clip]{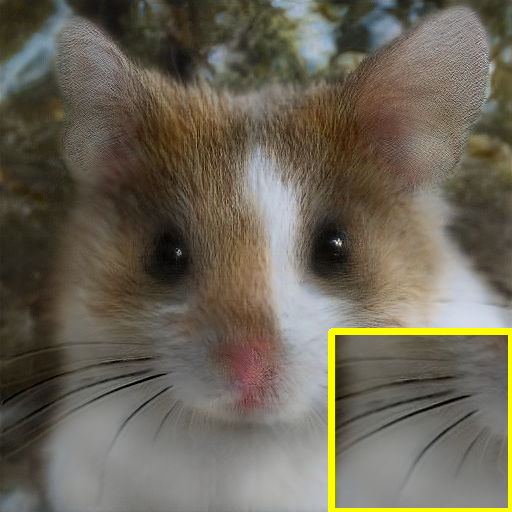}}}\hfill
\mpage{0.01}{\raisebox{0pt}{\rotatebox{90}{\small{prompt 2\footnotemark}}}}  \hfill
\mpage{0.11}{\frame{\includegraphics[width=\linewidth, trim=0 0 0 0, clip]{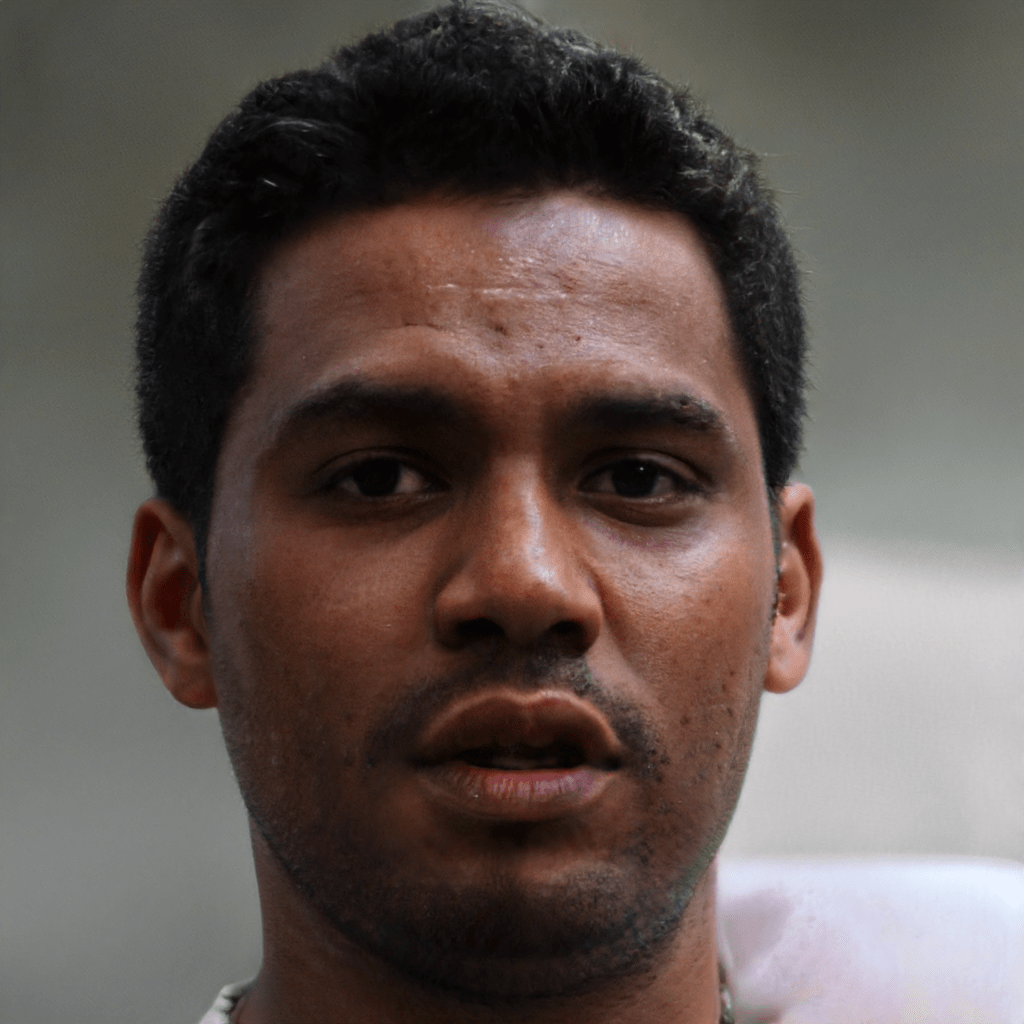}}}\hfill
\mpage{0.11}{\frame{\includegraphics[width=\linewidth, trim=0 0 0 0, clip]{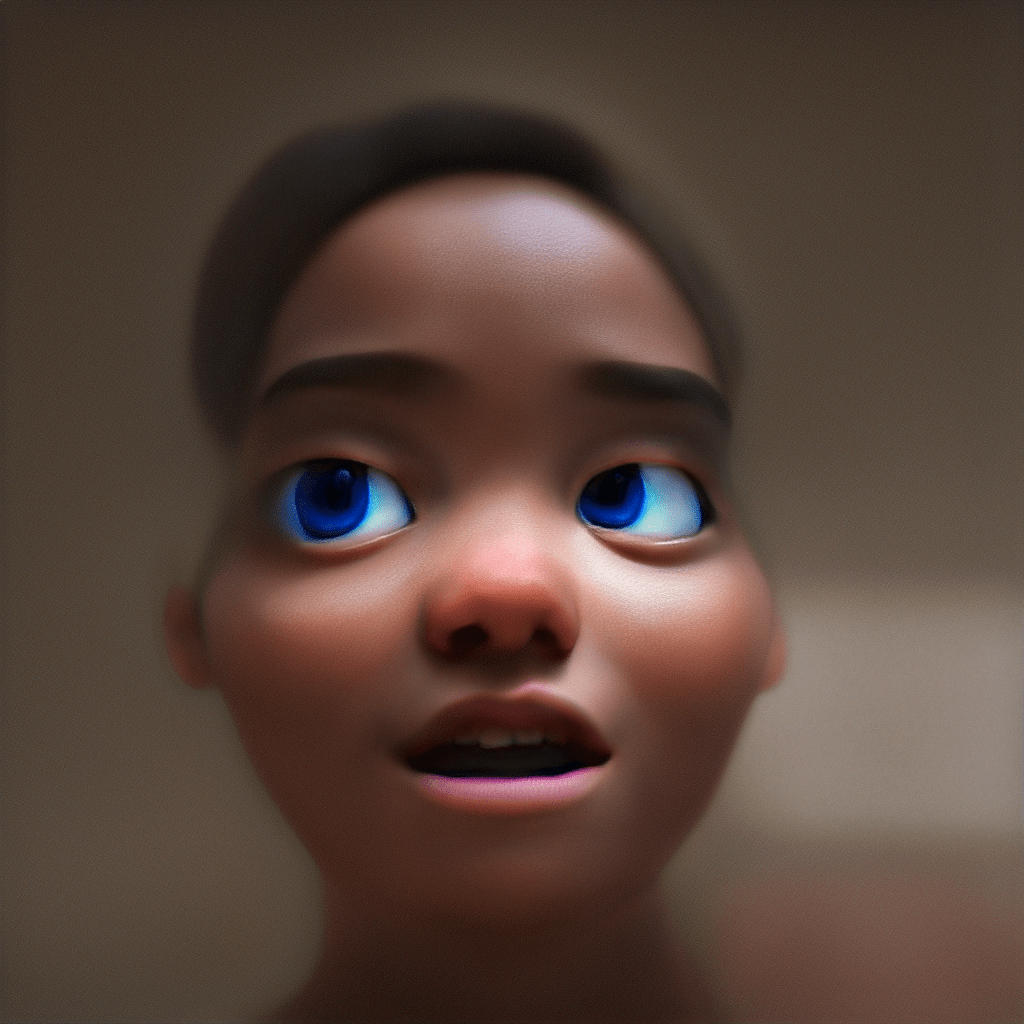}}}\hfill
\mpage{0.11}{\frame{\includegraphics[width=\linewidth, trim=0 0 0 0, clip]{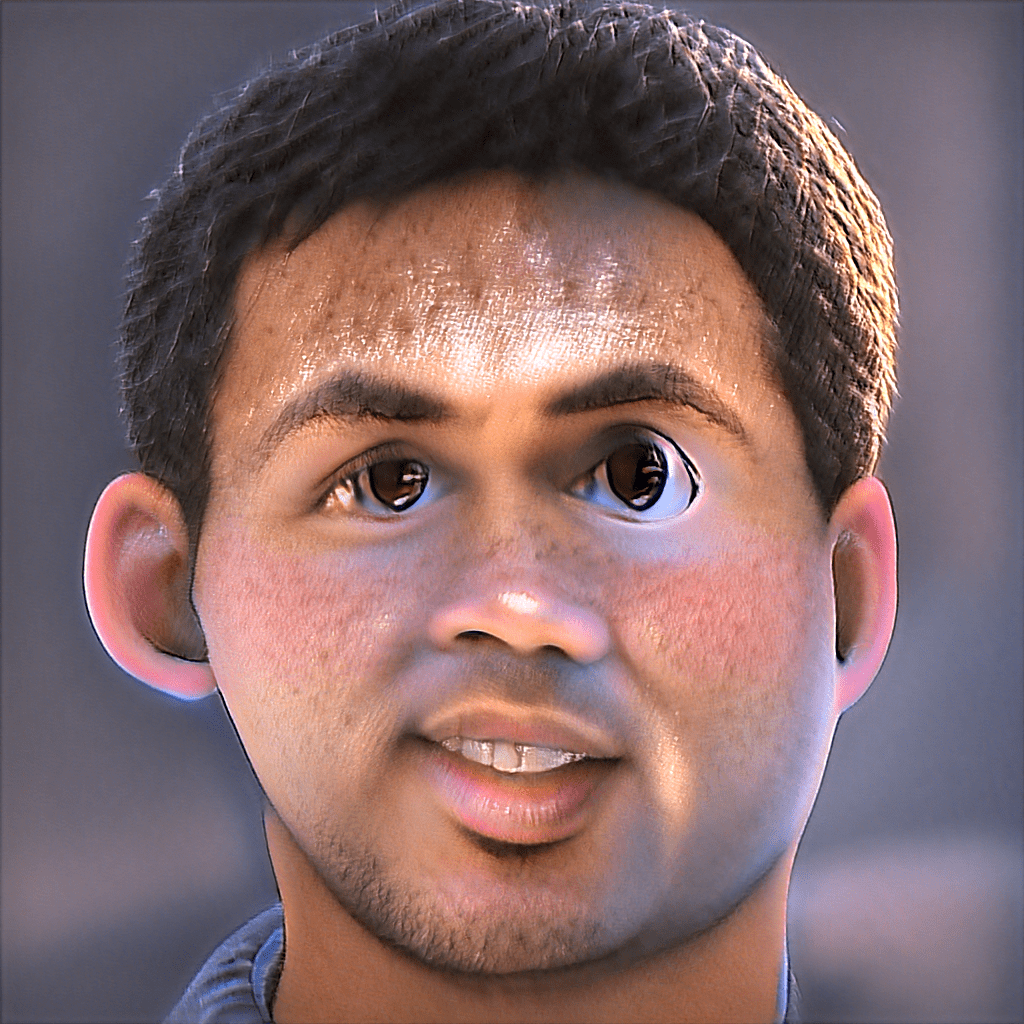}}}\hfill
\mpage{0.11}{\frame{\includegraphics[width=\linewidth, trim=0 0 0 0, clip]{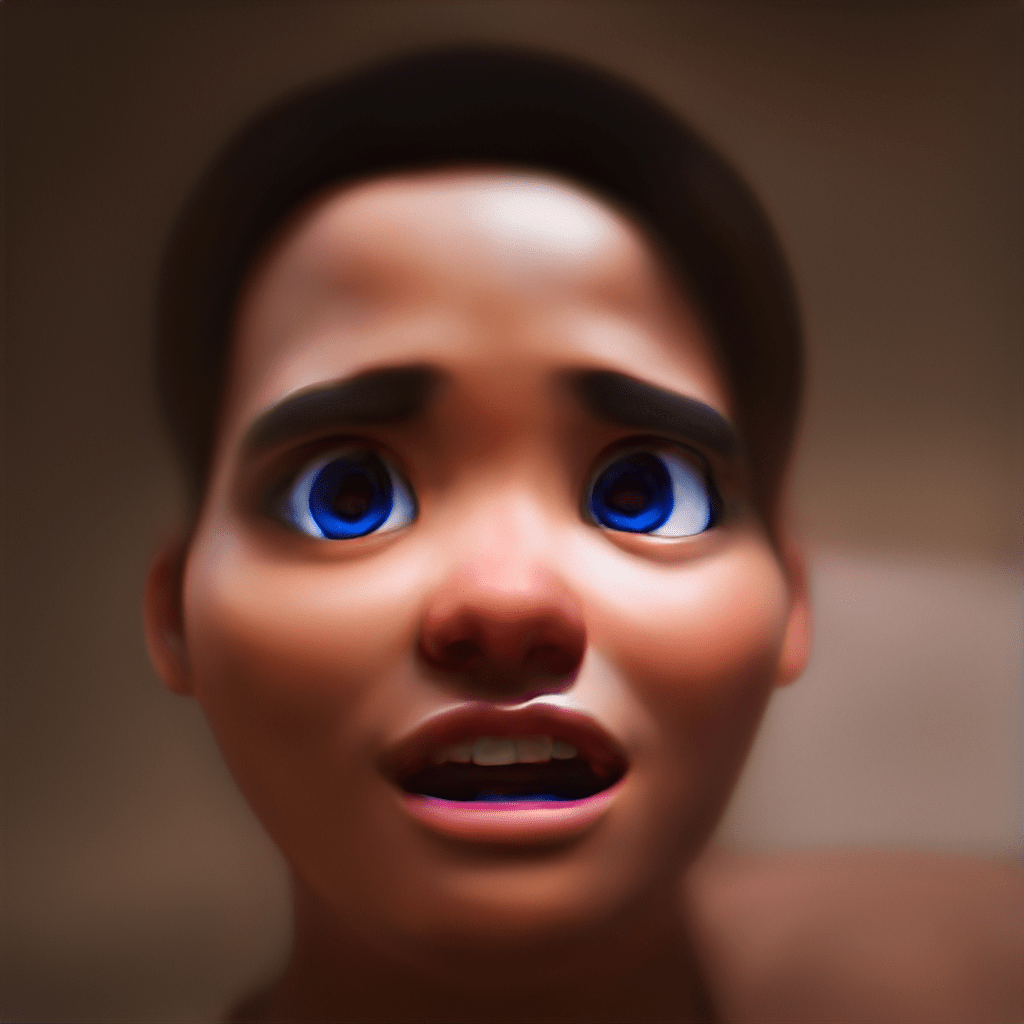}}}\hfill\\

\vspace{1mm}

\mpage{0.01}{\raisebox{0pt}{\rotatebox{90}{\small{Cat $\rightarrow$} lion}}}  \hfill
\mpage{0.11}{\frame{\includegraphics[width=\linewidth, trim=0 0 0 0, clip]{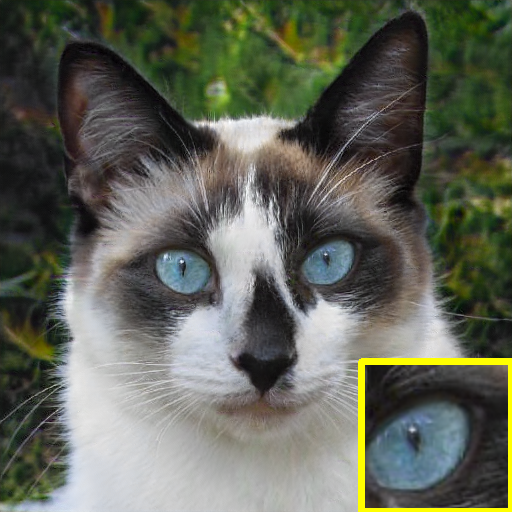}}}\hfill
\mpage{0.11}{\frame{\includegraphics[width=\linewidth, trim=0 0 0 0, clip]{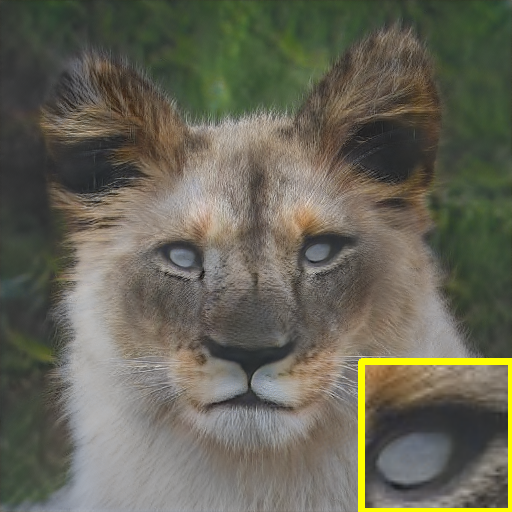}}}\hfill
\mpage{0.11}{\frame{\includegraphics[width=\linewidth, trim=0 0 0 0, clip]{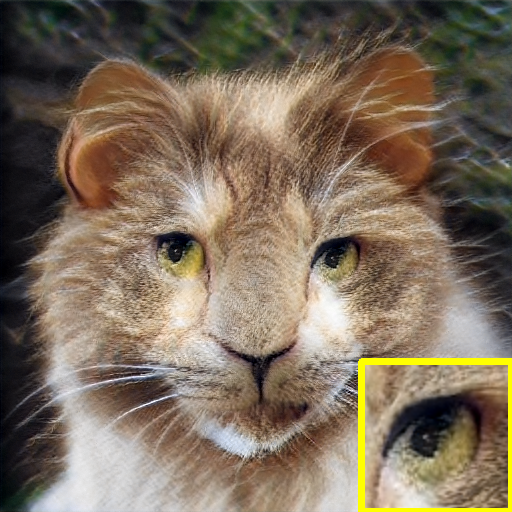}}}\hfill
\mpage{0.11}{\frame{\includegraphics[width=\linewidth, trim=0 0 0 0, clip]{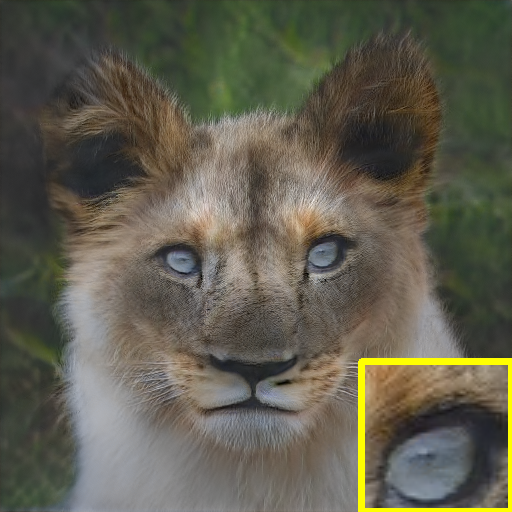}}}\hfill
\mpage{0.01}{\raisebox{0pt}{\rotatebox{90}{\small{prompt 3\footnotemark}}}}  \hfill
\mpage{0.11}{\frame{\includegraphics[width=\linewidth, trim=0 0 0 0, clip]{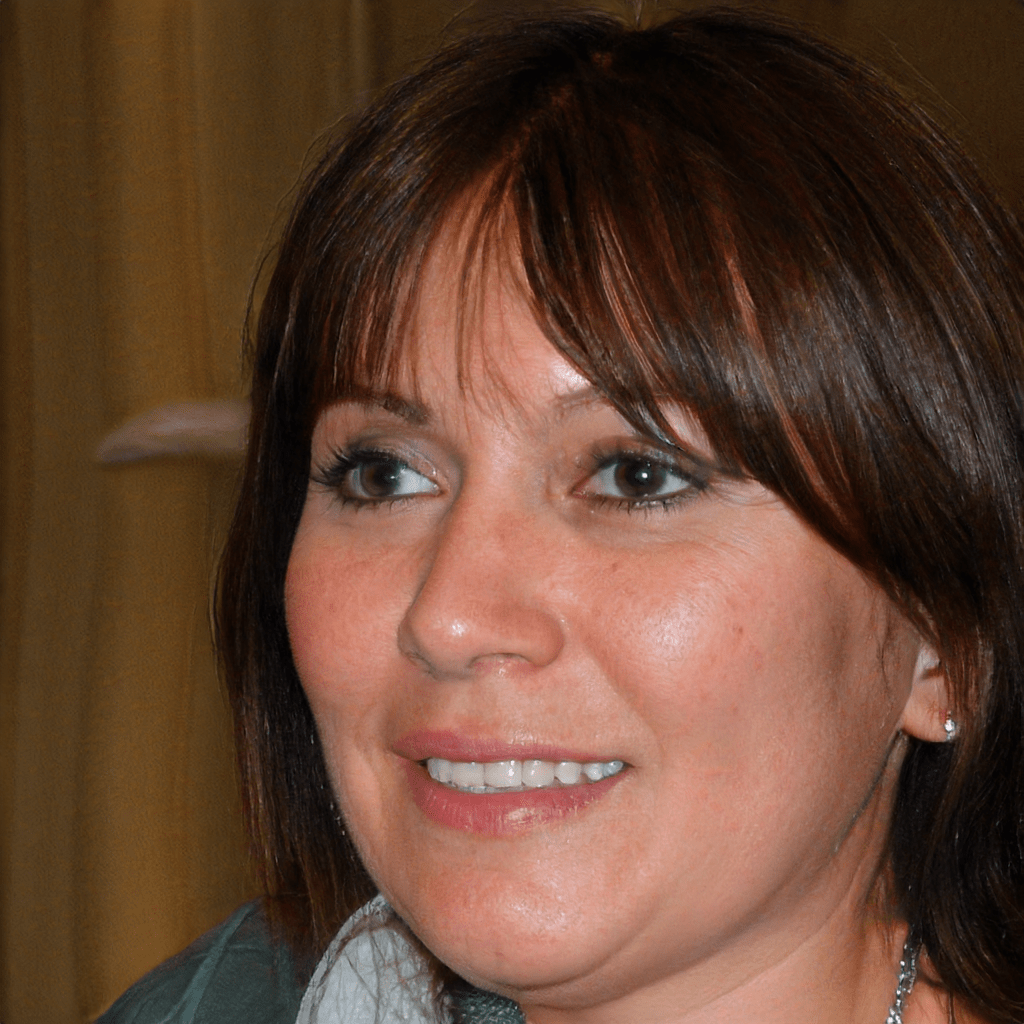}}}\hfill
\mpage{0.11}{\frame{\includegraphics[width=\linewidth, trim=0 0 0 0, clip]{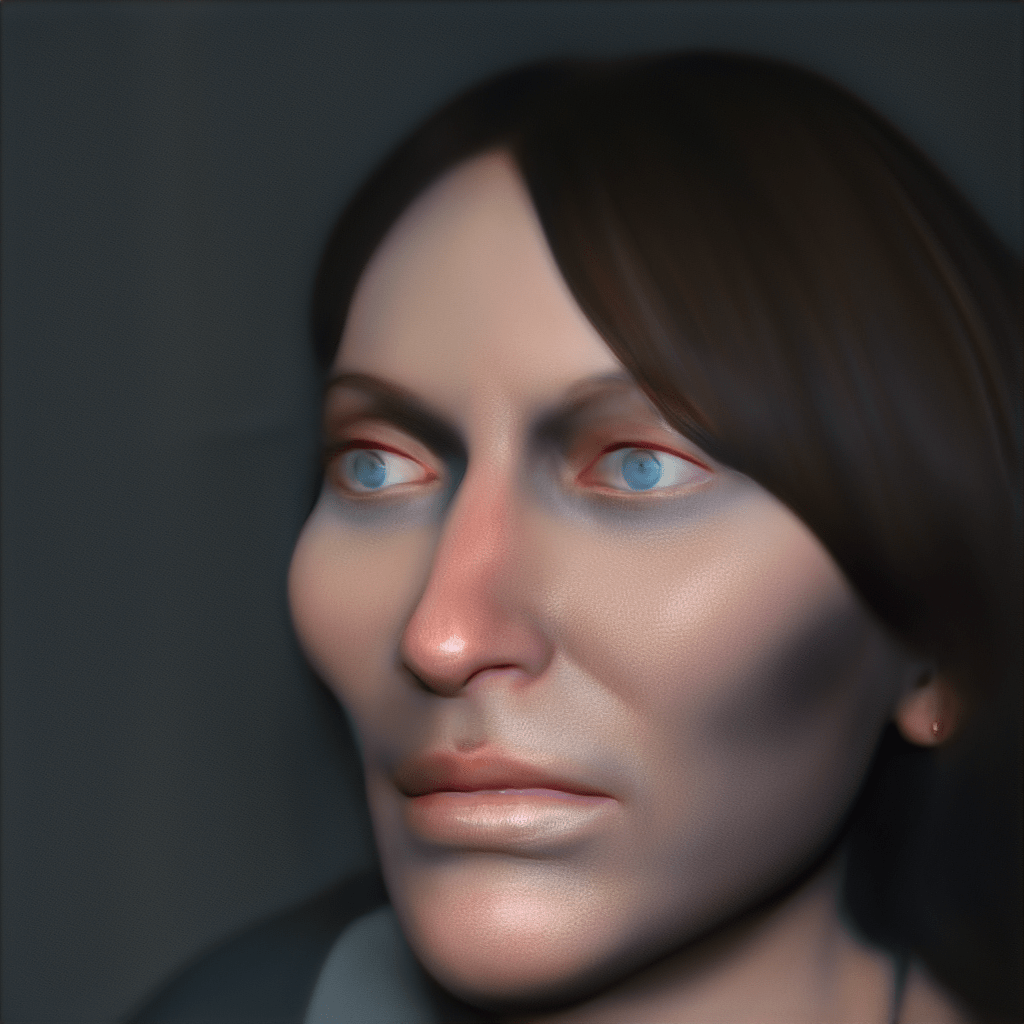}}}\hfill
\mpage{0.11}{\frame{\includegraphics[width=\linewidth, trim=0 0 0 0, clip]{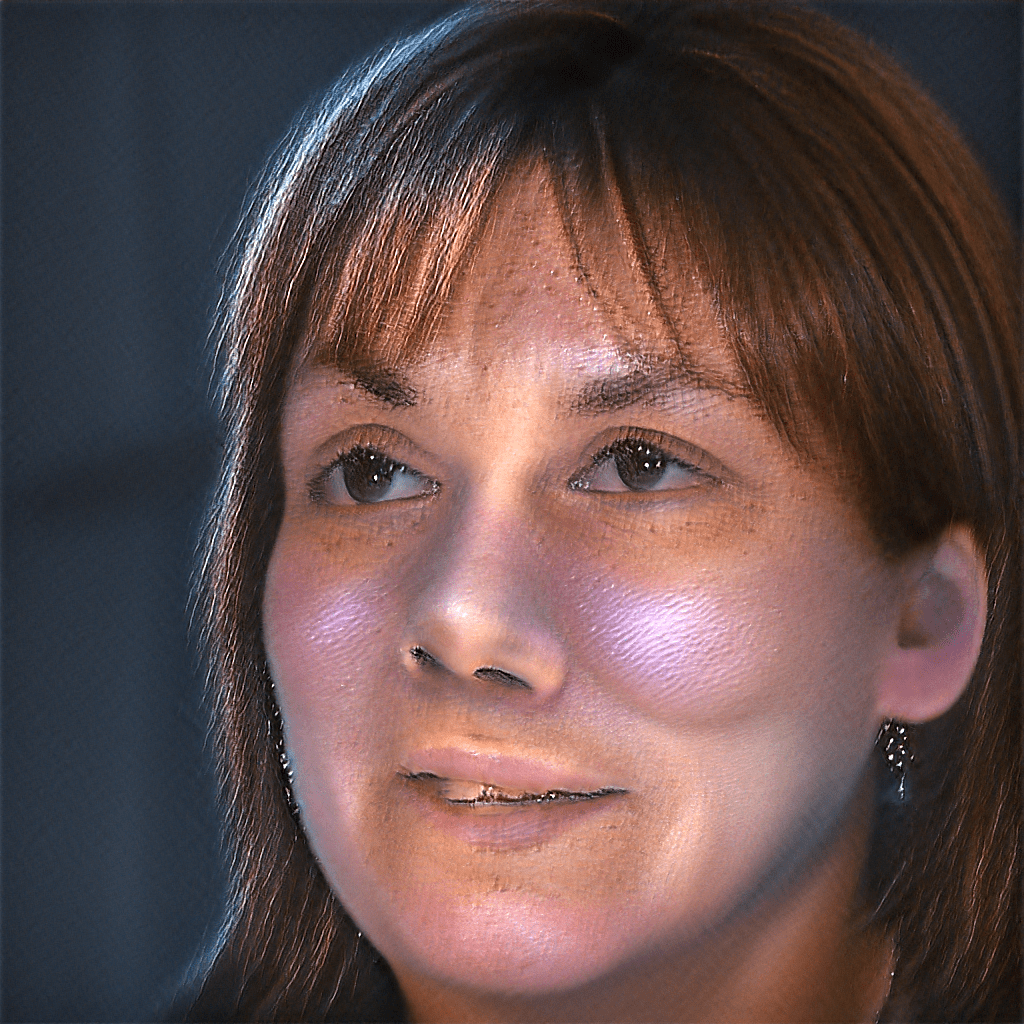}}}\hfill
\mpage{0.11}{\frame{\includegraphics[width=\linewidth, trim=0 0 0 0, clip]{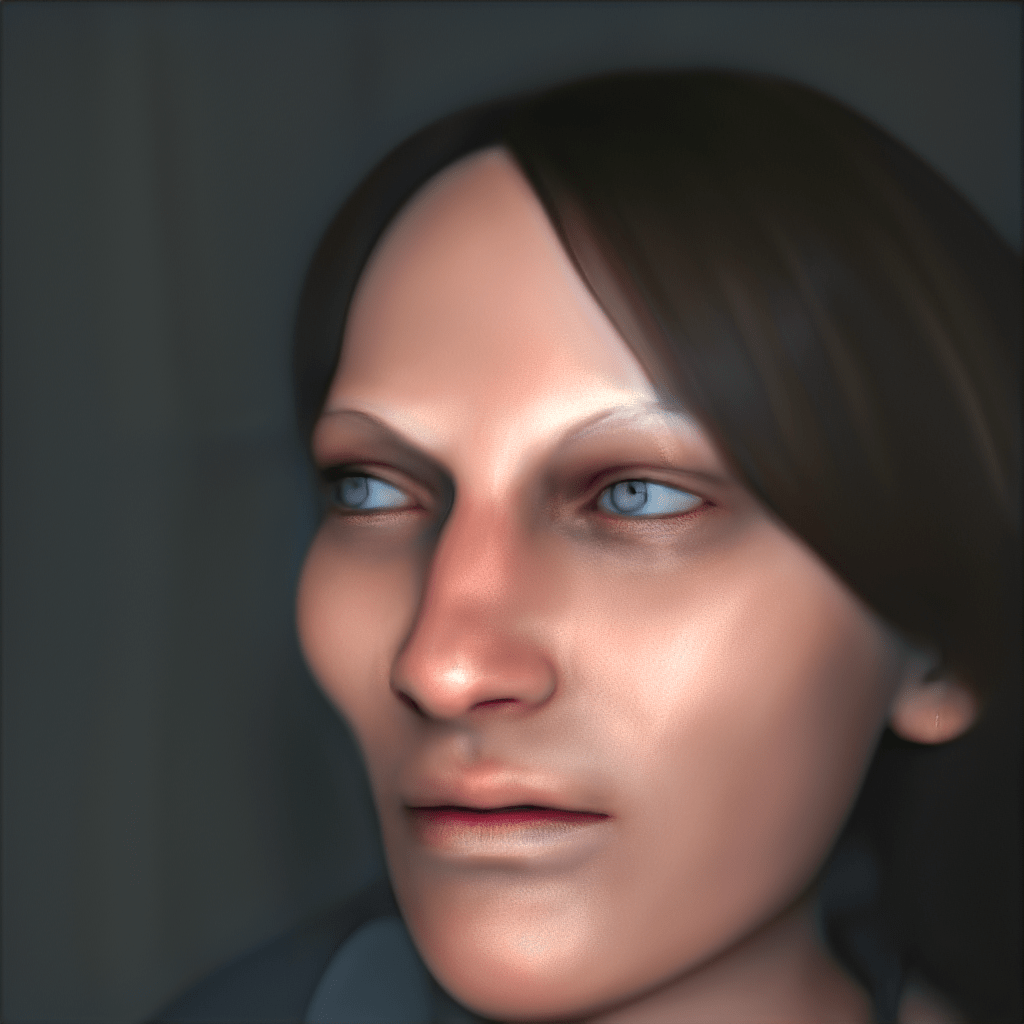}}}\\

\vspace{1mm}

\mpage{0.01}{{{\small{\!}}}}\hfill
\mpage{0.105}{{{\small{Source}}}}\hfill
\mpage{0.115}{{{\small{StyleGANFusion}}}}\hfill
\mpage{0.11}{{{\small{StyleGAN-NADA}}}}\hfill
\mpage{0.11}{{{\small{Ours}}}}\hfill
\mpage{0.02}{{{\small{\!}}}}\hfill
\mpage{0.105}{{{\small{Source}}}}\hfill
\mpage{0.12}{{{\small{StyleGANFusion}}}}\hfill
\mpage{0.11}{{{\small{StyleGAN-NADA}}}}\hfill
\mpage{0.11}{{{\small{Ours}}}}\\

\vspace{1mm}

\caption{\textbf{Comparison with StyleGANFusion~\cite{song2022diffusion} and StyleGAN-NADA~\cite{gal2022stylegannada}.}
We show uncurated samples from different methods. 
We test \emph{short prompts} like ``photo of a/an X'' on the left, 
and \emph{long prompts}
on the right.
Compared to baselines, our approach shows higher fidelity (\eg, whiskers) and keeps the attributes from the source image (\eg, pupils). We show the complete text prompts in the footnote.}
\label{fig:comparison_GAN}
\end{figure*}

We generate 2,000 samples from the adapted generator after the training for the quantitative evaluations.
We first show FID scores of different approaches on various target domains in Table~\ref{tab:quan_FID}. 
Groundtruth images are extracted from AFHQ dataset~\cite{choi2020stargan}. 
Some of the labels are provided by~\cite{song2022diffusion}.
Following StyleGANFusion~\cite{song2022diffusion}, we compare the FID scores on ``Cat $\rightarrow$ Dog/Fox/Lion/Tiger/Wolf''.
Our FID scores outperform other baselines in most cases.
This indicates that our method can help gain a more similar distribution to the target domain and improve the quality.

\begin{figure*}[t]
\centering

\mpage{0.15}{{{\small{$L_{LSD}$}}}}\hfill
\mpage{0.15}{{{\small{$L_{LSD}+L_{KL}$}}}}\hfill
\mpage{0.15}{{{\small{$L_{FM}$}}}}\hfill
\mpage{0.15}{{{\small{$L_{FM}+L_{KL}$}}}}\hfill
\mpage{0.15}{{{\small{$L_{FM}+L_{LSD}$}}}}\hfill
\mpage{0.15}{{{\small{$L_{FM}+L_{LSD}+L_{KL}$}}}}\\

\vspace{1mm}

\mpage{0.15}{\frame{\includegraphics[width=\linewidth, trim=0 0 0 0, clip]{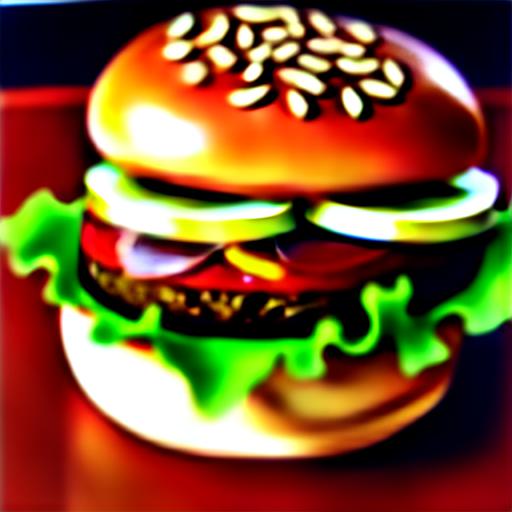}}}\hfill
\mpage{0.15}{\frame{\includegraphics[width=\linewidth, trim=0 0 0 0, clip]{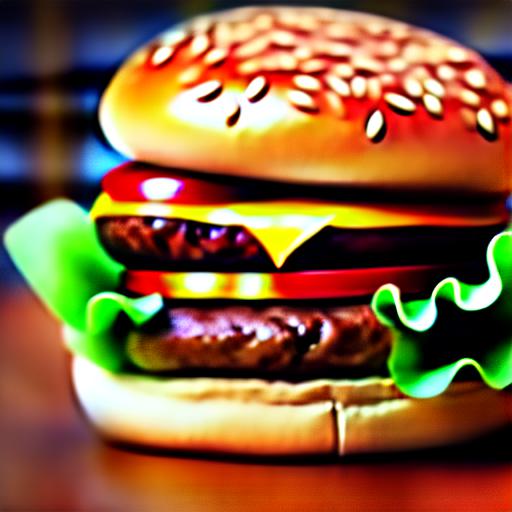}}}\hfill
\mpage{0.15}{\frame{\includegraphics[width=\linewidth, trim=0 0 0 0, clip]{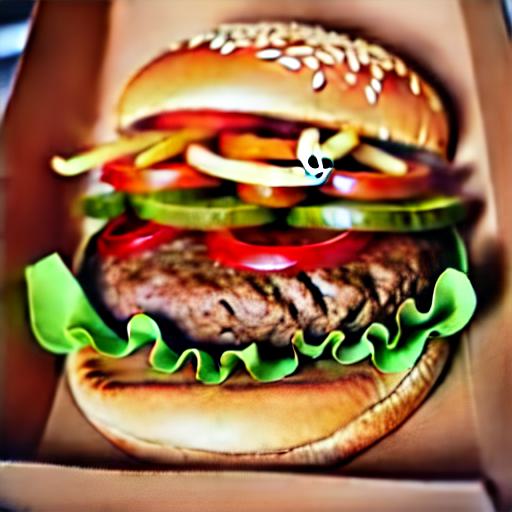}}}\hfill
\mpage{0.15}{\frame{\includegraphics[width=\linewidth, trim=0 0 0 0, clip]{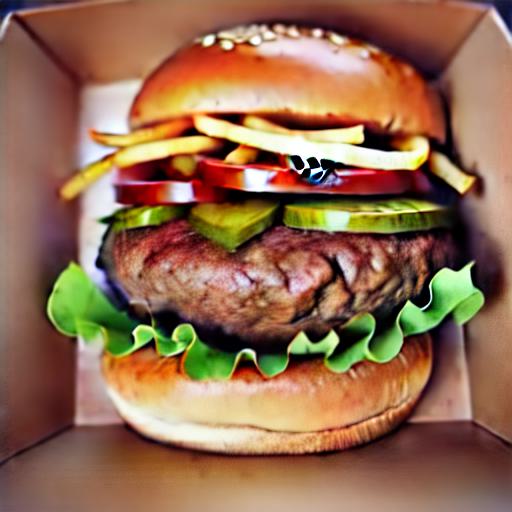}}}\hfill
\mpage{0.15}{\frame{\includegraphics[width=\linewidth, trim=0 0 0 0, clip]{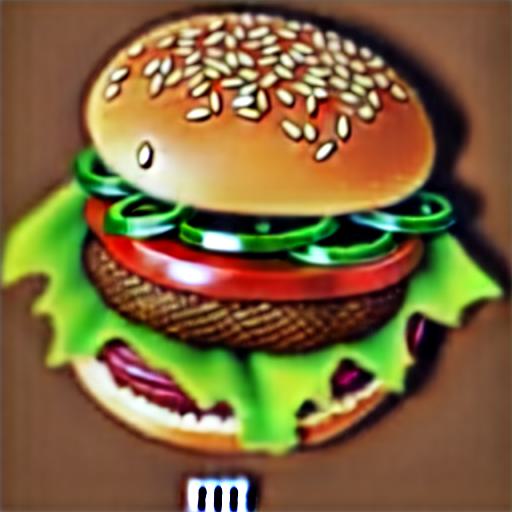}}}\hfill
\mpage{0.15}{\frame{\includegraphics[width=\linewidth, trim=0 0 0 0, clip]{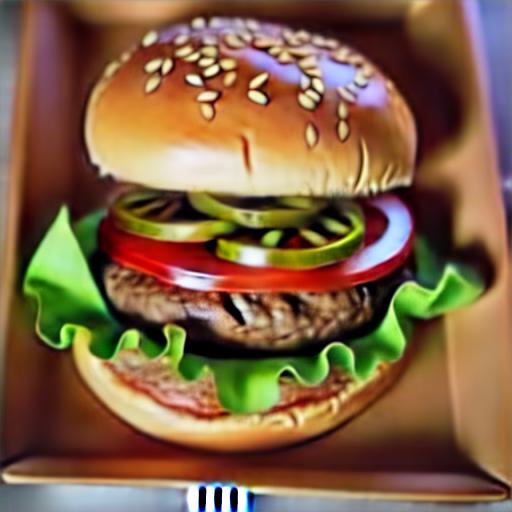}}}\\

\vspace{1mm}


\captionof{figure}{
\textbf{Ablation study.} 
We test our two component multi-level feature matching loss $L_{FM}$ and KL regulizier $L_{KL}$on an image-syntheis.
}
\label{fig:ablation}
\end{figure*}

\begin{figure*}[t]
\centering

\mpage{0.32}{{{\small{Text-to-3D}}}}\hfill
\mpage{0.32}{{{\small{StyleGAN adaptation}}}}\hfill
\mpage{0.32}{{{\small{Layered image editing}}}}\\

\vspace{1mm}

\mpage{0.15}{\frame{\includegraphics[width=\linewidth, trim=0 0 0 0, clip]{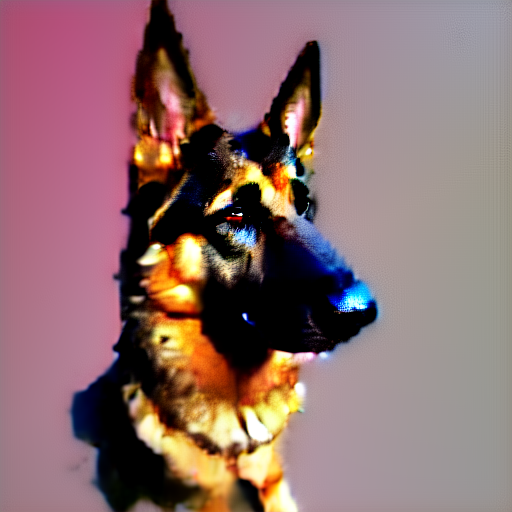}}}\hfill
\mpage{0.15}{\frame{\includegraphics[width=\linewidth, trim=0 0 0 0, clip]{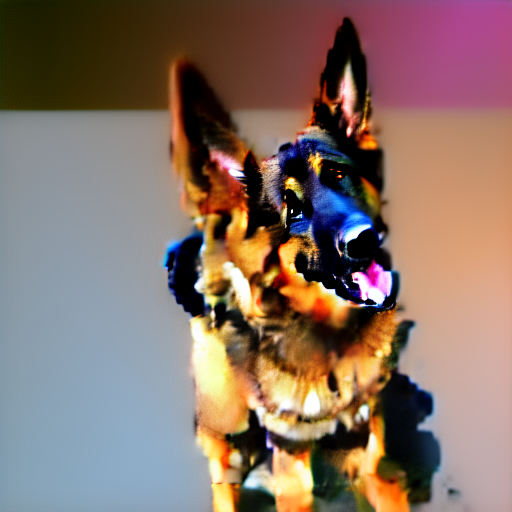}}}\hfill
\mpage{0.15}{\frame{\includegraphics[width=\linewidth, trim=0 0 0 0, clip]{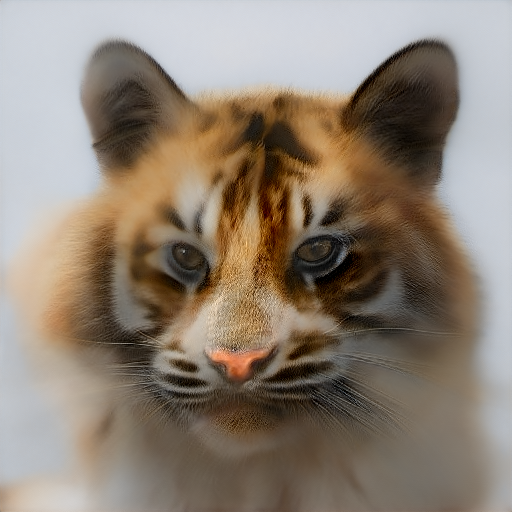}}}\hfill
\mpage{0.15}{\frame{\includegraphics[width=\linewidth, trim=0 0 0 0, clip]{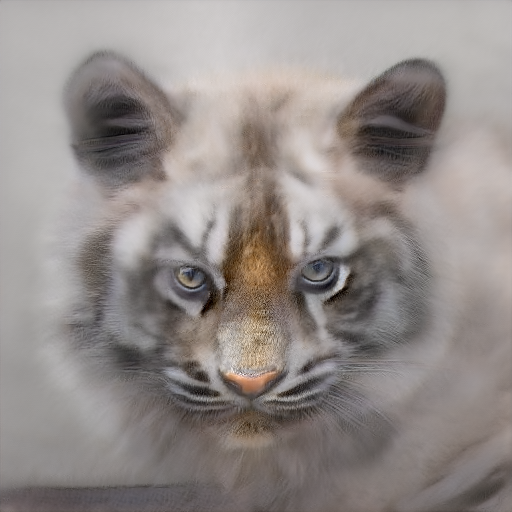}}}\hfill
\mpage{0.15}{\frame{\includegraphics[width=\linewidth, trim=256 0 0 0, clip]{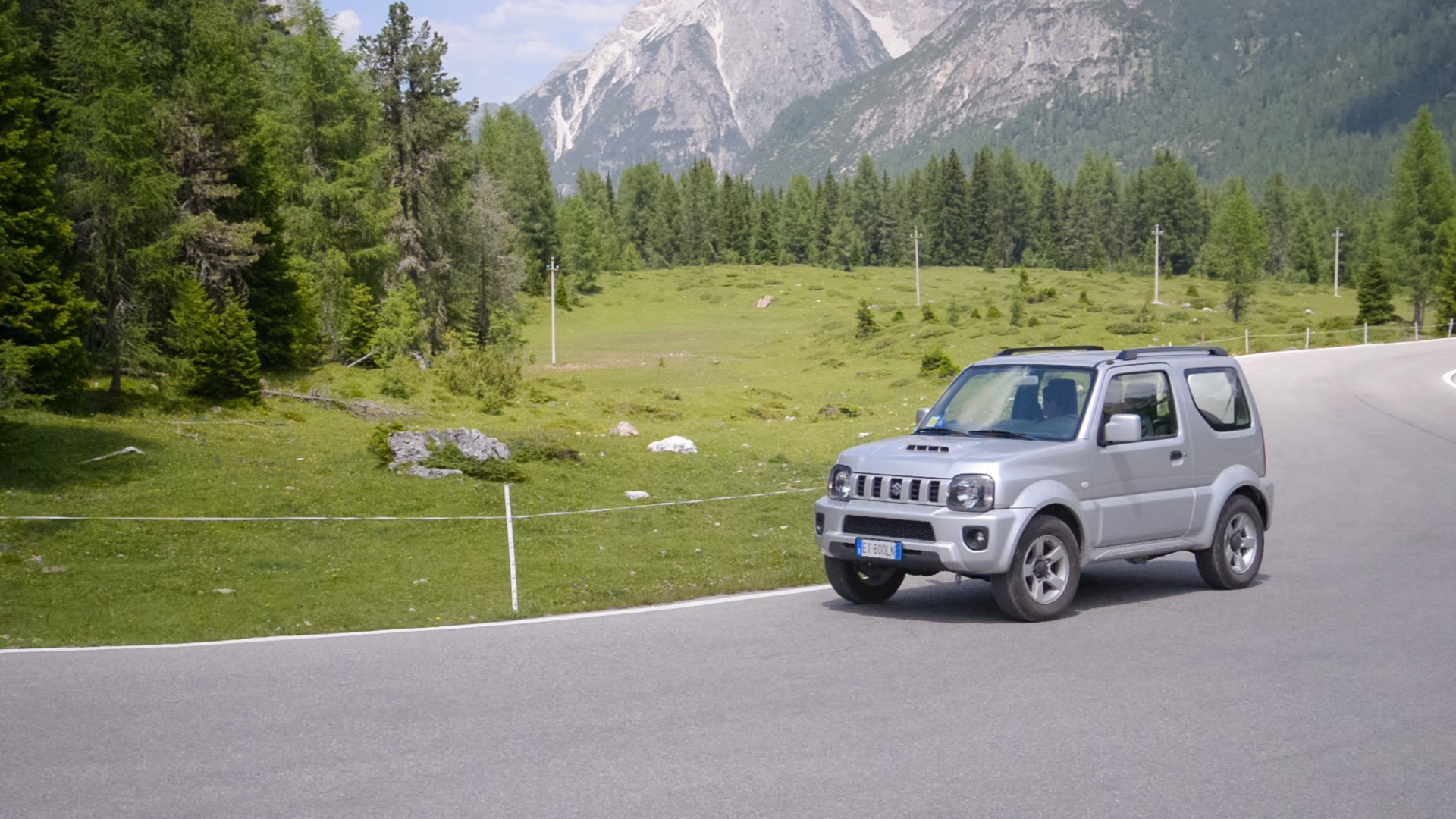}}}\hfill
\mpage{0.15}{\frame{\includegraphics[width=\linewidth, trim=398 0 0 0, clip]{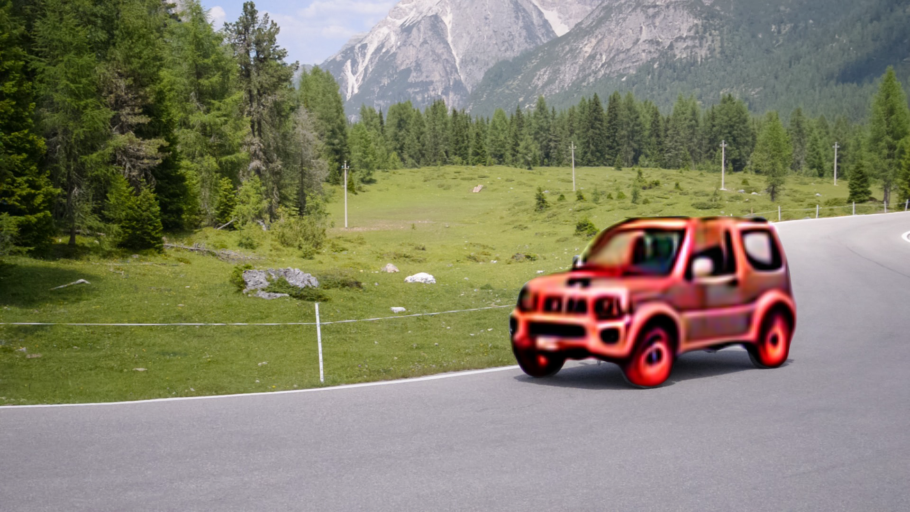}}}\hfill\\

\vspace{1mm}

\mpage{0.15}{{{\small{View 1}}}}\hfill
\mpage{0.15}{{{\small{View 2}}}}\hfill
\mpage{0.15}{{{\small{Sample 1}}}}\hfill
\mpage{0.15}{{{\small{Sample 2}}}}\hfill
\mpage{0.15}{{{\small{Input}}}}\hfill
\mpage{0.15}{{{\small{``rusty car''}}}}\\

\captionof{figure}{
\textbf{Limitations.} Text-to-3D: multi-face,  StyleGAN adaptation: out-of-focus, Layered image editing: over-saturation issue.}
\label{fig:limitation}
\end{figure*}

\footnotetext[1]{{``A high quality photo of a delicious burger.''}}
\footnotetext[2]{{``A high quality photo of a jug made of blue and white porcelain.''}}
CLIP score is used to evaluate the similarity between the text prompt and the output image.
LPIPS score measures all the pairs in the samples to examine the \emph{diversity} of the generated samples. 
Our method achieves competitive CLIP and LPIPS scores compared to the baselines. 
For some domains, \eg, Dog, Hamster, Badger, Otter, and Pig, our LPIPS score outperforms other baselines by a large margin. 
\footnotetext[3]{{``3d cute cat, closeup cute and adorable, cute big circular reflective eyes, long fuzzy fur, Pixar render, unreal engine cinematic smooth, intricate detail, cinematic''}}
\footnotetext[4]{{``3d human face, closeup cute and adorable, cute big circular reflective eyes, Pixar render, unreal engine cinematic smooth, intricate detail, cinematic''}} 
\footnotetext[5]{{``photo of a face [SEP] A realistic detailed portrait, single face, science fiction, artstation, volumetric lighting, octane render''}}
Note that the results shown here have discrepancies with the reported results in StyleGANFusion~\cite{song2022diffusion}.
We consulted with the authors and were informed that the reported results were obtained from \emph{per-dataset tuning}.
We run the experiments with the default parameters provided by the authors.
The qualitative comparisons of human faces and cats are illustrated in Fig.~\ref{fig:comparison_GAN}. 
As reported in~\cite{song2022diffusion}, StyleGAN-NADA cannot handle long prompts well but achieves competitive CLIP scores for short prompts. 
We show results with short prompts, like ``a photo of X'', and long prompts. 
Other baselines tend to have over-smoothed texture (StyleGANFusion) or lower fidelity (StyleGAN-NADA). 
Our method can generate sharper textures and better overall image quality.

\topic{Layered image editing.}
We demonstrate the application on text-driven image editing tasks. 
We follow the layered-editing approach in Text2LIVE~\cite{bar2022text2live}, which utilizes a CNN generator to output an RGBA editing layer blended with the input image. 
The generator is trained with CLIP~\cite{clip} guidance via test-time optimization. 
We adopt the same CNN generator $\theta_{CNN}$ and introduce an additional trainable latent code $\theta_{latent}$, allowing to generate richer details.
The trainable parameter set is $\theta = \{\theta_{CNN}, \theta_{latent}\}$. 
The parameters are updated using a combination of latent score distillation $L_{LSD}$ and the proposed \emph{feature matching loss} $L_{FM}$ along with KL loss $L_{KL}$. 
Besides, we use object mask supervision for the blending alpha map to improve editing quality. 

We compare our method with Text2LIVE, and the latent-score-distillation-only (\ie, $L_{LSD}$) baseline with the same configuration as our model in Fig.~\ref{fig:comparison_layered_editing}.
Text2LIVE fails to manipulate the appearance effectively and produces noticeable artifacts, particularly in the ``zebra'' case.
On the other hand, while the $L_{LSD}$-only baseline can guide the editing effectively, it fails to generate fine details. 
It cannot adequately synthesize the zebra stripes in Example 2 and the tiger's face in Example 3. 
In contrast, our proposed method can generate more detailed results than the $L_{LSD}$-only baseline, such as the feathers of the white swan in Example 1. 
Overall, our proposed method produces visually better results and richer details than the baselines.

\subsection{Ablation}
We conduct an ablation study in Fig.~\ref{fig:ablation} to assess the effectiveness of the feature matching loss $L_{FM}$ and the KL regularizer $L_{KL}$, along with the baseline latent score distillation loss $L_{LSD}$ in a simple image-synthesis task. The results demonstrate that incorporating $L_{FM}$ and $L_{KL}$ improved the overall image quality performance. Specifically, the feature matching loss $L_{FM}$ is able to bring more detail to the image, while the KL regularizer mitigates the color over-saturation problem. For more quantitative and qualitative results of ablation studies on the three applications, please refer to the supplementary material.

\section{Implementation Details}
In this section, we elaberate our implemantation detail for each application.
\subsection{Text-to-3D.}
\label{sec:text}
We evaluate our method on two text-guided 3D generative models, Latent NeRF~\cite{metzer2022latent} and Jacobian NeRF~\cite{sjc}. We follow the original setup of using Stable Diffusion~\cite{rombach2022high} v1.4 for Latent NeRF and Stable Diffusion v1.5 for Jacobian NeRF. 
We integrate their method with our $L_{FM}$ and KL regularizer. For Jacobian NeRF, we run experiments for 10,000 iterations, and for Latent NeRF, we run experiments for 5,000 iterations with an additional 2,000 refined iterations, following the default setting.
The objective after integrating our method will be $L=\lambda_{1}L_{FM}+\lambda_{2}L_{KL}+\lambda_{3}L_{LSD}$, where $\lambda_1=10^{-1}, \lambda_2=10^{-1}, \lambda_3=1.0$ for Jacobian NeRF and $\lambda_1=10^{-2}, \lambda_2=10^{-1}, \lambda_3=1.0$ for Latent NeRF.

\subsection{StyleGAN adaptation.}
\label{sec:gan}
We implement our method on StyleGAN2~\cite{stylegan2} with our feature matching loss $L_{FM}$. 
Our method is built upon StyleGANFusion~\cite{song2022diffusion}.
We add our $L_{FM}$ and KL regularizer. 
We run the experiments for 2,000 training iterations and a learning rate of 0.002. 
We generate 2,000 samples from the adapted generator after the training for the quantitative evaluations.
The objective after integrating our method will be $L=\lambda_{1}L_{FM}+\lambda_{2}L_{KL}+\lambda_{3}L_{LSD}$, where $\lambda_1=3.0, \lambda_2=0,5 \times 10^{-1}, \lambda_3=1.0$.

\subsection{Layered image editing}
\label{sec:layer}
\begin{figure}
    \centering
    \includegraphics[width=\linewidth]{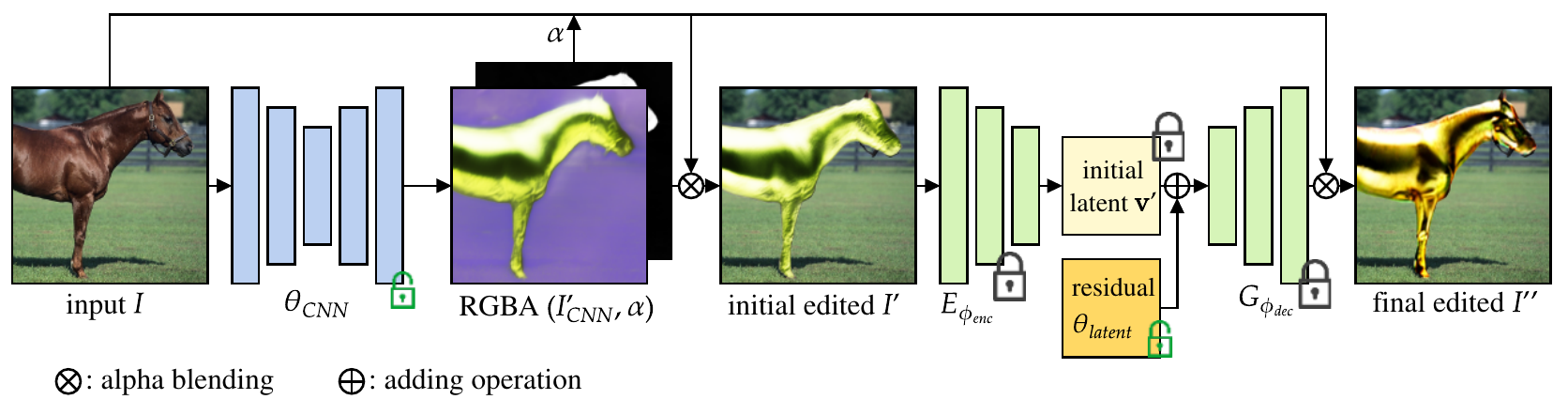}
    \caption{\textbf{Layered image editing pipeline.}}
    \label{fig:supp_layered_edit_method}
\end{figure}

We follow the layered image editing approach of Text2LIVE~\cite{bar2022text2live}. 
As illustrated in Fig.~\ref{fig:supp_layered_edit_method}, we first feed the input image $I$ to the CNN generator of Text2LIVE to produce an initial editing layer and alpha map $(I'_{CNN}, \alpha)$. 
The initial edited image can be synthesized by alpha blending $I'=I'_{CNN}\cdot\alpha+I\cdot(1-\alpha)$. 
We then input the initial edited image $I'$ into the frozen encoder $E_{\phi_{enc}}$ of the latent diffusion model~\cite{rombach2022high} to obtain the initial latent $\mathbf{v}'$. 
We exploit an additional learnable residual latent $\theta_{latent}$ to add to the initial latent $\mathbf{v}''=\mathbf{v}'+\theta_{latent}$ to obtain richer details. 
Lastly, the frozen decoder $G_{\phi_{dec}}$ generates the final edited image $I''$ from the latent $\mathbf{v}''$ followed by the same alpha blending. The parameter set $\{\theta_{CNN}, \theta_{latent}\}$ (\ie, the parameters of Text2LIVE's CNN generator and the additional residual latent) is trained by the diffusion prior, either the $L_{LSD}$-only baseline or our proposed $L_{FM} + L_{KL} +L_{LSD}$ full loss. 
We compute the diffusion-guided losses on $\mathbf{v}''$ and $I''$. 
In addition, to enhance the editing quality, we use an additional mask supervision loss on learning the alpha map $\alpha$. The final loss is $L_{final}=\lambda_{1}L_{FM}+\lambda_{2}L_{KL}+\lambda_{3}L_{LSD} + \|\alpha-M\|_1$, where $\lambda_1=10^{-5}, \lambda_2=10^{-7}, \lambda_3=10^{-6}$, and $M$ is the mask obtained from MaskRCNN~\cite{maskrcnn}.

\label{sec:layer}

\subsection{Image Synthesis}
\label{sec:img}
We conduct an ablation study of the two primary components of our proposed method, the
feature matching loss $L_{FM}$ and the KL regularizer $L_{KL}$ on a straightforward image synthesis task. We begin by initializing a $64 \times 64 \times 4$ random noise map and then optimizing it with a combination of different losses for 1000 iterations. The final results are rendered from the decoder of the pretrained diffusion model. 
We use the AdamW optimizer with a learning rate of $10^{-1}$ for optimization. 
The objective of the full model is  $L=\lambda_{1}L_{FM}+\lambda_{2}L_{KL}+\lambda_{3}L_{LSD}$, where $\lambda_1=3.0, \lambda_2=10^{-1}, \lambda_3=1.0$.

\section{Limitations}
\label{sec:limitations}

In Text-to-3D task, the learned 3D objects sometimes encounter the multiple faces Janus problem (Fig.~\ref{fig:limitation} first case), in which the model will appear with more than one front face. 
The problem also comes up in DreamFusion, which might be caused by less knowledge of 3D geometry in the diffusion model. 
Or as mentioned in DreamFusion and Latent NeRF, the text prompt often interprets from the canonical views and might not be a good condition for sampling other viewpoints. Besides, with the guidance of diffusion score distillation, we found that few cases would suffer from the issue of color over-saturation and out-of-focus despite the use of our KL loss $L_{KL}$ (Fig.~\ref{fig:limitation} second and third case). 
The over-saturation issue can also be found in previous score distillation works~\cite{metzer2022latent,poole2022dreamfusion}.

\section{Conclusions}
\label{sec:conc}

In this paper, we proposed a framework that uses a diffusion model as a prior for visual synthesis tasks. 
Our core contributions are a feature-matching loss to extract detailed information and a KL loss for regularization. Through extensive experimental evaluations, we show that our method improves the quality compared to strong baselines on text-to-3D, StyleGAN adaptation, and layered image editing tasks.

{\small
\bibliographystyle{ieee_fullname}
\bibliography{egbib}
}

\end{document}